\documentclass{style/naturep}

\newboolean{isarxiv}
\setboolean{isarxiv}{true} 

\newcommand{\maybeClearpage}{
  \ifthenelse{\boolean{isarxiv}}{
  }{
    \clearpage 
  }
}

\usepackage{amssymb}
\usepackage{amsmath}
\usepackage{algorithmicx}
\usepackage{algorithm}
\usepackage{adjustbox}
\usepackage{graphicx}
\usepackage{siunitx}
\usepackage[noend]{algpseudocode}
\usepackage{bbm}
\usepackage{lineno}
\usepackage[margin=0.6in]{geometry}
\usepackage{ragged2e}
\usepackage{setspace}
\usepackage{longtable}
\usepackage{hyperref}
\usepackage{cleveref}
\usepackage{lineno}
\usepackage{setspace}
\usepackage{anyfontsize}
\usepackage{float}
\usepackage{booktabs}
\usepackage{multirow}
\usepackage{blindtext}
\usepackage{tabularx}
\usepackage{caption}
\usepackage{subcaption}
\usepackage{enumitem, kantlipsum}
\usepackage{xspace}
\usepackage{pifont}

\usepackage{amsmath}
\usepackage{amssymb}
\usepackage{booktabs}
\usepackage{environ}
\usepackage{float}
\usepackage{ifthen} 
\usepackage{placeins}
\usepackage[table,x11names]{xcolor}
\definecolor{maroon}{cmyk}{0,0.87,0.68,0.32}
\definecolor{gray}{rgb}{0.3,0.3,0.3}

\newcommand{\ours}{\textsc{Threads}\xspace}
\newcommand{\mean}{\textsc{Mean Pooling}\xspace}
\newcommand{\prism}{\textsc{Prism}\xspace}
\newcommand{\gigapath}{\textsc{GigaPath}\xspace}
\newcommand{\chief}{\textsc{Chief}\xspace}
\newcommand{\conch}{\textsc{CONCHv1.5}\xspace}
\newcommand{\ctranspath}{\textsc{CTransPath}\xspace}
\newcommand{\virchow}{\textsc{Virchow}\xspace}
\newcommand{\abmil}{\textsc{ABMIL}\xspace}
\newcommand{\mbtg}{\textsc{MBTG-47k}\xspace}
\newcommand{\resnet}{\textsc{ResNet50-IN}\xspace}

\usepackage{amsmath,amsfonts,bm}

\def\valpha{{\bm{\alpha}}}

\def\vs{{\mathbf{s}}}

\def\mX{{\mathbf{X}}}

\newcommand{\R}{\mathbb{R}}

\newcommand{\softmax}{\mathrm{softmax}}

\newcommand{\gelu}{\textsc{Gelu}\xspace}
\newcommand{\relu}{$\mathrm{ReLU}$\xspace}

\newcommand\Heading[1]{
  \noindent\textbf{\Large{#1}}
}

\newcommand\heading[1]{
  \noindent\textbf{\large{#1}}
}

\newcommand\hheading[1]{
  \noindent\textbf{#1}
}

\title{\begin{flushleft}{\begin{spacing}{1}
     Molecular-driven Foundation Model for Oncologic Pathology
\end{spacing}}\end{flushleft}}

\begin{document}

\maketitle

\ifthenelse{\boolean{isarxiv}}{
  \renewcommand{\baselinestretch}{1} 
}{
  \doublespacing
}

\maketitle
\vspace{-20mm}
\begin{spacing}{1.8}
\noindent
Anurag Vaidya$^{1,2,3,4,\boldsymbol{\ddag}}$,
Andrew Zhang$^{1,2,3,4,\boldsymbol{\ddag}}$,
Guillaume Jaume$^{1,2,3,\boldsymbol{\ddag}}$,
Andrew H. Song$^{1,2,3,\boldsymbol{\ast}}$,
Tong Ding$^{1,2,3,5,\boldsymbol{\ast}}$,
Sophia J. Wagner$^{1,6,7,\boldsymbol{\ast}}$,
Ming Y. Lu$^{1,2,3,8}$,
Paul Doucet$^{1}$,
Harry Robertson$^{1,9}$,
Cristina Almagro-Pérez$^{1,2,3,4}$,
Richard J. Chen$^{1,2,3}$,
Dina ElHarouni$^{3,10}$,
Georges Ayoub$^{3,10}$,
Connor Bossi$^{3,10}$,
Keith L. Ligon$^{1,3,10,11}$,
Georg Gerber$^{1}$,
Long Phi Le$^{2,\boldsymbol{+}}$,
Faisal Mahmood$^{1,2,3,12,\boldsymbol{+}}$
\end{spacing}
\vspace{-10mm}

\begin{spacing}{1.2}
\begin{affiliations}
    \item Department of Pathology, Brigham and Women's Hospital, Harvard Medical School, Boston, MA
    \item Department of Pathology, Massachusetts General Hospital, Harvard Medical School, Boston, MA
    \item Cancer Program, Broad Institute of Harvard and MIT, Cambridge, MA
    \item Health Sciences and Technology, Harvard-MIT, Cambridge, MA
    \item Harvard John A. Paulson School of Engineering and Applied Sciences, Harvard University, Cambridge, MA
    \item Helmholtz Munich – German Research Center for Environment and Health, Munich, Germany
    \item School of Computation, Information and Technology, TUM, Munich, Germany
    \item Electrical Engineering and Computer Science, Massachusetts Institute of Technology (MIT), Cambridge, MA
    \item  Sydney Precision Data Science Center, The University of Sydney, Camperdown, New South Wales, Australia
    \item Department of Oncologic Pathology, Dana-Farber Cancer Institute, Boston, MA 02215, USA
    \item Department of Pathology, Boston Children’s Hospital, Boston, MA 02115, USA
    \item Harvard Data Science Initiative, Harvard University, Cambridge, MA \\
    $\boldsymbol{\ddag}$ Co-first authors, 
    $\boldsymbol{\ast}$ Co-second authors, 
    $\boldsymbol{+}$ Co-senior authors\\
 \textbf{Lead Contact}: Faisal Mahmood (faisalmahmood@bwh.harvard.edu) 
\end{affiliations}
\end{spacing}


\vspace{-0.6cm}
\begin{spacing}{1.}
\noindent\textbf{Foundation models are reshaping computational pathology by enabling transfer learning, where models pre-trained on vast datasets can be adapted for downstream diagnostic, prognostic, and therapeutic response tasks. Despite these advances, foundation models are still limited in their ability to encode the entire gigapixel whole-slide images without additional training and often lack complementary multimodal data. Here, we introduce $\ours$, a slide-level foundation model capable of generating universal representations of whole-slide images of any size. $\ours$ was pretrained using a multimodal learning approach on a diverse cohort of 47,171 hematoxylin and eosin (H\&E)-stained tissue sections, paired with corresponding genomic and transcriptomic profiles—the largest such paired dataset to be used for foundation model development to date. This unique training paradigm enables $\ours$ to capture the tissue’s underlying molecular composition, yielding powerful representations applicable to a wide array of downstream tasks. In extensive benchmarking across 54 oncology tasks, including clinical subtyping, grading, mutation prediction, immunohistochemistry status determination, treatment response prediction and survival prediction $\ours$ outperformed all baselines while demonstrating remarkable generalizability and label efficiency. It is particularly well-suited for predicting rare events, further emphasizing its clinical utility. We intend to make the model publicly available for the broader community.
}
\end{spacing}

\clearpage
\maybeClearpage

\Heading{Introduction}

\noindent With the advancement of precision medicine and targeted therapies, problem statements in oncology focus increasingly on rare conditions and targeted populations. As research questions become more specific, the assumption of data abundance, which underpinned early successes in AI for pathology\cite{song2023artificial,van2021deep}, such as Gleason grading\cite{bulten2022artificial} and metastasis detection\cite{bejnordi2017diagnostic}, no longer applies. Many current problem statements in oncology, especially for patient prognostication, and treatment response prediction, involve small patient cohorts, frequently fewer than 100 patients. The limitation of data scarcity is compounded by the size of digitized tissue sections (whole-slide images, WSIs), which can be several gigabytes each. Consequently, most computational pathology predictive models operate in a scenario where the input data size vastly exceeds the number of available samples for model training, making model training incredibly complex.

In response to these challenges, numerous foundation models specifically designed for pathology have been developed\cite{chen2024towards,xu2024whole,vorontsov2024foundation}. These models enable transfer learning from large pretraining data with hundreds of thousands of WSIs and billions of cells to narrow applications, such as biomarker prediction. However, most of these models are patch encoders and, by design, are restricted to encoding small regions of interest, orders of magnitude smaller than clinical whole-slide imaging data, which can be several gigabytes. Existing models address the limitation by training an additional model, which can be computationally expensive to train and may require a lot of downstream labels. Addressing this limitation is critical for advancing foundation models in pathology to more varied tasks and overcoming the data abundance requirement. Some models have explored whole-slide image representation learning to derive off-the-shelf slide embeddings that can be used for various downstream tasks at minimal cost\cite{chen2022scaling,jaume2024transcriptomics,jaume2024multistain,xu2024whole,wang2024chief}. However, they remain limited in scope by the diversity of training data with organ and disease-specific models\cite{jaume2024transcriptomics,jaume2024multistain}, and the predictive capabilities of the resulting representations.

\begin{figure}[!hbt]
\centering
\includegraphics[width=0.95\textwidth]{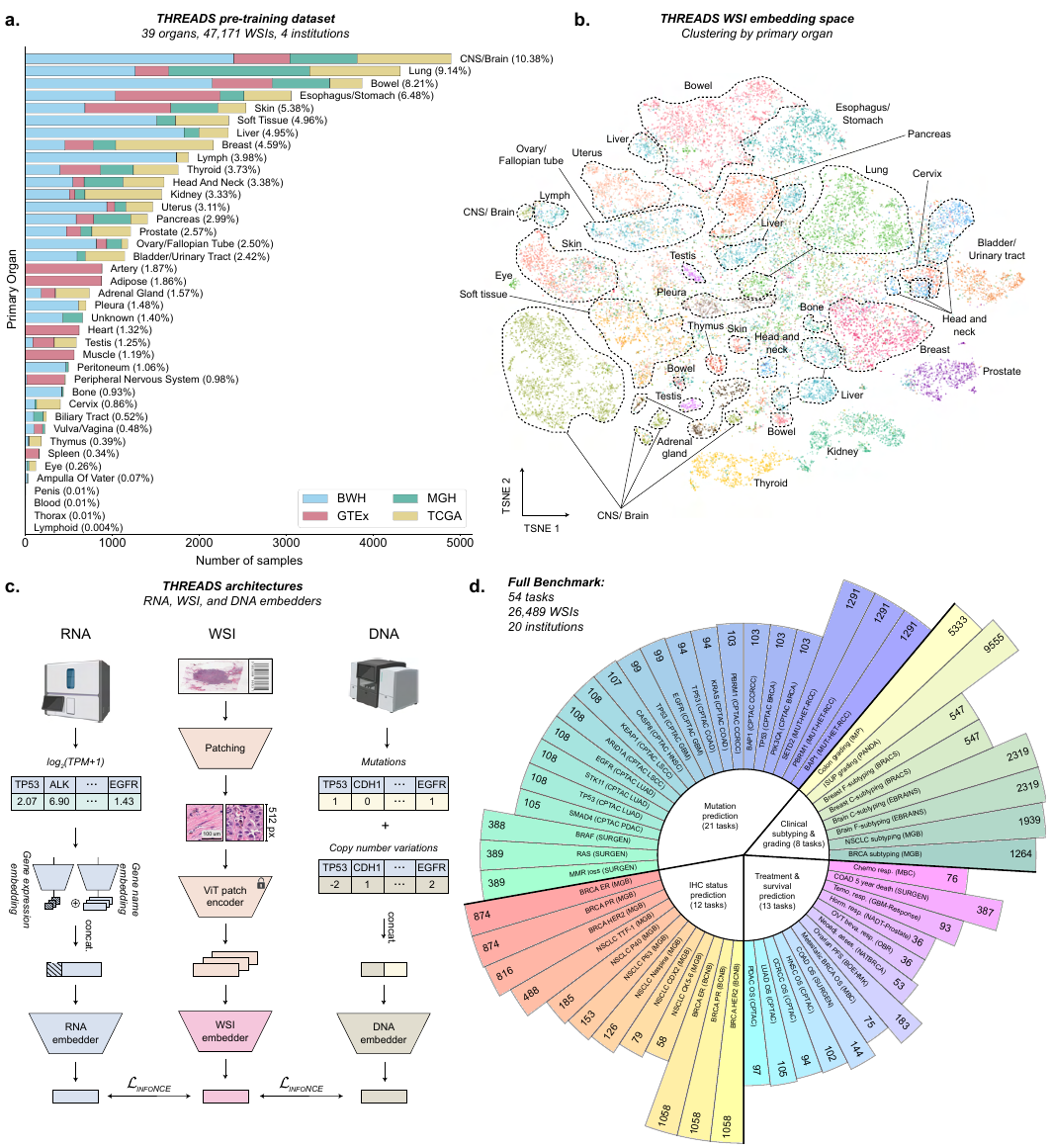}
\caption{\textbf{Study overview.}
\textbf{a.} Tissue site distribution of \mbtg used for $\ours$ pretraining.
\textbf{b.} 2-dimensional tSNE \cite{van2008visualizing} representation of $\ours$ WSI embedding space on \mbtg colored by primary organ. Each point is a WSI.
\textbf{c.} Block diagram of $\ours$ architecture for WSI representation learning. 
\textbf{d.} Overview of $\ours$ downstream evaluation composed of 54 tasks. Tasks are grouped into four families: clinical subtyping and grading (n=8 tasks), gene mutation prediction (n=21 tasks), immunohistochemistry status prediction (n=12 tasks), and treatment response and survival prediction (n=13 tasks).
WSI: whole-slide image;
tSNE: t-distributed stochastic neighbor embedding;
MGH: Mass General Hospital;
BWH: Brigham and Women's Hospital.
}
\label{fig:fig1}
\end{figure}

Here, we introduce a new foundation model for pathology, \ours, a general-purpose encoder model that can generate WSI embeddings. $\ours$ was pretrained through multimodal contrastive learning, where molecular profiles obtained with next-generation sequencing are used as a guide for learning the slide representation. We posit that molecular data brings a holistic and unbiased view of the tissue morphology that encapsulates biologically and clinically relevant information\cite{jaume2024transcriptomics,jaume2024multistain}. To train $\ours$, we assembled the most extensive multimodal training dataset to date, named \mbtg, consisting of more than 47,000 samples. Each sample includes a WSI and its corresponding molecular profile obtained from an adjacent tissue section (\textbf{Figure \ref{fig:fig1}.a}). \mbtg was curated from Massachusetts General Hospital (MGH, 6,899 samples, or 14.6\%), Brigham and Women’s Hospital (BWH, 20,556 samples or 43.6\%), The Cancer Genome Atlas Program (TCGA, 10,209 samples or 21.6\%), and the Genotype–Tissue Expression (GTEx, 9,507 samples or 20.2\%) consortium\cite{gtex2015genotype} (\textbf{Extended Data Table \ref{tab:pretrain-counts}}). This pretraining strategy leads to a slide embedding space that encodes rich information about tissue morphology, disease, and composition (\textbf{Figure \ref{fig:fig1}.b}). 

We validate \ours across a wide range of tasks in oncology, covering clinical tasks for cancer subtyping and grading, gene mutation prediction, immunohistochemistry status prediction, and treatment response and survival prediction (\textbf{Figure \ref{fig:fig1}.d}). In total, our model is evaluated on 54 pathology tasks from 23 cohorts across 17 different sources. \ours achieves state-of-the-art performance, significantly outperforming three whole-slide encoder models \prism\cite{shaikovski2024prism} (P-value$<$0.001), \gigapath\cite{xu2024whole} (P-value$<$0.001) and \chief\cite{wang2024chief} (P-value$<$0.001), and attention-based multiple instance learning classification baselines (P-value$<$0.001). \ours can also serve as an effective initialization for additional model fine-tuning, which brings a significant improvement over training a model from scratch (P-value$<$0.001). This establishes \ours as a foundational model that can drive AI advancements in histopathology.

\Heading{Results}

\hheading{Whole-slide image classification with \ours}

\noindent \textbf{\ours design.} \ours consists of two components. An ROI encoder model (\conch\cite{lu2024towards}) consisting of a Vision Transformer-Large\cite{dosovitskiy2021image,vaswani2017attention} (ViT-L) model trained on millions of image patches via multimodal learning between ROIs and text captions; and a slide encoder that aggregates tile embeddings into a slide representation using attention-based modeling (\textbf{Figure \ref{fig:fig1}.c} and \textbf{Extended Data Figure \ref{fig:edf1}}). We use two types of next-generation sequencing (NGS) data for \ours pretraining: transcriptomic profiles obtained with bulk RNA sequencing (MGH, TCGA, and GTEx samples), and genomic profiles capturing single nucleotide variants (SNVs), insertions and deletions (indels), and copy number variants (CNVs) of a targeted gene panel (BWH samples). The transcriptomic profiles are encoded using a single-cell foundation model pretrained on 5.7 million cells of various cancer types\cite{cui2024scgpt}, and the genomic profiles using a multi-layer perceptron model\cite{chen2020pathomic}. We employ cross-modal contrastive learning to align the slide representation with the corresponding molecular embedding. Additional information is provided in the \textbf{Online Methods, Pretraining dataset curation}. 

\noindent \textbf{Downstream evaluation.} We propose a large benchmark for assessing foundation models in hematoxylin and eosin (H\&E)-stained whole-slide imaging. Our evaluation includes 54 tasks from 23 different cohorts, covering four families of tasks: clinical subtyping and grading (n=8 tasks, 20,427 WSIs), gene mutation prediction (n=21 tasks, 3,503 WSIs), immunohistochemistry (IHC) status prediction (n=12 tasks, 2,469 WSIs), and patient prognostication including treatment response and survival prediction (n=13 tasks, 2,857 WSIs). We curated tasks from a set of in-house data (n=12 tasks including 3,550 WSIs from three cohorts) and publicly available data (n=42 tasks including 23,161 WSIs from 17 cohorts). The diversity of tasks makes our benchmark suitable for assessing the predictive performance of slide encoders under different scenarios, from well-established clinical tasks with data abundance, such as colorectal cancer grading and breast cancer subtyping, to specific problem statements in treatment response prediction typically characterized by small patient cohorts. Our evaluation constitutes, to date, the most comprehensive benchmark introduced in computational pathology. All tasks follow a unified evaluation with either five-fold cross-validation into 80:20 splits or 50-train-test splits, depending on cohort size with label- and patient-stratified splits. An overview of each evaluation task is provided in \textbf{Figure \ref{fig:fig1}.d}, with additional descriptions in the \textbf{Online Methods}, \textbf{Downstream tasks and datasets}, statistics for each task in \textbf{Extended Data Table \ref{tab:task_summary.morphological_subtyping},\ref{tab:task_summary.tumor_grading},\ref{tab:task_summary.molecular_subtyping},\ref{tab:task_summary.mutation_prediction},\ref{tab:task_summary.treatment_response},\ref{tab:task_summary.survival_prediction}}, and links to access public cohorts detailed in \textbf{Extended Data Table \ref{tab:downstream-links}}.

\noindent \textbf{Baselines.} We compare \ours against three foundation models for encoding WSIs: \prism, \gigapath, and \chief. \prism is based on the Virchow\cite{vorontsov2024foundation} patch encoder (ViT-Huge, 632 million parameters) followed by a Perceiver model\cite{jaegle2021perceiver} (45 million parameters) pretrained using contrastive learning with matched patient-level pathology reports (195,344 specimens). \gigapath is based on a ViT-Giant patch encoder (1.13 billion parameters) pretrained on 171,000+ WSIs ($>$30,000 patients) using DINOv2\cite{oquab2023dinov2}, and a LongNet\cite{ding2023longnet} slide encoder pretrained using masked autoencoding. Finally, \chief is based on the CTransPath\cite{wang2021transpath,wang2022transformer} patch encoder (Swin-Transformer, 28 million parameters) followed by an attention-based multiple instance learning (ABMIL) model\cite{ilse2018attention} pretrained on 60,530 WSIs using contrastive learning with the tissue site. Additional information is provided in the \textbf{Online Methods} and \textbf{Baselines}. 

\noindent We evaluate \ours and baselines using linear probing (i.e., by learning a logistic regression model) to classify slide embeddings into the downstream task label. To prevent overfitting, we avoid hyperparameter tuning and set a fixed cost, number of iterations, and solver in linear probe models. We evaluate model performance using the area under the receiver operating characteristic (AUC) for all binary classification tasks, quadratic Cohen's kappa for grading, balanced accuracy for multi-class clinical subtyping, and concordance-index (c-index) for survival tasks. We provide additional information in the \textbf{Online Methods}, \textbf{Evaluation metrics}.

\begin{figure}[!htp]
\centering
\includegraphics[width=0.77\textwidth]{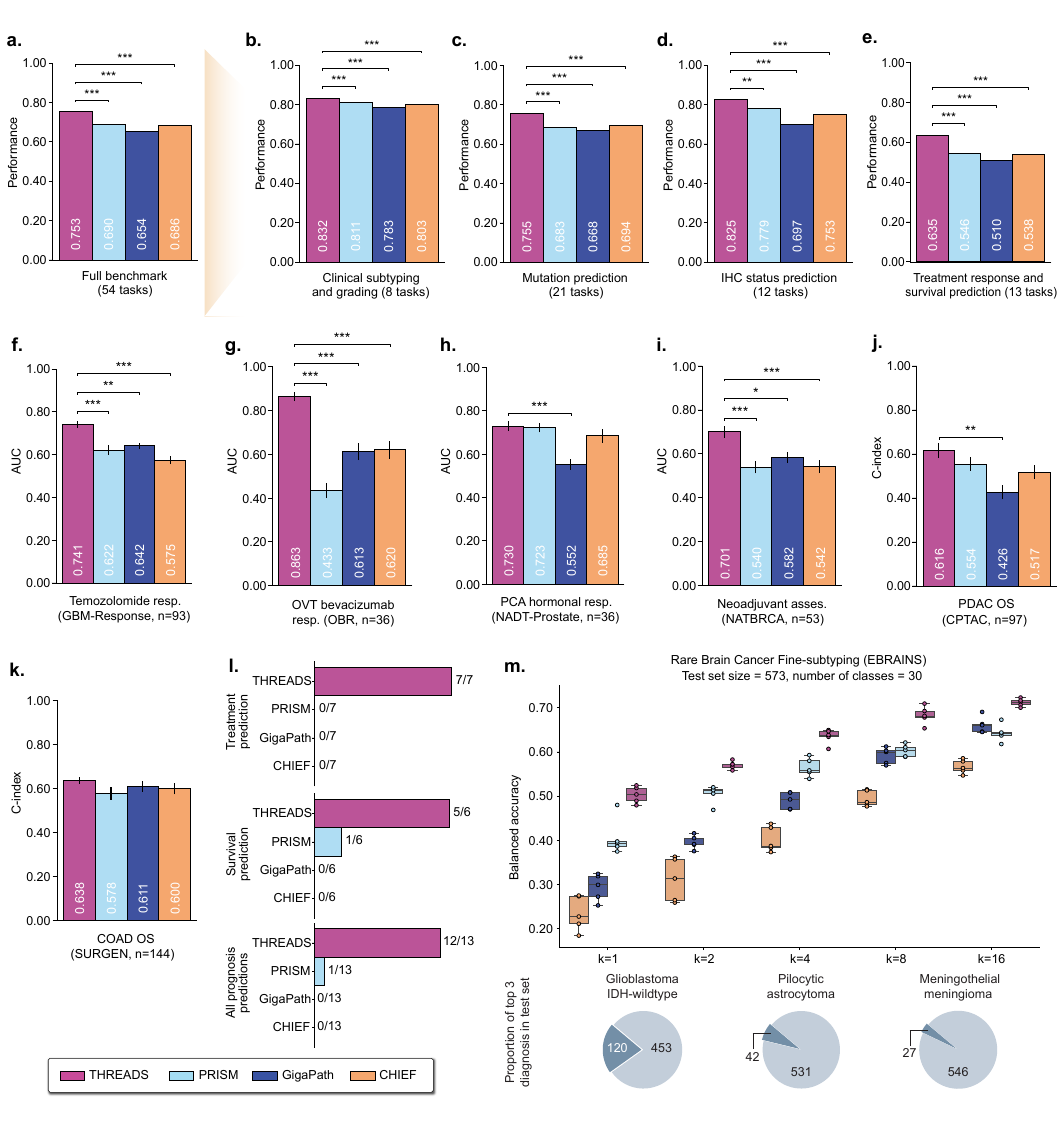}
\caption{\textbf{Evaluation of \ours and baselines with linear probing.}
\textbf{a.} Average performance of \ours and baselines on 54 tasks. \ours is compared against \prism, \gigapath, and \chief.
Average performance per family of tasks: 
\textbf{b.} clinical subtyping and grading (8 tasks), 
\textbf{c.} mutation prediction (21 tasks),
\textbf{d.} IHC status prediction (12 tasks), and 
\textbf{e.} treatment response and survival prediction tasks (13 tasks). 
\textbf{f--k} \ours performance on treatment response and prognostication tasks characterized by label scarcity (n=36 to n=144 patients). Binary tasks (\textbf{f--i}) are measured with AUC. Survival tasks (\textbf{j,k}) are measured with concordance-index.
\textbf{f.} Temozolomide treatment response in glioblastoma (GBM).
\textbf{g.} Bevacizumab treatment response in ovarian cancer (OV).
\textbf{h.} Neoadjuvant response assessment in invasive breast cancer (BRCA).
\textbf{i.} Hormonal therapy response in prostate adenocarcinoma (PRAD).
\textbf{j.} Overall survival (OS) prediction in pancreatic ductal adenocarcinoma (PDAC).
\textbf{k.} Overall survival prediction in colon adenocarcinoma (COAD).
\textbf{l.} Number of tasks where each model (\ours and baselines) reaches highest performance across all tasks (n=54 tasks), treatment response (n=6 tasks) and survival tasks (n=7 tasks). 
\textbf{m.} Few-shot learning performance of \ours against baselines in brain tumor subtyping. $k$ refers to the number of training samples per class. 
Error bars represent the standard error measured across multiple folds.
Boxes indicate quartile values of model performance (n=5 runs), and whiskers extend to data points within 1.5-fold the interquartile range.
Task-wise P-values were determined using two-sided Tukey Honest Significance Difference tests accounting for multiple comparisons following a positive result (P$<$0.05) of a two-way ANOVA.
Statistical significance across multiple tasks (e.g., for each family) was assessed using a mixed-effects model. 
P$<$0.05: *, P$<$0.01: **, P$<$0.001: ***.
 }
\label{fig:fig2}
\end{figure}

\noindent \textbf{Linear probing.} \ours provides state-of-the-art performance in linear probing evaluation. \ours leads to an absolute performance gain over \prism, \gigapath, and \chief of 6.3\%, 9.9\%, 6.7\%, respectively (\textbf{Figure \ref{fig:fig2}.a}). Employing a mixed-effects statistical model to compare the overall performance (\textbf{Online Methods, Statistical analysis}), we showed that \ours significantly outperformed \prism (P$<$0.001), \chief (P$<$0.001), and \gigapath (P$<$0.001). When investigating each family of tasks, \ours demonstrates absolute performance improvements of
2.1\% in clinical subtyping and grading over \prism, the second-best model (P$<$0.001, \textbf{Figure \ref{fig:fig2}.b}),
6.1\% in mutation prediction over \chief (P$<$0.001, \textbf{Figure \ref{fig:fig2}.c}),
4.6\% in IHC status prediction over \prism (P$<$0.01, \textbf{Figure \ref{fig:fig2}.d}), and 
8.9\% in prognostication over \prism (P$<$0.001, \textbf{Figure \ref{fig:fig2}.e}).
At the individual task level, \ours outperforms \prism in 44/54 tasks, \chief in 49/54 tasks, and \gigapath in 54/54 tasks. The performance per task is detailed in
\textbf{Extended Data \cref{tab:allshot_mgb_brca,tab:allshot_mgb_lung,tab:allshot_bcnb,tab:allshot_mut-het-rcc,tab:allshot_imp,tab:allshot_panda,tab:allshot_cptac_brca,tab:allshot_cptac_ccrcc,tab:allshot_cptac_coad,tab:allshot_cptac_gbm,tab:allshot_cptac_hnsc,tab:allshot_cptac_lscc,tab:allshot_cptac_luad,tab:allshot_cptac_pda,tab:allshot_bracs,tab:allshot_ebrains,tab:allshot_ovarian,tab:allshot_nadt,tab:allshot_gbm_rt_tmz,tab:allshot_natbrca,tab:allshot_sr386_,tab:allshot_mbc_,tab:survival_cptac_pda,tab:survival_cptac_luad,tab:survival_cptac_ccrcc,tab:survival_cptac_hnsc,tab:survival_sr386_,tab:survival_mbc_,tab:survival_boehmk_}}.

\noindent When investigating performance in diagnostic tasks (cancer subtyping and grading), \ours with a linear model achieves performance levels that are competitive with specialist models\cite{bulten2022artificial,oliveira2023cad}. For instance, \ours reaches 98.3\% AUC in breast cancer subtyping (invasive lobular carcinoma \textit{vs.} invasive ductal carcinoma), and 98.2\% AUC in lung cancer subtyping (lung adenocarcinoma \textit{vs.} lung squamous cell carcinoma). In comparison, an attention-based MIL model\cite{ilse2018attention} trained on the same data with $>$1.5 million parameters reaches 98.3\% and 98.0\%, respectively (\textbf{Extended Data Table \ref{tab:allshot_mgb_brca} and \ref{tab:allshot_mgb_lung}}). In colorectal cancer grading, \ours with linear probing reaches 91.9\% quadratic Cohen's kappa score, just 2.3\% lower than training a dedicated ABMIL model (94.2\%) (\textbf{Extended Data Table \ref{tab:allshot_imp}}). These findings underscore the extensive capabilities of \ours in providing rich slide representations for clinical use.

\noindent To validate the superior performance of \ours in linear probing evaluation, we additionally benchmarked \ours and baselines with varying the regularization cost (\textbf{Extended Data Figure \ref{fig:edf5} and Extended Data Table \ref{tab:cost-scale-linprobe}}). \ours provides significant performance gain over all baselines for all the regularization costs explored: 4.3\% absolute performance gain over \prism (second-best performer) with large regularization (C=0.01), and 5.7\% over \chief (second-best performer) with small regularization (C=10). \ours is also less sensitive to changes in regularization than baselines across all regularization strengths, showing the model's robustness and versatility.  

\noindent\textbf{Transferability of \ours.} We also investigated whether linear models trained with \ours embedding show generalization properties when tested in external cohorts. To this end, we selected a subset of tasks from our evaluation pipeline for which we have an external test set. Specifically, we first train a linear probe classifier on the entirety of one dataset and test on the entirety of the external set. We studied transferability in six different types of cancer: prediction of BAP1 and PRMB1 mutations in clear cell renal cell carcinoma, IDH mutation in a cohort of glioblastoma and low-grade glioma, prediction of ER/PR status in invasive breast cancer, subtyping in lung cancer, and survival prediction in pancreatic adenocarcinoma. Overall, \ours provides strong transferability properties that lead to significantly better performance than \prism (P-value$<$0.001), \gigapath (P-value$<$0.001), and \chief (P-value$<$0.001) as shown in \textbf{Extended Data Figure \ref{fig:fig3}} and \textbf{Extended Data Table \ref{tab:generalizability}}. \ours outperforms all baselines in 8/9 tasks (task-wise P-values $<$0.001 in 7/9 tasks in comparison to the second-best baseline). In lung and breast cancer subtyping, \ours preserves a high predictive performance of 98.4\% and 96.5\% AUC, respectively, on the external cohort. Similarly, in ER and PR status prediction, \ours leads to 88.5\% and 79.4\% AUC, maintaining high predictive performance. These results highlight the ability of \ours to capture clinically and biologically relevant information without overfitting on cohort- and institution-specific features. 

\hheading{Data and label efficiency of \ours}
 
\noindent As pathology and oncology progress, problem statements become increasingly specific, resulting in scenarios that inherently face constraints in data availability. Such limitations are particularly prevalent in predicting patient treatment response and resistance. As part of our evaluation pipeline, we curated six treatment prediction tasks and one treatment assessment task covering several cancer types. Specifically, we tested \ours to predict response in patients treated with
temozolomide in glioblastoma (93 patients, \textbf{Figure~\ref{fig:fig2}.f}),
bevacizumab in ovarian cancer (36 patients, \textbf{Figure~\ref{fig:fig2}.g}), 
hormonal therapy in prostate adenocarcinoma (53 patients, \textbf{Figure~\ref{fig:fig2}.h}), and
neoadjuvant chemotherapy in high-grade serous ovarian cancer (183 patients, \textbf{Extended Data Table \ref{tab:survival_boehmk_}}), and
platinum (and taxane for a subset) in ovarian metastasis of metastatic breast cancer (75 patients, \textbf{Extended Data Table \ref{tab:survival_mbc_}}). 
\ours also provides a tool for response assessment to detect signs of vascular invasion in patients with breast cancer treated with neoadjuvant chemotherapy (53 WSIs, \textbf{Figure~\ref{fig:fig2}.i}).
\ours provides better performance than baselines in all seven tasks, overall significantly outperforming baselines as assessed using a mixed-effect model statistical analysis (P-value$<$0.001, \textbf{Figure~\ref{fig:fig2}.l}). When considering individual tasks, \ours significantly outperforms all baselines in 4/7 (P-value$<$0.05). 

\ours embeddings can also be used for patient survival prediction. We employ this approach for predicting overall survival in patients with pancreatic adenocarcinoma (\textbf{Figure~\ref{fig:fig2}.k}), colon and rectum adenocarcinoma (\textbf{Figure~\ref{fig:fig2}.l}), clear cell renal cell carcinoma (\textbf{Extended Data Table \ref{tab:survival_cptac_ccrcc}}), head and neck squamous carcinoma (\textbf{Extended Data Table \ref{tab:survival_cptac_hnsc}}), and lung adenocarcinoma (\textbf{Extended Data Table \ref{tab:survival_cptac_luad}}). Across all six survival tasks from our evaluation, \ours provides the best predictive performance in five of them, overall providing significantly better performance than all baselines (P-value$<$0.001, \textbf{Figure~\ref{fig:fig2}.m}). The survival analysis using Kaplan Meier estimators also reveals the superior stratification capabilities of \ours, which provide better separation between groups of patients considered as low and high risk than all baselines (\textbf{Extended Data Figure \ref{fig:edf4}}).

\noindent To complement our analysis in data-scarce problem statements, we benchmarked \ours and baselines in few-shot learning experiments, where we monitor the test performance when training on an increasing number of samples: $k=1,2,4,8,16,32$, where $k$ is the number of training samples per class. We use the GBM-Treatment response dataset for treatment response prediction (\textbf{Extended Data Figure~\ref{fig:edf3}.a} and \textbf{Extended Data Table \ref{tab:fewshot_gbm_rt_tmz}})), EBRAINS dataset\cite{roetzer2022ebrains} for fine-grained (n=30 classes) and coarse-grained (n=12) brain tumor subtyping (\textbf{Figure~\ref{fig:fig2}.m}, \textbf{Extended Data Figure~\ref{fig:edf3}.b}, and \textbf{Extended Data Table \ref{tab:fewshot_ebrains}}), the BRACS dataset\cite{brancati2021bracs} for fine-grained (n=7) and coarse-grained (n=3) breast tumor subtyping (\textbf{Extended Data Figure~\ref{fig:edf3}.c,d} and \textbf{Extended Data Table \ref{tab:fewshot_bracs}}), and the BCNB dataset\cite{xu2021predicting} for ER status prediction (\textbf{Extended Data Figure~\ref{fig:edf3}.e,f} and \textbf{Extended Data Table \ref{tab:fewshot_bcnb}}). \ours provides the best linear probing performance, outperforming baselines for most values of $k$. The predictive capabilities of \ours are particularly highlighted in subtyping rare brain tumors, where \ours performance with $k$=4 is superior to \prism performance (second best performer) with $k$=16 (\textbf{Figure \ref{fig:fig2}.n}).

\hheading{\ours fine-tuning}

\noindent \ours can also serve as weight initialization for further finetuning on a downstream task. This approach combines the benefit of large-scale pretraining while letting the model adapt to the nuances of the downstream application. Here, we fine-tuned a $\ours$-initialized model on every downstream task in our evaluation pipeline. We employ a unified fine-tuning recipe that is applied to all tasks (\textbf{Online Methods}, \textbf{Baselines}). To mitigate overfitting and costly hyperparameter searches, we did not apply layer-wise learning rate decay, weight decay, or gradient accumulation. We apply a similar strategy to fine-tune \chief. For \gigapath, we follow the recommended recipe of gradient accumulation, weight decay, and layer-wise learning rate decay. Fine-tuning recipe for \prism is not provided.

\begin{figure}[!htp]
\centering
\includegraphics[width=0.85\textwidth]{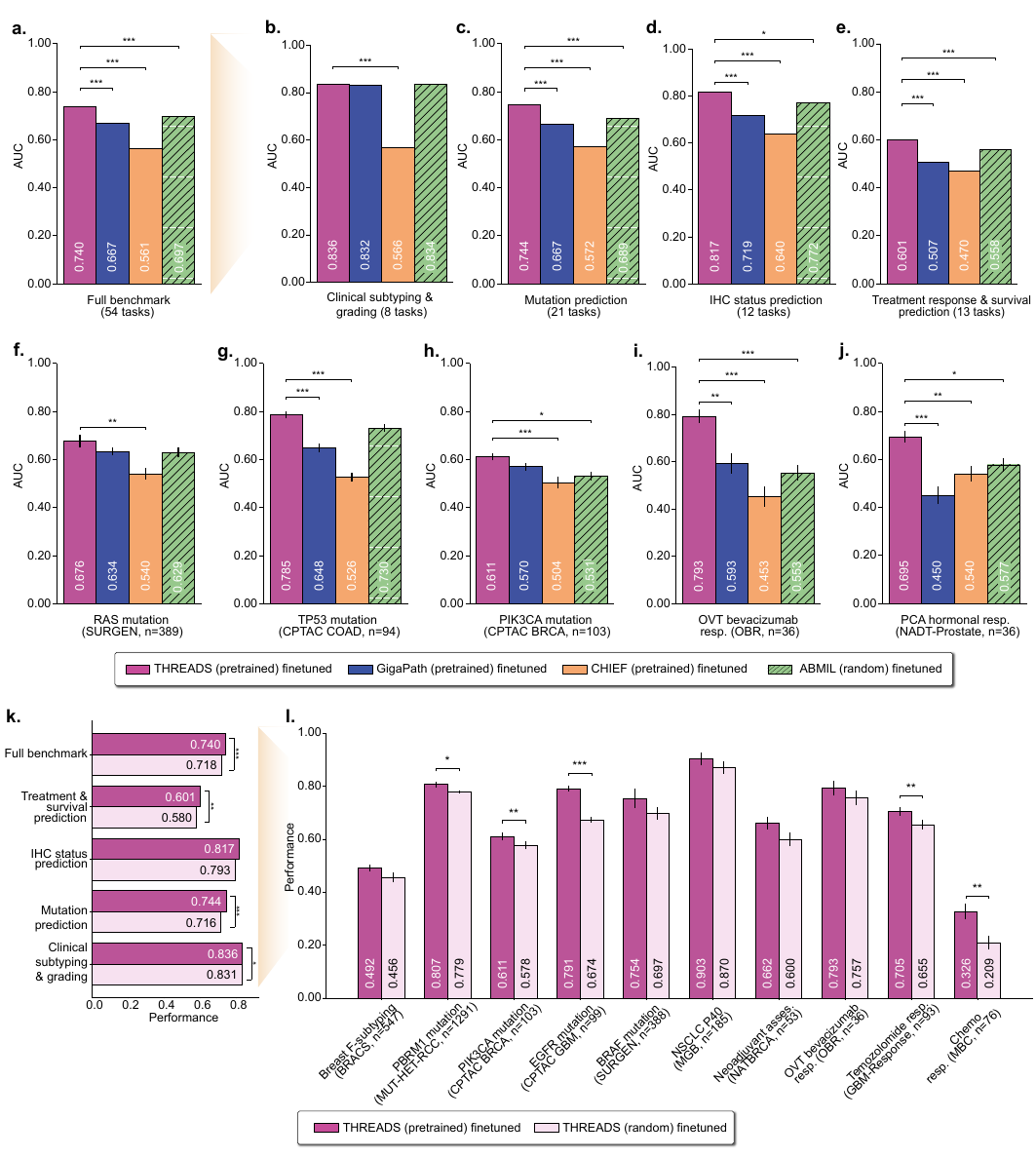}
\caption{\textbf{\ours fine-tuning.}
\textbf{a.} Average performance of \ours and baselines finetuned on 54 benchmarking tasks, along with average performance for each family of tasks: 
\textbf{b.} clinical subtyping and grading (8 tasks), 
\textbf{c.} mutation prediction (21 tasks),
\textbf{d.} IHC status prediction (12 tasks), and 
\textbf{e.} treatment response and survival prediction (13 tasks).
Task-wise comparison of \ours and baselines finetuned on individual tasks:
\textbf{f.} RAS status prediction in SURGEN colorectal adenocarcinoma (COAD).
\textbf{g.} TP53 mutation prediction in CPTAC-COAD. 
\textbf{h.} PIK3CA mutation prediction in CPTAC breast invasive carcinoma (BRCA).
\textbf{i.} Bevacizumab response prediction in ovarian cancer with fine-tuning.
\textbf{j.} Temozolomide response prediction in MGB glioblastoma (GBM). 
\textbf{k.} Comparison of \ours fine-tuning \textit{vs.} training a \ours model from scratch on our benchmark and families of tasks. 
\textbf{l.} Task-wise performance of \ours fine-tuning \textit{vs.} \ours randomly initialized on ten representative tasks. 
Error bars represent standard error, and the centers correspond to the mean computed values of each metric.
Task-wise P-values were determined using two-sided Tukey Honest Significance Difference tests accounting for multiple comparisons following a positive result (P$<$0.05) of a two-way ANOVA.
Statistical significance across multiple tasks (e.g., for each family) was assessed using a mixed-effects model. 
P$<$0.05: *, P$<$0.01: **, P$<$0.001: ***.
 }
\label{fig:fig4}
\end{figure}

\noindent \ours leads to significantly better performance than \chief fine-tuning (absolute gain of 17.9\% and P-value$<$0.001 assessed with mixed-effects statistical modeling) and \gigapath (absolute gain of 7.3\%, P-value$<$0.001) in our 54-task evaluation pipeline. When inspecting individual tasks, \ours leads to significantly better performance than \chief and \gigapath in 54/54 tasks and 40/54 tasks (P-value$<$0.001), respectively (\textbf{Figure \ref{fig:fig4}.a}). In addition, \ours fine-tuning leads to a 4.3\% absolute performance boost over an attention-based MIL baseline trained from scratch (P-value$<$0.001 assessed with mixed-effects modeling). \ours leads to the largest performance gain in challenging tasks characterized by small to medium-size cohorts, such as mutation prediction (absolute gain of 5.5\% over ABMIL and 7.7\% over \gigapath, \textbf{Figure \ref{fig:fig4}.b}), treatment response and survival prediction (gain of 4.3\% over ABMIL and 9.4\% over \gigapath, \textbf{Figure \ref{fig:fig4}.c}) and IHC status prediction (gain of 4.5\% over ABMIL and 9.8\% over \gigapath, \textbf{Figure \ref{fig:fig4}.d}). In clinical subtyping and grading tasks, characterized by larger cohorts, \ours performance is comparable to \gigapath fine-tuning and ABMIL training \textbf{Figure \ref{fig:fig4}.e}. Additional results for specific tasks are provided in \textbf{Figure \ref{fig:fig4}.f,g,h,i,j}. 

\noindent We additionally compared \ours fine-tuning with a randomly initialized \ours model. Overall, fine-tuning leads to an average absolute gain of 2.2\% across all 54 tasks (P-value$<$0.001 assessed with mixed-effects statistical modeling as shown in \textbf{Figure \ref{fig:fig4}.k}). In examining task performance across different families of tasks, we find that fine-tuning yields the most significant improvement, a 2.8\% increase, in mutation prediction tasks, which typically involve challenging tasks and cohorts of small to medium size. Conversely, it shows the smallest improvement, a 0.5\% increase, in clinical subtyping and grading, where the training cohorts tend to be larger and the tasks more objective. Additional results for specific tasks are provided in (\textbf{Figure \ref{fig:fig4}.l}). 

\hheading{Retrieval capabilities of \ours}

\noindent \ours is designed to provide off-the-shelf slide and patient embeddings. This property enables case and patient retrieval without additional model training or fine-tuning. To this end, we extract \ours slide (and patient) embeddings for a collection of samples for which a diagnosis has already been made. We use these samples as a reference database to compare new query cases, which are first embedded using \ours, and then compared to the $k$ most similar embeddings (\textbf{Figure \ref{fig:fig4}.a}). The other three slide encoders (\prism, \gigapath, and \chief) are processed and evaluated in a similar manner. We evaluate retrieval performance using mean Average Precision at $k$ (mAP@$k$), which measures the average number of relevant results within the top $k$ retrieved items, weighted by their rank and averaged over all queries. 

\noindent In rare brain tumor retrieval assessed with the EBRAINS dataset (30 classes, n=2,319 cases), \ours provides the best overall performance for all values of $k$, significantly outperforming all baselines (P-value$<$0.001 for 3/3 baselines at all $k$) (\textbf{Figure \ref{fig:fig4}.b}). We additionally study retrieval performance on the CPTAC consortium data, which aggregates cases from 10 cancer types for a total of 2,115 slides. \ours outperforms all three baselines for $k=1, 5, \text{and } 10$ (P-value$<$0.05 for 3/3 baselines at mAP@1, P-value$<$0.001 for 2/3 baselines at mAP@5, and P-value$<$0.001 for 2/3 baselines at mAP@10) (\textbf{Figure \ref{fig:fig4}.c}). These results highlight how \ours can encode clinically relevant information and retrieve similar cases for comparison and investigation. Additional results are provided in \textbf{Extended Data Table \ref{tab:retrieval-cptac_organ} and \ref{tab:retrieval-ebrains}}.

\hheading{Molecular prompting with \ours}

\begin{figure}[!b]
\centering
\includegraphics[width=0.9\textwidth]{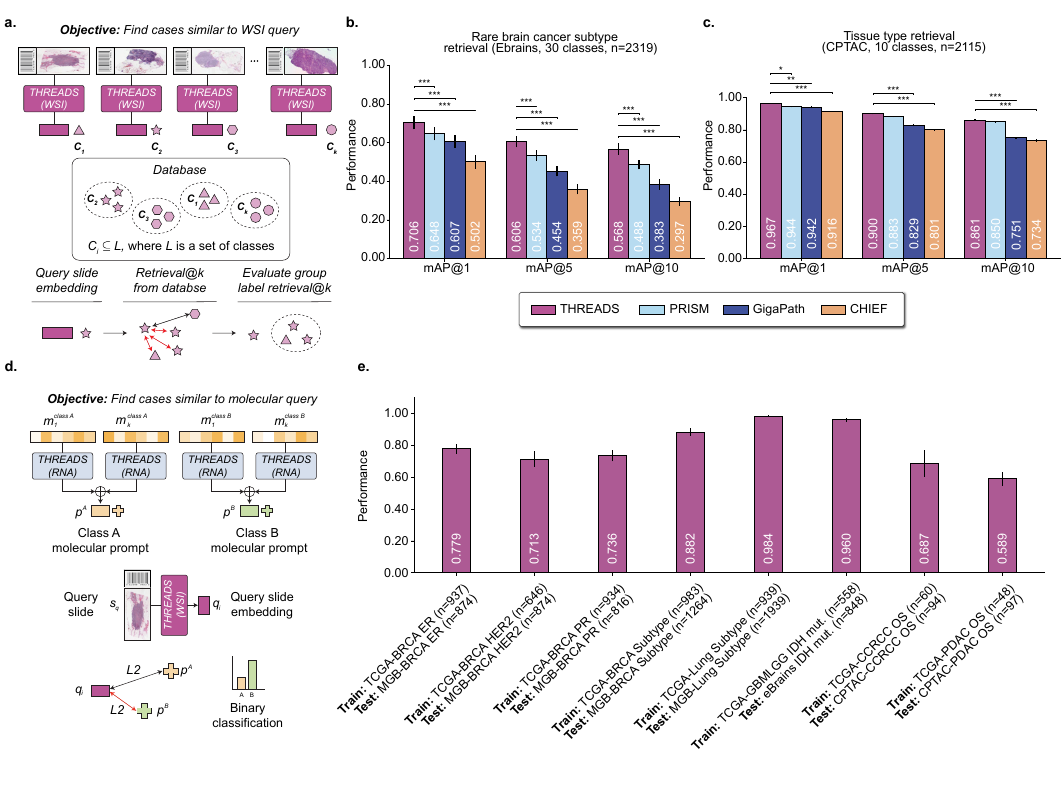}
\caption{\textbf{Retrieval and prompting capabilities of \ours.}
\textbf{a.} Method overview for case retrieval.
\textbf{b.} Rare brain tumor subtype retrieval in EBRAINS (n=30 subtypes) evaluated using mean Average Precision (mAP).
\textbf{c.} Cancer subtype retrieval in CPTAC (n=10 cancer subtypes) evaluated using mAP.
\textbf{d.} Method overview for molecular prompting. 
\textbf{e.} Molecular prompting performance on eight tasks.  
Error bars represent 95\% confidence intervals and the centers correspond to computed values of each metric.
P-values were determined using two-sided Tukey Honest Significance Difference tests accounting for multiple comparisons following a positive result (P$<$0.05) of a two-way ANOVA. P$<$0.05: *, P$<$0.01: **, P$<$0.001: ***.
}
\label{fig:fig5}
\end{figure}

\noindent A hallmark characteristic of multimodal foundation models is to enable transfer and generalization without task supervision. In vision-language models like CLIP~\cite{radford2021learning} and CONCH~\cite{lu2024towards}, such capabilities include zero-shot classification, in which by formulating class labels (\textit{e.g.}, ``Lung Adenocarcinoma'', ``Lung Squamous Cell Carcinoma'') as text prompts via natural language, tasks such as NSCLC subtyping can be performed without requiring training data. In \ours, we introduce a novel multimodal capability known as ``molecular prompting'' (\textbf{Figure \ref{fig:fig5}.d}), in which canonical molecular profiles (\textit{e.g.}, molecular representations of their corresponding disease states) can be leveraged to perform clinical tasks without requiring any task-specific model development. To perform molecular prompting, for each class, we average the representations of molecular profile data from a support dataset (encoded using the molecular branch of \ours) to create class-wise molecular prototypes, which can then be used for cross-modal slide retrieval and classification. At inference time, we classify a query WSI by assigning it to the class of the nearest molecular prompt based on $L2$ distance.

\noindent We evaluated molecular prompting across eight tasks, including clinical cancer subtyping, gene mutation and IHC status prediction, and prognostication (\textbf{Figure \ref{fig:fig5}.e}). Classification with molecular prompts achieves competitive performance across diverse tasks. Building IDH wild-type and mutant prompts from TCGA-GBMLGG and testing on EBRAINS yields a high AUC of 0.960, comparable to linear probing with \ours WSI embedding (0.961). Molecular prompts can also represent typical high- and low-risk profiles, allowing patient survival to be estimated based on similarity to these prompts (additional information is provided in \textbf{Online Methods, Evaluation}). For instance, high- and low-risk prompts generated from TCGA-CCRCC and applied to prognosis prediction on CPTAC-CCRCC achieve a competitive C-index of 0.687. Additional results can be found in \textbf{Extended Data Table \ref{tab:prompting]}}.

\hheading{Insights into \ours.}

\begin{figure}[!btp]
\centering
\includegraphics[width=1.0\textwidth]{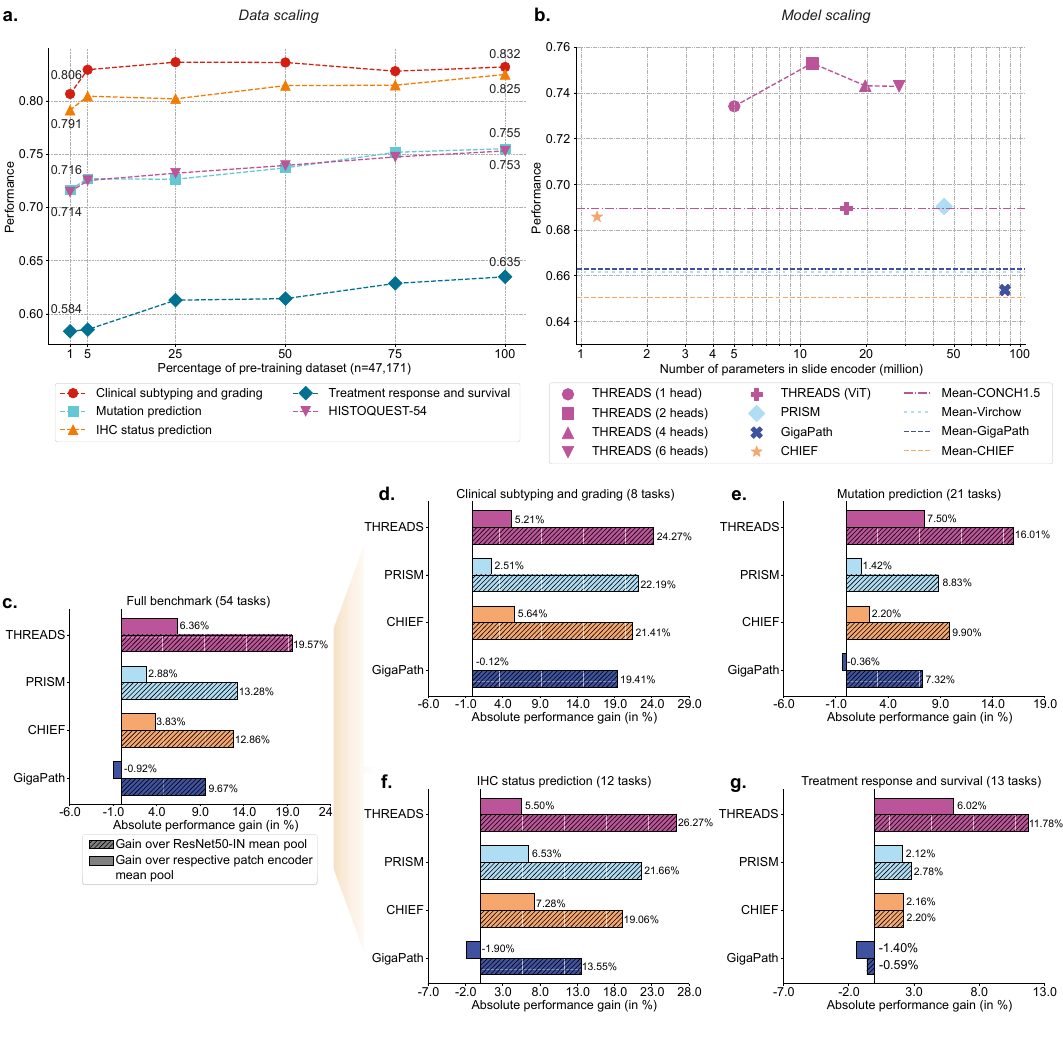}
\caption{\textbf{Properties of \ours.}
\textbf{a.} Data scaling law of \ours across all tasks and families of tasks. Percentage of training slides varies from 1\% to 100\% of \mbtg. 
\textbf{b.} Model scaling law of \ours across all tasks and families of tasks.
\textbf{c.} Absolute performance gain of \ours and slide encoder baselines over the mean pooling baselines from their respective  patch encoder and \resnet. Performance averaged on our 54-task benchmark.  
Absolute performance gain of \ours and slide encoder baselines over the mean pooling baselines from their respective  patch encoder and \resnet for each family of tasks: \textbf{d.} clinical subtyping and grading (8 tasks), \textbf{e.} mutation prediction (21 tasks), \textbf{f.} immunohistochemistry status (IHC) prediction (12 tasks), \textbf{g.} treatment response and survival prediction (13 tasks). \resnet is ResNet50 model pretrained on ImageNet (IN). 
}
\label{fig:fig6}
\end{figure}

\noindent\textbf{Scaling laws.} We additionally study scaling laws in \ours, building on existing works in foundation models that highlighted the benefits of larger training datasets and model sizes\cite{vorontsov2024foundation,chen2024towards}. To this end, we pretrained \ours using subsets of \mbtg of varying sizes. We sampled 1\%, 5\%, 25\%, 50\%, and 75\% of the data from each source, ensuring uniform sampling across major tissue sites. This resulted in the creation of MBTG-1 (473 histomolecular pairs), MBTG-5 (2,356 histomolecular pairs), MBTG-25 (11,791 histomolecular pairs), MBTG-50 (23,584 histomolecular pairs), and MBTG-75 (35,377 histo-molecular pairs). \ours highlights a data scaling law, as shown in \textbf{Figure \ref{fig:fig6}.a}. Across all tasks, we observe a +3.9\% performance increase when using 1\% to 100\% of \mbtg. All families of tasks benefit from data scaling, with treatment response and survival prediction tasks showing the largest performance gain (+5.2\%). When comparing \ours against baselines, we also observe that our approach is more data-efficient than \prism (trained on 195,344 specimens), \gigapath (trained on 171,189 whole-slide images), and \chief (trained on 60,530 whole-slide images). Additional information is provided in \textbf{Extended Data Table \ref{tab:data-scale-linprobe}}. 

\noindent We also assessed model scaling laws by ablating \ours using a varying number of attention heads, resulting in models with a single head (5.0 million parameters in the slide encoder), two heads (proposed approach, 11.3 million parameters), four heads (19.7 million parameters), six heads (28.1 million parameters) and a ViT model with two Transformer layers (16.1 million parameters). We observe that model scaling peaks with a two-head model and then plateaus or leads to decreased performance (-1.0\% when using six vs. two attention heads). A ViT baseline trained with \ours leads to lower performance than our proposed architecture by 6.3\%. Comparing \ours against \prism and \gigapath highlights the parameter-efficiency of \ours. Despite being 4.0$\times$ and 7.5$\times$ smaller than \prism and \gigapath slide encoders, \ours leads to significantly better performance on our benchmark. \chief is lightweight due to its compact architecture but provides significantly lower performance than our single-head model. This analysis highlights that scaling model size in slide encoder does not necessarily lead to better performance and that other factors are more important. Additional information is provided in \textbf{Extended Data Table \ref{tab:model-scale-linprobe}}.

\noindent\textbf{Mean pooling.} To better understand the superior performance of \ours over baselines, we conducted additional ablations. First, we compared the quality of \ours patch embeddings (based on \conch) against \gigapath patch encoder, \prism patch encoder (based on \virchow), and \chief patch encoder (based on \ctranspath). To this end, we adopt mean pooling to derive a slide embedding, which we then use for linear probing classification. \conch with mean pooling reaches an average of 68.9\% across all tasks outperforming \virchow, \gigapath and \chief by 2.7\%, 2.6\% and 3.8\%, respectively (\textbf{Figure \ref{fig:fig6}.a} and \textbf{Extended Data Table \ref{tab:data-scale-linprobe}}). We hypothesize that this gain stems from (i) vision-language fine-tuning in \conch, whereas \virchow, \gigapath, and \ctranspath are vision-only models, and (ii) from extracting patch features on larger regions (512$\times$512-pixel regions vs. 256$\times$256-pixel in baselines) which can better capture morphological context. We also note that \conch is a ViT-Large model (307 million vision parameters), whereas \prism uses a ViT-H (632 million parameters, 2.0$\times$ more than \conch), and \gigapath used a ViT-G (1.13 billion parameters, 3.7$\times$ more than \conch), highlighting the parameter-efficiency of our pipeline.

We additionally compare slide encoders with their respective mean pooling baselines, i.e., \conch mean pooling and \ours, \gigapath mean pooling and \gigapath, \ctranspath mean pooling and \chief, and \virchow mean pooling and \prism (\textbf{Figure \ref{fig:fig6}.c}). Using linear probing evaluation across all tasks, \ours leads to a gain of 6.36\% over \conch mean pooling (P-value$<$0.001) and 19.3\% over \resnet mean pooling (P-value$<$0.001). In contrast, \gigapath slide encoder leads to lower performance than mean pooling (-0.83\%). Both \prism and \chief lead to a performance gain over their respective mean pooling (2.86\% in \prism and 4.40\% in \chief), lower than \ours performance gain. This observation highlights the complexity of whole-slide representation learning in capturing information-dense off-the-shelf embeddings. 

\noindent\textbf{Clustering in \ours embedding space.} To study the superior performance of \ours in predicting clinically and biologically relevant information from whole slide images, we explored the clustering capabilities of the latent space compared to baselines. To this end, we embedded all slides from the ten CPTAC cohorts (n=2,115 WSIs) using \ours and baselines. We then applied K-means clustering, where the number of clusters was set to the number of cancer types (n=10). From there, we computed two clustering metrics: the adjusted Rand index (ARI) and mutual information (MI) (\textbf{Extended Data Figure \ref{fig:edf2}}). \ours highlights better clustering capabilities than \prism, \gigapath, and \chief (for instance ARI=0.654 for \ours \emph{vs.} 0.354 for \chief). In addition, we derived tSNE visualization of the latent space, which further highlights the separability of \ours compared to baselines. We conducted a similar analysis with EBRAINS for fine brain tumor subtyping (30 classes). We observe similar trends with \ours highlighting better clustering and linear separability than baselines (for instance, MI=2.104 for \ours \emph{vs.} 1.44 for \gigapath).  


\Heading{Discussion}

\noindent In this study, we introduced \ours, a foundation model for pathology that can provide biologically and clinically relevant representations of H\&E-stained whole-slide images. \ours uses a novel multimodal pretraining strategy, where the learned slide representation is guided by its corresponding molecular profile. Using this strategy, the resulting slide representations can capture morphological features reflective of the underlying molecular composition of the tissue. \ours was thoroughly tested on a wide benchmark of 54 tasks, covering four families of tasks: clinical cancer subtyping and grading, gene mutation prediction, immunohistochemistry status prediction, and treatment response and survival prediction. \ours consistently shows state-of-the-art performance under several evaluation scenarios, including in- and out-of-domain generalization, few-shot learning, and case retrieval. Importantly, \ours can reach clinical-grade performance on subtyping and grading tasks using simple linear models built upon our slide embeddings. \ours also highlights great potential for patient outcome prediction and can help identify patients who will respond to certain treatments.

\ours sets apart from existing methods using its unique pretraining strategy based on multimodal alignment with molecular profiles. Unlike \prism\cite{shaikovski2024prism}, which relies on matching pathology reports, molecular data provide an unbiased, objective perspective on cellular and tissue states, free from the subjective influences inherent in written reports. On the other hand, \chief\cite{wang2024chief} and \gigapath\cite{xu2024whole} employ weaker pretraining signals, relying on contrastive alignment with tissue sites and masked autoencoding, respectively. We hypothesize that these approaches lack the capacity to capture the subtle morphological features essential for addressing most clinical tasks. Our investigation into scaling laws of \ours further reveals that the saturation point of model and data scale remains an open question in slide representation learning. We found that simpler clinical tasks, such as cancer grading, do not benefit significantly from larger pretraining datasets. However, more challenging tasks--particularly those involving treatment response prediction and molecular predictions--show substantial performance gains when models are trained on larger and more diverse datasets. This indicates that data diversity and pretraining strategy are critical factors influencing the efficacy of the resulting model. \ours is significantly smaller than existing models, being 7.5 $\times$ smaller than GigaPath and 4.0 $\times$ smaller than \prism. This suggests that in slide representation learning, simply scaling model size may not be the most influencing factor for building general-purpose models. In addition, \ours is trained on a highly diverse dataset that includes 39 main human tissue sites following the highest level of the OncoTree cancer classification system. In contrast, \gigapath, \chief, and \prism are trained on less diverse tissue sites, often with a skewed distribution toward skin, breast, and lung cases. We hypothesize that the broader diversity in \ours likely contributes to its enhanced generalizability and robustness across a wide range of tasks and tissue types.

Despite these advancements, certain limitations remain. Although \ours was pretrained on an unprecedented cohort of over 47,000 histomolecular pairs, it cannot encompass the full spectrum of molecular and morphological heterogeneity. As next-generation sequencing becomes more widely deployed in clinical settings, the potential to scale \ours’ pretraining dataset by orders of magnitude may reshape this landscape, potentially uncovering new scaling laws that are currently beyond reach with our existing cohorts. Additionally, extending our molecular-guided approach to include other molecular assays, such as immunohistochemistry and special stains, could broaden the scope of \ours\cite{jaume2024multistain}. In addition, \ours architecture uses a multihead attention-based model, which treats patch embeddings independently. Attempts to replace our backbone with a Vision Transformer model fail to match the performance of simpler models, even the ones with a single attention head. The use of larger image patches (512 $\times$ 512 pixels) instead of the typical 256 $\times$ 256 pixels in most patch encoders may reduce the need for explicit context modeling. Alternatively, slide encoders based on Transformers may need a larger pretraining cohort size for significant performance improvements. 

\ours has the potential to impact various aspects of computational pathology and oncology. First, it can bring off-the-shelf integration in data-scarce scenarios. \ours can be readily used to prototype new tasks and assess predictive performance at a minimal cost. In addition, the reduced training data requirements enable the development of clinical-grade predictive systems for rare diseases. Researchers and clinicians can also utilize \ours pretrained weights as initialization for additional fine-tuning on specific tasks. This transfer learning approach can accelerate model development and can improve performance on specialized tasks, such as rare molecular alteration classification or treatment response and resistance prediction. Finally, the case retrieval capabilities of \ours make it well-suited for identifying rare conditions in clinical settings. Overall, our study highlights the rich biological information contained in molecular assays, which can be transferred to slide encoders to advance the development of diagnostic and prognostic tools. Future work will focus on scaling the pretraining dataset size and increasing the biological richness of the training signal by including additional modalities.

\Heading{Online Methods}

\noindent The Mass General Brigham (MGB) institutional review board approved the retrospective analysis of pathology slides (whole-slide images or WSIs), corresponding next-generation sequencing (NGS) assays, and corresponding reports used in this study. Research conducted in this study involved a retrospective analysis of pathology slides and NGS assays, and the participants were not directly involved or recruited for the study. The requirement for informed consent to analyze archival pathology slides and NGS assays was waived. Before scanning and digitization, all pathology slides were de-identified to ensure anonymity. The sample sizes were determined by the availability of the data.

\clearpage
\heading{Pretraining dataset curation} 

\noindent We present \ours pretraining dataset, \mbtg, a large and diverse dataset composed of paired formalin-fixed paraffin-embedded (FFPE) haematoxylin and eosin (H\&E) whole-slide images (WSIs), tissue bulk RNA expression, and DNA variant data including single nucleotide variations (SNV), insertions and deletions (indels), and copy number variations (CNV). Sourced from the Massachusetts General Hospital (MGH), the Brigham and Women's Hospital (BWH), the Genotype-Tissue Expression (GTEx) consortium, and The Cancer Genome Atlas (TCGA), the dataset comprises 47,171 WSIs from 39 major organs, totaling 125,148,770 512 $\times$ 512-pixel histology images tiled at $20\times$ magnification. 26,615 WSIs have associated RNA expression data, and 20,556 WSIs have associated DNA variant data. The total size of the dataset is 40.7 TB. We describe each data source contributing to the \mbtg dataset. Additional information is provided in \textbf{Extended Data Table \ref{tab:pretrain-counts}}. 

\noindent\textbf{MGH.} Anchored multiplex PCR\cite{zheng2014anchored} and next-generation sequencing (NGS) of total nucleic acid were applied to generate bulk RNA expression data using FusionPlex (Integrated DNA Technologies, Coralville, IA). The Solid Fusion Assay V2 is a clinically validated pan-cancer RNA assay that targets canonical exons involved in fusion variants of 59 genes and control genes. The assay generates predominantly RNA reads to detect gene fusions, splicing, exon-skipping events, and gene expression. In this study, we focused on bulk gene expression data. Transcript abundance for each gene was quantified using the Kallisto 0.50.1 pseudo alignment software, with an index built from GENCODE Human Release v45 (genome assembly GRCh38.p14). Measurements from different isoforms were summed to acquire a single total per gene. RNA expression was summarized as transcripts per million (TPM) and further normalized by taking the $\text{log}_{2}$ of TPM across 54 genes. No additional batch effect normalization techniques were applied. The associated H\&E glass slides were scanned using an Aperio GT450 scanner at $40\times$. In total, 6,899 FFPE H\&E WSI and bulk RNA expression pairs from 25 tissue sites were utilized. The final MGH dataset amounted to 11.0 TB.

\noindent\textbf{BWH.} OncoPanel is an Agilent SureSelect hybrid capture targeted DNA NGS assay designed to detect SNVs, indels, CNVs, and some structural variations. Library preparation, sequencing, and bioinformatics analysis have been previously described\cite{garcia2017validation}. Similar to the MGH Solid Fusion Assay V2, OncoPanel testing was performed according to a routine clinical workflow with expert molecular pathologist review. For \ours, we sourced SNV, CNV, and indel data associated with 20,556 FFPE H\&E slides from 32 tissue sites. We subsetted the data to 239 common genes across the OncoPanel versions used in this study (ranging from 2012 to 2020). CNVs were categorized into four groups: two-copy deletion, loss, gain, and amplification. SNVs and indels were categorized into three groups: small coding change, large coding change, and non-coding change. The mutation status of each gene was multi-hot encoded and concatenated together to form a vector of length 7; therefore, across the 239 genes, the variant status vector has a length of 1,673. The H\&E WSIs associated with the OncoPanel testing were scanned using an Aperio GT450 scanner at 40$\times$ magnification. The final BWH dataset amounted to 12.0 TB.

\noindent\textbf{TCGA.} The Cancer Genome Atlas (TCGA) contributed 10,209 FFPE H\&E WSIs from 32 cancer types, paired with bulk RNA expression. TCGA includes histology data from over 11,000 cancer patients, with tissue samples scanned at 40$\times$ and 20$\times$ magnification using Aperio and Hamamatsu scanners. The expression data comprises bulk whole-transcriptomic RNA sequencing analysis from approximately 20,000 samples across 33 cancer types, with a sequencing depth of 50-200 million reads per sample, using Illumina HiSeq platforms. Exclusion criteria for WSIs were frozen tissue, benign and non-diagnostic, missing appropriate magnification information and metadata necessary for processing, and lacking associated RNA expression. While TCGA provides whole-transcriptome sequencing, we selected a set of 4,917 cancer-related genes\cite{jaume2024modeling}, which was reduced to 4,848 genes present in the vocabulary of our transcriptomic encoder (scGPT encoder\cite{cui2024scgpt}) used to generate molecular embeddings (\textbf{Online Methods, Model Design and Development}). RNA expression was measured in transcripts per million (TPM) and further normalized by taking the $\text{log}_{2}$ of TPM. No additional batch effect normalization techniques were applied. The final TCGA dataset amounted to 10.5 TB.

\noindent\textbf{GTEx.} The Genotype-Tissue Expression (GTEx)\cite{gtex2015genotype} contributed 9,507 FFPE H\&E WSIs and bulk RNA expression pairs from 29 tissue sites to \mbtg. GTEx includes FFPE WSIs from 24,782 non-cancerous samples scanned at 40$\times$ magnification using either Aperio AT2 or Hamamatsu NanoZoomer-XR. The expression data comprises whole-transcriptome bulk RNA sequencing from 17,382 samples, with a depth of 50-100 million reads per sample, using Illumina HiSeq 2000/2500 platforms. Exclusion criteria for WSIs include missing appropriate magnification information and metadata necessary for processing and lacking associated RNA expression. Even though GTEx provides whole-transcriptome sequencing, we selected a set of 5,000 genes showing maximum variation (as measured by the standard deviation of $\text{log}_{2}$ of TPM) across all organs. This gene set was reduced to 4,932 genes found in the vocabulary of our transcriptomic encoder (scGPT\cite{cui2024scgpt}). RNA expression was measured in transcripts per million (TPM) and further normalized by taking the $\text{log}_{2}$ of TPM. No additional batch effect normalization techniques were applied. The final GTEx dataset amounted to 7.2 TB.

\heading{Downstream tasks and datasets}

\noindent We provide a description of each task and cohort in our benchmark, which includes 54 tasks from 23 datasets across nine major organs. Our benchmark covers six types of tasks: morphological tumor subtyping (\textbf{Extended Data Table \ref{tab:task_summary.morphological_subtyping}, \ref{tab:label_counts_mgb_brca_subtype}, \ref{tab:label_counts_mgb_lung_subtype}, \ref{tab:label_counts_bracs_slidelevel_coarse},
\ref{tab:label_counts_bracs_slidelevel_fine},
\ref{tab:label_counts_ebrains_diagnosis_group},
\ref{tab:label_counts_ebrains_diagnosis}}), tumor grading (\textbf{Extended Data Table \ref{tab:task_summary.tumor_grading}, \ref{tab:label_counts_imp_grade}, \ref{tab:label_counts_panda_isup_grade}}), immunohistochemistry status prediction (\textbf{Extended Data Table \ref{tab:task_summary.molecular_subtyping}}), prediction of gene-level mutations (\textbf{Extended Data Table \ref{tab:task_summary.mutation_prediction}}), treatment response prediction (\textbf{Extended Data Table \ref{tab:task_summary.treatment_response}}), and survival prediction (\textbf{Extended Data Table \ref{tab:task_summary.survival_prediction}}).

\noindent\textbf{MGB-Breast.} We used an internal cohort of invasive breast cancer (BRCA) for morphological and immunohistochemistry status prediction\cite{vaidya2024demographic,jaume2024multistain}. MGB-Breast comprises 1,264 WSIs (mix of biopsies and resections) scanned from Brigham and Women's Hospital (one WSI per patient). We curated one morphological subtyping task (\textbf{Extended Data Table \ref{tab:label_counts_mgb_brca_subtype}}) and three immunohistochemistry (IHC) status prediction tasks: estrogen receptor (ER) status prediction, progesterone receptor (PR) status prediction, and human epidermal growth factor receptor 2 (HER2) status prediction. ER, PR, and HER2 status were manually extracted from pathology reports. Additional information is provided in \textbf{Extended Data Table \ref{tab:task_summary.morphological_subtyping}} and \textbf{Extended Data Table \ref{tab:task_summary.molecular_subtyping}}.

\noindent\textbf{MGB-Lung.} We used an internal cohort of lung cancer cases for morphological and IHC status prediction\cite{vaidya2024demographic,jaume2024multistain}. MGB-Lung comprises 1,939 WSIs scanned from Brigham and Women's Hospital (one WSI per patient). We curated one morphological subtyping task  (\textbf{Extended Data Table \ref{tab:label_counts_mgb_lung_subtype}}) and six immunohistochemistry tasks: (1) thyroid transcription factor-1 (TTF-1) status prediction, (2) protein 40 (P40) status prediction, (3) protein 63 (P63) status prediction, (4) Napsin A status prediction, (5) caudal type homeobox 2 (CDX2) status prediction, and (6) cytokeratin 5 and 6 (CK5-6) status prediction. Additional information is provided in \textbf{Extended Data Table \ref{tab:task_summary.morphological_subtyping}} and \textbf{Extended Data Table \ref{tab:task_summary.molecular_subtyping}}.

\noindent\textbf{BCNB.} We used the public BCNB dataset\cite{xu2021predicting} (Breast Cancer Core-Needle Biopsy) for IHC status prediction in breast cancer. BCNB comprises 1,058 WSIs (one WSI per patient) which we use for ER status prediction, PR status prediction, and HER2 status prediction. Additional information is provided in \textbf{Extended Data Table \ref{tab:task_summary.morphological_subtyping}}.

\noindent\textbf{MUT-HET-RCC.} We used the MUT-HET-RCC dataset\cite{acosta2022intratumoral} for mutation prediction in renal cell carcinoma. MUT-HET-RCC comprises 1,291 WSIs (one WSI per patient) which we use for (1) BAP1 mutation prediction, (2) PBRM1 mutation prediction, and (3) SETD2 mutation prediction. Additional information is provided in \textbf{Extended Data Table \ref{tab:task_summary.mutation_prediction}}.

\noindent\textbf{IMP.} We used the public IMP-CRS 2024 dataset (IMP)\cite{neto2024interpretable} for colorectal cancer grading. IMP consists of 5,333 WSIs collected from colorectal biopsies and polypectomies. IMP is used for 3-class tumor grading into non-neoplastic lesions, low-grade lesions (adenomas with low-grade dysplasia), and high-grade lesions (adenomas with high-grade dysplasia and invasion) (see \textbf{Extended Data Table \ref{tab:label_counts_imp_grade}}). Additional information is provided in \textbf{Extended Data Table \ref{tab:task_summary.tumor_grading}}.

\noindent\textbf{Prostate cANcer graDe Assessment (PANDA).} We used the public PANDA data for prostate cancer grading (ISUP grading)\cite{bulten2022artificial}. PANDA comprises 10,616 core needle biopsies from Radboud University Medical Center and Karolinska Institute, each annotated with an ISUP grade (6-class classification task). We follow prior work\cite{chen2024towards,pati2022hierarchical} in excluding slides with equivocal labels (\url{https://www.kaggle.com/competitions/prostate-cancer-grade-assessment/discussion/169230}), which resulted in 9,555 slides with the label breakdown shown in \textbf{Extended Data Table \ref{tab:label_counts_panda_isup_grade}}. We used the train split (7,647 WSIs) and test split (954 WSIs) from \gigapath \cite{xu2024whole}, and we did not use their validation split (954 WSIs). Additional information is provided in \textbf{Extended Data Table \ref{tab:task_summary.tumor_grading}}.

\noindent\textbf{Clinical Proteomic Tumor Analysis Consortium (CPTAC).} We used the public CPTAC data for pan-cancer mutation prediction\cite{edwards2015cptac,thangudu2020cptac}. Specifically, we included (1) CPTAC-BRCA (invasive breast cancer) for PIK3CA and TP53 mutation prediction,  (2) CPTAC-CCRCC (clear-cell renal-cell carcinoma) for BAP1 and PBRM1 mutation prediction, (3) CPTAC-COAD (colon adenocarcinoma) for KRAS and TP53 mutation prediction, (4) CPTAC-GBM (glioblastoma) for EGFR and TP53 mutation prediction, (5) CPTAC-HNSC (head and neck squamous cell carcinoma) for CASP8 mutation prediction, (6) CPTAC-LSCC (squamous cell lung carcinoma) for KEAP1 and ARID1A mutation prediction, (7) CPTAC-LUAD (lung adenocarcinoma) for EGFR, STK11 and TP53 mutation prediction, and (8) CPTAC-PDAC (pancreatic ductal adenocarcinoma) for SMAD4 mutation prediction. We also used overall survival data for CPTAC-CCRCC, CPTAC-PDAC, CPTAC-LUAD, and CPTAC-HNSC\cite{liao2023proteogenomics}. Additional information is provided in \textbf{Extended Data Table \ref{tab:task_summary.mutation_prediction}}.

\noindent\textbf{BReAst Carcinoma Subtyping (BRACS).} We used the public BRACS dataset\cite{brancati2021bracs} for coarse- and fine-grained breast morphological subtyping. BRACS consists of 547 breast carcinoma WSIs from 189 patients sourced from IRCCS Fondazione Pascale. Each WSI is used for coarse-grained (\textbf{Extended Data Table \ref{tab:label_counts_bracs_slidelevel_coarse}}) and fine-grained (\textbf{Extended Data Table \ref{tab:label_counts_bracs_slidelevel_fine}}) morphological subtyping. Due to the limited size of the official test set, we redefined train-test splits with an 80:20 ratio. Because BRACS-Fine and BRACS-Coarse are slide-level prediction tasks with multiple slides per case, we kept all slides belonging to the same patient together, ensuring one patient does not end up in both train and test. Therefore, this dataset was not explicitly label-stratified (as each patient would have multiple labels). Additional information is provided in \textbf{Extended Data Table \ref{tab:task_summary.morphological_subtyping}}.

\noindent\textbf{EBRAINS.} We used the EBRAINS dataset\cite{roetzer2022ebrains} for coarse- and fine-grained brain tumor subtyping. EBRAINS consists of 3,114 WSIs acquired by the EBRAINS Digital Tumor Atlas at the University of Vienna. We reused splits from UNI\cite{chen2024towards}, which kept categories with at least 30 samples, resulting in 2,319 slides. Each WSI is used for coarse-grained (\textbf{Extended Data Table \ref{tab:label_counts_ebrains_diagnosis_group}}) and fine-grained (\textbf{Extended Data Table \ref{tab:label_counts_ebrains_diagnosis}}) morphological subtyping. Splits are stratified by patients to ensure slides from a patient are not found in both train and test splits. Additional information is provided in \textbf{Extended Data Table \ref{tab:task_summary.morphological_subtyping}}.

\noindent\textbf{OV-Bevacizumab.} We used the OV-Bevacizumab dataset\cite{wang2022weakly} for treatment response prediction in ovarian cancer. OV-Bevacizumab consists of 288 WSIs from 78 patients. Non-responders are defined as having a measurable regrowth of the tumor or as a serum CA-125 concentration of more than twice the value of the upper limit of normal during the treatment course for the bevacizumab therapy. We kept all patients who received bevacizumab as their first-line treatment and additionally removed four cases (case IDs: P00181938C, 2630938, 2224393, 2937351), which were labeled as both responders and non-responders,
yielding 85 WSIs from 36 patients. Additional information is provided in \textbf{Extended Data Table \ref{tab:task_summary.treatment_response}}.

\noindent\textbf{NADT-Prostate.} We used the neoadjuvant androgen deprivation therapy (NADT)-Prostate dataset\cite{wilkinson2021nascent} for hormonal therapy response prediction in prostate adenocarcinoma. Baseline tumor volumes were estimated using multiparametric magnetic resonance imaging (mpMRI). After 6 months of NADT combined with enzalutamide, patients underwent a second mpMRI before radical prostatectomy (RP). The final pathologic response to treatment was defined by a residual cancer burden of 0.05 cubic centimeters, distinguishing responders from non-responders. While the entire dataset consists of 1,401 WSIs with various stains, we only used the H\&E stained WSIs, yielding 449 WSIs from 36 patients (20$\times$ magnification). Additional information is provided in \textbf{Extended Data Table \ref{tab:task_summary.treatment_response}}.

\noindent\textbf{Treatment response in glioblastoma (GBM-Treatment).} We collected an internal cohort of 93 glioblastoma patients, accounting for 347 H\&E-stained slides, who received radiotherapy and temozolomide\cite{touat2020mechanisms, ayoub2024path}. Based on patient survival in months following treatment initiation (all patients deceased) and a cutoff of 15 months \cite{domingo2015next}, we stratified the patients into responders and non-responders. Additional information is provided in \textbf{Extended Data Table \ref{tab:task_summary.treatment_response}}.

\noindent\textbf{Post-NAT-BRCA.} We used the post-neoadjuvant therapy (NAT) breast invasive carcinoma (Post-NAT-BRCA) dataset\cite{martel2019assessment} to assess the presence of lymphovascular invasion in post-NAT WSIs. The dataset contains 53 H\&E-stained WSIs from 50 patients (20$\times$ magnification). Additional information is provided in \textbf{Extended Data Table \ref{tab:task_summary.treatment_response}}.

\noindent\textbf{SURGEN.} We used public cases from the SR386 cohort of SurGen\cite{myles2024leveraging}, which includes 389 patients with colon and rectum adenocarcinoma. For each patient, we predict mismatch repair (MMR) loss, BRAF mutation, KRAS mutation, 5-year death, and overall survival. Treatment information is available only for a subset of patients. Additional information is provided in \textbf{Extended Data Table \ref{tab:task_summary.survival_prediction} and \ref{tab:task_summary.mutation_prediction}}.

\noindent\textbf{MBC.} We used the public Bergstrom dataset\cite{galland2022mbc, bergstrom2024deep} from which we retrieved 77 metastatic breast cancer patients (MBC) with corresponding H\&E WSIs (n=99 WSIs, 1 to 2 WSI per patient). All 77 patients were treated with platinum, with a subset of 54 who were additionally treated with taxane. We predict Response Evaluation Criteria in Solid Tumors (RECIST1.1) and overall survival. Since all patients in MBC received the same treatment, predicting survival can be considered as predicting response to the treatment given. Additional information is provided in \textbf{Extended Data Table \ref{tab:task_summary.survival_prediction} and \ref{tab:task_summary.mutation_prediction}}.

\noindent\textbf{BOEHMK.} We used the public BOEHMK\cite{boehm2022multimodal} dataset comprising 183 patients for which we could retrieve the H\&E WSI and corresponding metadata, including overall and progression free survival. Patients were diagnosed with high-grade serous ovarian cancer and treated with neoadjuvant chemotherapy followed by interval debulking surgery, or underwent primary debulking surgery. Since all patients in BOEHMK received the same treatment, predicting progression free survival can be considered as predicting response to the treatment given. Additional information is provided in \textbf{Extended Data Table \ref{tab:task_summary.treatment_response} and \ref{tab:task_summary.survival_prediction}}. 

\noindent\textbf{TCGA (generalizability).} \textbf{TCGA-GBMLGG} consists of 1,123 WSIs from 558 patients with glioblastomas multiforme (GBM) and lower-grade gliomas (LGG). The WSIs are classified into two classes: isocitrate dehydrogenase (IDH) mutation (425 WSIs) and no IDH mutation (698 WSIs). \textbf{EBRAINS} serves as an external cohort for this task (IDH MUT: 333 WSIs, IDH WT 540 WSIs). \textbf{TCGA-BRCA} (invasive breast carcinoma) consists of 1,048 WSIs from 983 patients. The WSIs are classified into two cancer subtypes: invasive ductal carcinoma (IDC) (838 WSIs) and invasive ductal carcinoma (ILC) (210 WSIs). MGB-Breast subtyping serves as external cohort for this task. \textbf{TGCA-BRCA} is also used for IHC status prediction: ER (996 WSIs total, 78.3\% WSIs positive), PR (993 WSIs total, 68.2\% WSIs positive), HER2 (692 WSIs total, 22.8\% WSIs positive)\cite{jaume2024multistain}. \textbf{BCNB} serves as external cohort IHC status prediction. \textbf{TCGA-Lung} (lung cancer) consists of 1,043 WSIs from 946 patients with non-small cell lung cancer (NSCLC). The WSIs are classified into two classes: lung adenocarcinoma (LUAD, 531 slides) and lung squamous cell carcinoma (LUSC, 512 slides). MGB-Lung subtyping serves as the external cohort for this task. We define \textbf{TCGA-LUAD} as only the adenocarcinomas and use this dataset for overall survival prediction (509 WSIs from 446 patients, 60.1\% censored). \textbf{CPTAC-LUAD} is used as the external test cohort for this task. \textbf{TCGA-PDAC} (pancreatic ductal adenocarcinoma) consists of 180 WSIs from 166 patients with PDAC. We use overall survival labels (47.2\% censored) while testing on \textbf{CPTAC-PDAC} for external validation.

\subsection{Data splits.} We created two types of splits: $k=\text{All}$ splits, which distribute all available samples between train and test, and fewshot splits, which restrict the size of the training set to only a few examples (``shots''). For certain datasets (EBRAINS, PANDA, IMP), we use ``official'' single-fold $k=\text{All}$ splits that have been publicly released. Otherwise, we create 80:20 train:test splits using 5-fold cross-validation or 50-fold bootstrapping. We also create fewshot splits with $k \in \{1, 2, 4, 8, 16, 32\}$ examples per class. For fewshot splits, if there is more than one $k=\text{All}$ fold, then corresponding fewshot splits are created by sampling $k$ items from the training set of each $k=\text{All}$ fold, and masking the remainder. Otherwise, bootstrapped fewshot splits are created by repeatedly sampling $k$ items from the single training fold. Note that the test set of all splits for each task is the same. For certain tasks with classes containing too few labeled examples, we omit $k=32$ fewshot splits.


\heading{Model design and development}
\label{sec:supp-modeldesign}

\noindent Each WSI goes through three steps: (1) tissue detection and patching, (2) feature extraction from each patch, and (3) slide encoding using \ours. 

\noindent\textbf{Tissue segmentation and patching.} Each slide is tiled into fixed-size image patches and processed using a pretrained vision model to extract patch-level feature embeddings. For compute efficiency, we only process patches overlapping with tissue and ignore background regions. Background \textit{vs.} tissue segmentation is performed using a deep learning feature pyramidal network (FPN) fine-tuned from the \texttt{segmentation-} \texttt{models-pytorch} package\cite{lin2017feature} on an in-house dataset of mask annotations. Non-overlapping 512$\times$512-pixel patches are extracted at 20$\times$ magnification ($\sim 0.5$ \si{\um}/px) for each slide.

\noindent\textbf{Patch encoder.} We use the \conch patch encoder, the next iteration of CONCH\cite{lu2024towards}. \conch was trained by initializing UNI weights\cite{chen2024towards} followed by full multimodal fine-tuning using image captions. UNI is a state-of-the-art vision-only foundation model for pathology trained on 100 million image patches of size 224$\times$224 pixels using a Vision Transformer Large (ViT-L)\cite{dosovitskiy2021image}. Fine-tuning was conducted with 1.17 million vision-language pairs (pathology image/caption pairs) using CoCa\cite{yu2022coca} on 448$\times$448-pixel patches, as described in \cite{lu2024towards,lu2024pathchat}. 512$\times$512-pixel patches were resized to 448$\times$448, and normalized using default ImageNet mean and standard deviation parameters before being passed to \conch. An overview of \conch training hyperparameters is provided in \textbf{Extended Data Table~\ref{tab:conch_hyperparams}}.

\noindent\textbf{Slide encoder.} \ours consists of an attention-based multiple instance learning (ABMIL) model \cite{ilse2018attention,lu2021data} with single or multiple attention heads, depending on the configuration. For the single-headed configuration, raw patch embeddings are projected from 768-dimensional \conch features to 1024-dimensional features $\mX$ using a pre-attention network with three hidden layers, layer normalization, \gelu activation, and 0.1 dropout. The attention head processes batched input patch features $\mX \in \R^{N\times 1024}$, where $N$ is the number of patches. The gated attention mechanism comprises three fully connected layers: two parallel layers ($a$ and $b$) with a hidden dimension of 1024 and 25\% dropout, followed by a final layer ($c$). The attention weights $\valpha$ are computed as:
\begin{equation}
\valpha = c(\tanh(a\mX) \odot \sigma(b\mX))
\end{equation}
where $\odot$ denotes element-wise multiplication, $\tanh$ and $\sigma$ represent the hyperbolic tangent and sigmoid functions, respectively. The final slide-level features $s$ are computed by multiplying the softmax-normalized attention scores by the patch features:
\begin{equation}
\vs = \softmax(\valpha)^\top \mX
\label{eq:softmax_attention}
\end{equation}

\noindent In the multi-headed configuration, the pre-attention network includes a third hidden layer (\gelu activation, 0.1 dropout), which projects from hidden dimension 1024 to $M \times 1024$, where $M$ is the number of heads. The output of this layer is chunked into $M$ feature vectors, each processed separately by its corresponding attention head. The aggregated slide-level features from each head are concatenated and projected with a post-attention linear layer $L: \R^{(M \times 1024)} \to \R^{1024}$ to derive the final 1024-dimensional slide embedding.

\noindent\textbf{Gene expression encoder.} We encode gene expression data using a modified scGPT model\cite{cui2024scgpt}. scGPT is a single-cell foundation model based on the Transformer architecture\cite{vaswani2017attention}. While originally developed for single-cell gene expression encoding, we adapted it to operate on bulk RNA expression data\cite{wang2024path}. Expression data preprocessing was performed for each data source as described in \textbf{Online Methods, Pretraining dataset curation}.
scGPT consists of three encoders: a gene-identity encoder $G$, an expression-value encoder $E$, and a transformer encoder $T$. $G$ is a lookup table of learned 512-dimensional gene identity embeddings followed by layer normalization. $E$ is a 2-layer MLP that expands each 1-dimensional continuous expression value into a 512-dimensional vector, and is preceded by 0.2 dropout and followed by layer normalization. The outputs of $G$ and $E$ are summed and passed into $T$, which consists of 12 stacked Transformer blocks, each with eight attention heads. We bypass the final decoder layers of scGPT and pass the mean of all tokens (including CLS) from the last transformer layer into a 2-layer projection head $P: \R^{512} \to \R^{1024}$. During \ours pretraining, all layers of the gene expression encoder are fine-tuned. $G$, $E$, and $T$ are initialized from the \texttt{pancancer} checkpoint (pretrained on 5.7 million cells of various cancer types), while $P$ is randomly initialized.

\noindent\textbf{SNV and CNV encoder.} SNV and CNV data represent unstructured data that can be challenging to encode. Here, we adopt a simple strategy of using a multi-hot encoding passed through a 4-layer MLP (1 input, 2 hidden, and 1 output) with \relu activation and 0.2 dropout\cite{chen2020pathomic}. Details about the multi-hot encoding strategy are presented in the \textbf{Online Methods, Pretraining dataset curation}. The hidden dimension of the MLP is set to the input size (ie., 1673), and mapped to the final output dimension, 1024. Our gene mutation encoder has 10128068 parameters in total.

\noindent\textbf{Pretraining protocol.} We pretrained \ours using 4 $\times$ 80GB A100 GPUs. The model was trained with a batch size of 300 per GPU for a maximum of 101 epochs, with early stopping based on RankMe\cite{garrido2023rankme} (for details, see \textbf{Model Selection} below). We start with a 5-epoch linear warmup, gradually increasing the learning rate from 0 to $1 \times 10^{-5}$. After the warmup, we apply a cosine scheduler that decays the learning rate from $1 \times 10^{-5}$ down to $1 \times 10^{-8}$ by the end of training. The weight decay is set to 0.0001 throughout. To improve training efficiency and data diversity, we sample 512 patches per slide during each training iteration. The AdamW optimizer was employed with $\beta$ values of (0.9, 0.999). Hyperparameters and training settings are provided in \textbf{Extended Data Table \ref{tab:ssl-hparams} and \ref{tab:arch-hparams}}.

\noindent\textbf{Model selection.} During pretraining, assessing the quality of the latent space and knowing when to stop training can be challenging. Previous works have relied on monitoring the downstream task performance at regular intervals, e.g., at the end of each epoch. However, this evaluation can be computationally intensive and result in optimizing testing performance during training, which risks artificially inflating the results. Following prior work\cite{jaume2024multistain, jaume2024multistain}, we instead assess the expressivity of the embedding space by computing the smooth rank\cite{garrido2023rankme} of all slide embeddings within the training dataset after each epoch. Intuitively, a higher rank indicates greater diversity among the patch embeddings, ensuring that the representations have not collapsed into a limited number of modes, \textit{i.e.,} a low-rank space. Here, we compute the rank as the entropy of the $d$ (assuming $d<n$) L1-normalized singular values of the slide embedding matrix $H \in \mathcal{R}^{n \times d}$, which can be expressed as:
\begin{align}
    \text{RankMe}(H) &= \exp(-\sum_{k=1}^d p_k\log(p_k)),\,\, p_k = \frac{\sigma_k(H)}{\sum_{k=1}^n|\sigma_k(H)|} + \epsilon
\end{align}
where $\sigma_k$ denotes the $k$th singular value of $H$ (sorted from large to low), and $\epsilon$ is small constant set to $1e-7$ for numerical stability. A model checkpoint is saved if the rank of the training dataset increases. Rank monitoring is started after the initial learning rate ramp-up.

\noindent \textbf{Slide embedding extraction.} For evaluation, we take the final output of the slide encoder with dimensionality 1024 (\textbf{Eq. \ref{eq:softmax_attention}}). For patients with multiple WSIs, we provided the union of all patch embeddings from all WSIs (belonging to that patient) to \ours, resulting in a patient-level embedding. All patches in a slide are used while extracting slide embedding, \textit{i.e.,} no patch sampling is done Slide embeddings were extracted using bf16 precision on $1\times24$GB NVIDIA 3090Ti.

\noindent \textbf{Finetuning.} We finetune \ours slide encoder using the AdamW optimizer with a base learning rate of 0.000025. We do not apply weight decay, layer-wise learning rate decay, or gradient accumulation. All \ours finetuned models are trained with a weighted cross-entropy loss for five epochs with a batch size of 1, sampling 2048 patches per batch. We use the same learning rate scheduler as \gigapath. No early stopping was applied; the final model used is the one obtained after five epochs. All fine-tuning experiments were conducted on a $1\times24$GB NVIDIA 3090Ti with bf16 precision.

\heading{Baselines}

\noindent \ours is compared against four types of baselines: \gigapath, \prism, \chief and attention-based multiple instance learning (ABMIL). 

\hheading{\gigapath}

\noindent \textbf{Slide encoder.} \gigapath\cite{xu2024whole} is a slide encoder model consisting of a pretrained patch encoder and a pretrained slide encoder. The patch encoder is a Vision Transformer (ViT) pretrained on 171,000+ pan-cancer WSIs from over 30,000 patients using DINOv2\cite{oquab2023dinov2}. The slide encoder was trained using a LongNet\cite{ding2023longnet} model with masked autoencoding. 1536-dimensional patch features were extracted using the \gigapath patch encoder on 256$\times$256 patches at 20$\times$ magnification. We employed the official demo of \gigapath for extracting slide embeddings\footnote{https://github.com/prov-gigapath/prov-gigapath/blob/main/gigapath/slide$\_$encoder.py} and used the slide encoder checkpoint from Huggingface, resulting in a 768-dimensional pooled embedding. For patients with multiple WSIs, we processed each WSI separately and averaged the WSI embeddings. Slide embeddings were extracted using fp16 precision on $1\times24$GB NVIDIA 3090Ti.

\noindent\textbf{Finetuning.} To finetune \gigapath slide encoder, we follow the official recipe: The patch encoder is kept frozen, and we initialized \gigapath slide encoder using the HuggingFace checkpoint and added a randomly initialized linear classification head. As per the official codebase\footnote{https://github.com/prov-gigapath/prov-gigapath/blob/main/scripts/run$\_$panda.sh}, we finetuned \gigapath for all tasks using a batch size of 1, the AdamW optimizer, effective base learning rate of 0.00025, gradient accumulation over 32 steps, weight decay of 0.05, layer-wise learning rate decay of 0.95, and 5 epochs, without early stopping. We use a learning rate scheduler with half-cycle cosine decay after a one-epoch linear warmup up to the base learning rate. All patches were provided (no sampling) during training or testing. \gigapath finetuning was performed using $1\times80$GB NVIDIA A100 with fp16 precision.

\noindent\textbf{Mean pooling.} We additionally define a baseline where we average \gigapath patch embeddings, resulting into a slide embedding. 

\hheading{\prism}

\noindent \textbf{Slide encoder.} \prism is a vision-language slide encoder that uses Virchow (ViT-H/14)\cite{vorontsov2024foundation} patch encoding. Virchow was trained on $>2\times10^9$ patches (from over 100,000 patients) using DINOv2\cite{oquab2023dinov2}. \prism was then trained using a Perceiver\cite{jaegle2021perceiver} model with CoCa\cite{yu2022coca} on 587,000 WSIs-clinical reports pairs. Following the official Virchow demo\footnote{https://huggingface.co/paige-ai/Virchow}, we first tiled all WSIs into 256$\times$256-pixel patches at 20$\times$ and then used the publicly available patch encoder from HuggingFace to extract 2560-dimensional patch features. We then used the official \prism codebase\footnote{https://huggingface.co/paige-ai/Prism} to aggregate the patch embeddings of each WSI into a 1280-dimensional slide encoding. For patients with multiple WSIs, we provided the union of all patch embeddings from all WSIs (belonging to that patient) to \prism. Slide embeddings are extracted using fp16 precision on $1\times24$GB NVIDIA 3090Ti.

\noindent\textbf{Mean pooling.} We additionally define a baseline where we average \virchow patch embeddings into a slide embedding. 

\hheading{\chief} 

\noindent \textbf{Slide embedding extraction.} \chief\cite{wang2024chief} is an ABMIL-based slide encoder model that uses 768-dimensional \ctranspath patch embeddings\cite{wang2021transpath}. \chief was trained using contrastive learning by aligning the slide representation with a text embedding of the tissue site.  Following the official implementation, we compute 768-dimensional \chief pooled embeddings by passing CTransPath patch embeddings into the \chief slide encoder. For patients with multiple WSIs, we provided the union of all patch embeddings from all WSIs (belonging to that patient) to \chief. Slide embeddings are extracted using fp32 precision on $1\times24$GB NVIDIA 3090Ti.

\noindent \textbf{\chief finetuning} We employ the same finetuning recipe as in \ours as both models are based on attention-based multiple instance learning. \chief finetuning was performed using $1\times24$GB NVIDIA 3090Ti with fp32 precision.

\noindent\textbf{Mean pooling.} We additionally define a baseline where we average \ctranspath patch embeddings into a slide embedding.

\hheading{\resnet} 

\noindent\textbf{Mean pooling.} We first extract patch embeddings using a ResNet50\cite{he2016deep} model trained on ImageNet\cite{deng2009imagenet} (IN). We define a baseline where we average \resnet patch embeddings into a slide embedding.

\hheading{Attention-based multiple instance learning}

\noindent We use the widely employed attention-based multiple instance learning (\abmil) architecture. ABMIL assigns patch-level importance scores using a single-headed non-gated attention mechanism. These attention scores are used to weight the patch embeddings, which are then summed to derive a slide embedding used for classification. The model is designed with a pre-attention linear layer which preserves the dimensionality of the input patch features (768 for \conch) with \gelu activation and 0.1 dropout, an attention network with two layers (hidden dimension 512) where the first layer has $\tanh$ activation and 0.25 dropout, and a post-attention linear layer with \gelu activation and 0.1 dropout. The \abmil models were trained for 20 epochs with a batch size of 1, a learning rate of 0.0003 using a cosine scheduler, and the AdamW optimizer with a weight decay of $1\times10^{-5}$. We use the final checkpoint for evaluation without early stopping. During training, we randomly sample 2048 patch features from each WSI. In patient-level tasks, if a patient has more than one WSI, then sampling is done from the union of all patches from those WSIs. During testing, all patches are provided to the model. For classification problems, we use a balanced cross-entropy loss. For survival prediction, we use a negative log-likelihood (NLL) loss adapted for survival prediction \cite{chen2020pathomic}.

\heading{Evaluation}

\hheading{Linear probing}

\noindent \textbf{In-domain.} In classification tasks, we employ linear probing evaluation based on scikit-learn. We use a fixed cost set to 0.5, \texttt{lbfgs} solver, a maximum of 10,000 training iterations, and balanced class weights. Since we do not include a validation set, we did not perform any hyperparameter search, e.g. over the cost. Post-hoc evaluation of the impact of the cost is reported in \textbf{Extended Data Table \ref{tab:cost-scale-linprobe}} and \textbf{Extended Data Figure \ref{fig:edf3}}. To ensure fair comparisons, this evaluation recipe is applied to all sets of pooled features (\ours, \prism, \gigapath, \chief, and respective \mean baselines). In survival tasks, we use the CoxNet model from \texttt{sksurv}\cite{sksurv}, training all models for 10,000 iterations. We set the $\alpha$ parameter of the CoxNet model to 0.07 for overall survival prediction tasks and to 0.01 for progression-free survival prediction tasks. We make the following exceptions to ensure convergence: in CPTAC-CCRCC overall survival prediction, we set $\alpha$ to 0.01 in \chief, and in BOEHMK progression-free survival prediction, we set $\alpha$ to 0.02 in \prism. 

\noindent \textbf{Out-of-domain (transferability).} We run several sets of experiments to evaluate whether linear classifiers trained on one dataset can generalize to the same task on another dataset (\textbf{Extended Data Table \ref{tab:generalizability}}. We use the same setup as above and train on all samples of the "train" dataset and evaluate on a single fold containing all samples of the "test" dataset. We evaluate the performance over 100 bootstraps of the test set outputs.

\hheading{Retrieval}

\noindent We evaluate retrieval on tasks: cancer type retrieval (\textbf{Extended Data Table \ref{tab:retrieval-cptac_organ}}) and EBRAINS fine and coarse subtype retrieval (\textbf{Extended Data Table \ref{tab:retrieval-ebrains}}). For cancer type retrieval, we use 10 CPTAC cohorts: BRCA, CCRCC, COAD, HNSC, LUAD, LUSC, PDAC, GBM, OV, and UCEC. Each WSI within a cohort is labeled as the cancer type associated with that cohort. We use L2 distance as the similarity metric and simply compare the raw slide embedding of each slide in the test set with that of each slide in the training set. We consider up to the top 10 most similar retrieved slides and assess whether their class label matches that of the query slide. We compute mean average precision at $k$ (mAP@k) for $k \in \{1, 5, 10\}$. Consider a set of $n$ queries where for each query $i$ (where $i = 1, 2, \dots, n$), $y_i$ is the true label of the $i$-th query and $r_{ij}$ is the label of the retrieved item at rank $j$ for query $i$ (where $j = 1, 2, \dots, k$). Then, mAP@k is computed as:
\begin{nolinenumbers}\begin{equation*}\text{mAP@}k = \frac{1}{n} \sum_{i=1}^n \left( \frac{1}{k} \sum_{j=1}^k \delta_{r_{ij}, y_i} \cdot \frac{\sum_{s=1}^j \delta_{r_{is}, y_i}}{j} \right),\quad \text{where}\end{equation*}\begin{gather*}\delta_{a, b} = \begin{cases}1 & \text{if } a = b, \\0 & \text{if } a \ne b.\end{cases}\end{gather*}\end{nolinenumbers}

\vspace{-0.8cm}
\hheading{Prompting}

\noindent In prompting, a training dataset of RNA expression profiles with task labels is encoded using an RNA expression encoder (scGPT from \ours pretraining), producing class prompts by averaging profiles within each class. For testing with an independent dataset, a WSI is encoded via the \ours WSI encoder, and the L2 distance to all class prompts from the training set is computed. The final prediction is the class of the prompt nearest to the WSI embedding. For classification, all samples within each class are used to construct prompts. In survival prediction, prompts are based on the top and bottom 25\% of uncensored patients ranked by survival time, representing low- and high-risk categories. The risk score at test time is defined as the distance between the WSI query and high-risk prompt minus the distance between the WSI query and low-risk prompt. In prompting with $\conch$ $\mean$, WSI embeddings are formed by averaging patch embeddings and further averaged by class to create prompts. At test time, classification uses the distance between the WSI prompt and query slide embedding, while survival prediction applies the same approach, substituting high- and low-risk molecular prompts with $\mean$ WSI prompts.




\hheading{Metrics}

\noindent For binary classification tasks, we report macro-AUC. For multi-class subtyping tasks, we report balanced accuracy, and for multi-class grading tasks, we report quadratic weighted kappa score. For survival prediction tasks, we report the concordance index (c-index). \textbf{macro-AUC} is a threshold-free measure that computes the area under the receiver operating curve that plots the true positive rate against the false positive rate as the classification threshold changes. \textbf{Balanced accuracy} takes the class imbalance in the evaluation set into account by computing the unweighted average of the recall of each class. \textbf{Quadratic weighted Cohen's kappa} quantifies the agreement between two annotators (e.g., ground truth and model prediction) on a classification problem, adjusting for chance agreement and penalizing based on the distance between categories. The score ranges from -1 (complete disagreement) to 1 (perfect agreement), with 0 indicating chance-level agreement. \textbf{Concordance index} (C-index) evaluates the predictive accuracy of a risk model in survival analysis by considering the order of predicted risks and actual event times. It calculates the proportion of pairs where the individual with higher predicted risk either experiences the event earlier or is censored later. The C-index ranges from 0.5 (random) to 1 (perfect prediction). For a clear overview of the metric used for each task, see column "Metric" in \textbf{Extended Data Tables \ref{tab:task_summary.morphological_subtyping} to \ref{tab:task_summary.survival_prediction}}).

\hheading{Statistical analysis}

\noindent For all tasks with more than one test fold, we report the mean and standard error across all folds of the corresponding evaluation metric. For tasks with only a single test fold, we estimate 95\% confidence intervals with non-parametric bootstrapping using 100 bootstrap replicates.

\noindent To assess the performance of all baselines on a specific task, we first performed a two-way Analysis of Variance (ANOVA), where the null hypothesis was that mean performance values did not differ across methods. We leveraged consistent evaluation folds that enabled direct comparisons across methods. If the ANOVA showed there was a statistically significant result (\textit{i.e.,} P-value$<$0.05), a post-hoc two-sided, one-way Tukey's Honest Significant Difference (HSD) test was conducted to determine which pairs of methods differed significantly. Tukey's HSD test performs adjustment for multiple comparisons by default, so reported P-values have been adjusted for multiple comparisons\cite{dunn1961multiple}. Log-rank tests were used to compare Kaplan-Meier curves for statistically significant differences\cite{chen2020pathomic}.

\noindent In addition to comparing performance on individual tasks, we aimed to assess which model performs best across all tasks (the full benchmark and each family of tasks). To compare models on the full benchmark, we fit a mixed-effects model on the data, estimating the baseline performance while accounting for the random effect of each dataset. Contrasts between individual baselines were made using pairwise comparisons using the estimated marginal means approach \cite{searle1980population, robertson2024decoding}. Briefly, after fitting our mixed-effects model with dataset as a random effect and model type as a fixed effect, we calculate the estimated marginal means for each model. We then performed pairwise comparisons between models using Tukey's method to adjust for multiple comparisons. A similar methodology was applied for each of the four families of tasks.

\noindent For all few-shot settings (\textbf{Extended Data Fig. \ref{fig:edf4}}), we report results using box plots that indicate quartile values of model performance with whiskers extending to data points within 1.5$\times$ the interquartile range.

\heading{Computing hardware and software}

\noindent We used Python (version 3.10.12) and PyTorch (version 2.3.0, CUDA 12.3) (\url{https://pytorch.org/}) for all experiments and analyses in the study (unless otherwise specified), which can be replicated using open-source libraries as outlined below. To train \ours in a CLIP-style manner, we modified the original CLIP algorithm implemented by \url{https://github.com/mlfoundations/open_clip}. We used the implementation of scGPT from \url{https://github.com/bowang-lab/scGPT}. For pretraining, we used 4$\times$ 80GB NVIDIA A100 GPUs configured for multi-GPU training using distributed data-parallel (DDP). All other computations for downstream experiments were performed on single 24GB NVIDIA 3090 GPUs. All WSI processing was supported by OpenSlide (version 4.3.1), openslide-python (version 1.3.1), and Trident (\url{https://github.com/mahmoodlab/trident}), which additionally requires Pillow (version 10.2.0), segmentation-models-pytorch (version 0.0.3), and opencv-python (version 4.10.0.84). We use Scikit-learn\cite{pedregosa2011scikit} (version 1.5.0), Scikit-survival (version 0.23.0), and faiss (version 1.8.0) for training downstream machine learning models, specifically Logistic regression, and Cox PH. Implementations of other slide encoders benchmarked in the study are found at the following links: GigaPath (\url{https://github.com/prov-gigapath/prov-gigapath}), which additionally required fairscale (version 0.4.13), flash-attn (version 2.5.8), and ninja (version 1.11.1.1), PRISM (\url{https://huggingface.co/paige-ai/Prism}), and CHIEF (\url{https://github.com/hms-dbmi/CHIEF}).
Matplotlib (version 3.8.4) and Seaborn (version 0.13.2) were used to create plots and figures. Usage of other miscellaneous Python libraries is listed in the \textbf{Reporting Summary}.

\heading{Code availability}

\noindent Preprocessing code to (i) segment tissue from background, (ii) whole-slide image patching, and (iii) patch embedding extraction for \conch, \ctranspath, \gigapath, and \virchow can be accessed at \url{https://github.com/mahmoodlab/trident}.
Code to run our benchmark can be accessed from \url{https://github.com/mahmoodlab/patho-bench}.
Access to curated labels of publicly available cohorts, and data splits employed in the study can be found at \url{https://huggingface.co/datasets/MahmoodLab/patho-bench}.
\ours model weights and code to extract slide embeddings will be released upon publication.

\heading{Data availability}

\noindent
\textbf{\mbtg:} TCGA imaging data can be accessed through the NIH genomic data commons (\url{https://portal.gdc.cancer.gov}). TCGA transcriptomics data can be accessed through the Xena Hub (\url{https://xenabrowser.net/}).
GTEx imaging and transcriptomics data can be accessed through the GTEx portal (\url{https://www.gtexportal.org/home/}).
Pretraining data from BWH and MGH are proprietary patient data, and cannot be made publicly available.

\noindent\textbf{Benchmark:} Download links to access publicly available cohorts included as part of our benchmark are reported in \textbf{Extended Data Table~\ref{tab:downstream-links}}. Curated labels can be accessed via the \ours-Benchmarking GitHub repository. In-house cohorts cannot be made publicly available. 


 

\heading{Author contributions}

\noindent
A.V., A.Z., G.J. conceived the study and designed the experiments.
A.V., A.Z., G.J., A.H.S., C.A.P., P.D., S.W., collected the data for pretraining. D.L. and K.L. assisted with curating downstream tasks.
A.V., A.Z., G.J. T.D., M.Y.L. performed model development.
A.V., A.Z., G.J., P.D., M.Y.L., T.D. R.C organized the datasets and codebase for all downstream tasks.
A.V., A.Z., G.J., P.D., S.W., T.D. performed experimental analysis.
A.V., A.Z., G.J., A.H.S., D.L., K.L. G.G., L.P.L. interpreted the experimental results and provided feedback on the study.
A.V., A.Z., G.J. prepared the manuscript with input from all co-authors.
H.R. performed statistical analysis.
F.M. supervised the research.

\heading{Acknowledgments}

\noindent This work was partly supported by the BWH \& MGH Pathology, BWH president’s fund, Massachusetts Life Sciences Center, National Institute of General Medical Sciences (NIGMS) R35GM138216 (F.M.), and the BWH President's Scholar fund (G.G.).
S.J.W. was supported by the Helmholtz Association under the joint research school ``Munich School for Data Science - MUDS'' and the Add-on Fellowship of the Joachim Herz Foundation. This work was supported by a fellowship of the German Academic Exchange Service (DAAD).
M.Y.L. was supported by the Tau Beta Pi Fellowship and the Siebel Foundation.
We thank Muhammad Shaban and Timothy Janicki at BWH, and Richard Kenny and the system administration staff at the MGB Enterprise Research Infrastructure \& Services (ERIS) Research Computing Core, for their support in maintaining access to computing resources. We thank Prof. Ahrong Kim and Prof. Mihaela Aldea for interpreting some results and providing feedback on the study design. Some illustrations were created with BioRender.com. The content is solely the responsibility of the authors and does not reflect the official views of the NIH, NIGMS, NCI, DoD.

\newpage
\setcounter{figure}{0}
\renewcommand{\figurename}{Extended Data Figure}
\renewcommand{\tablename}{Extended Data Table}

\begin{figure}[t]
\centering
\includegraphics[width=\textwidth]{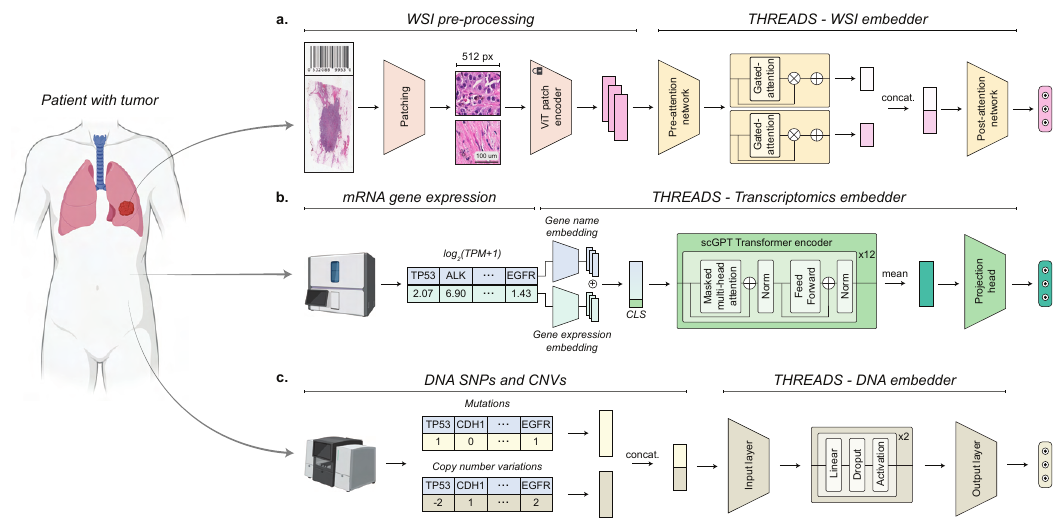}
\caption{\textbf{Detailed architecture of \ours.} \ours employs a multimodal contrastive learning approach to align a whole-slide image representation with its corresponding molecular profile, obtained either using a DNA or RNA assay. 
\textbf{a.} The vision encoding branch uses a multihead attention-based model to pool patch embeddings into a slide embedding. 
\textbf{b.} The RNA encoding branch uses an scGPT model pretrained on 5.7 million cells of various cancer types, which is fully fine-tuned to yield a transcriptome embedding. 
\textbf{c.} The DNA encoding branch uses a multilayer perceptron (MLP) to transform copy number variations (CNV), insertions and deletions (indels), and single nucleotide variants (SNV) into a genomic embedding. 
WSI: whole-slide image; ViT: vision transformer; concat.: concatenations; TPM: transcripts per million.
}
\label{fig:edf1}
\end{figure}

\clearpage
\begin{figure}[t]
\centering
\includegraphics[width=\textwidth]{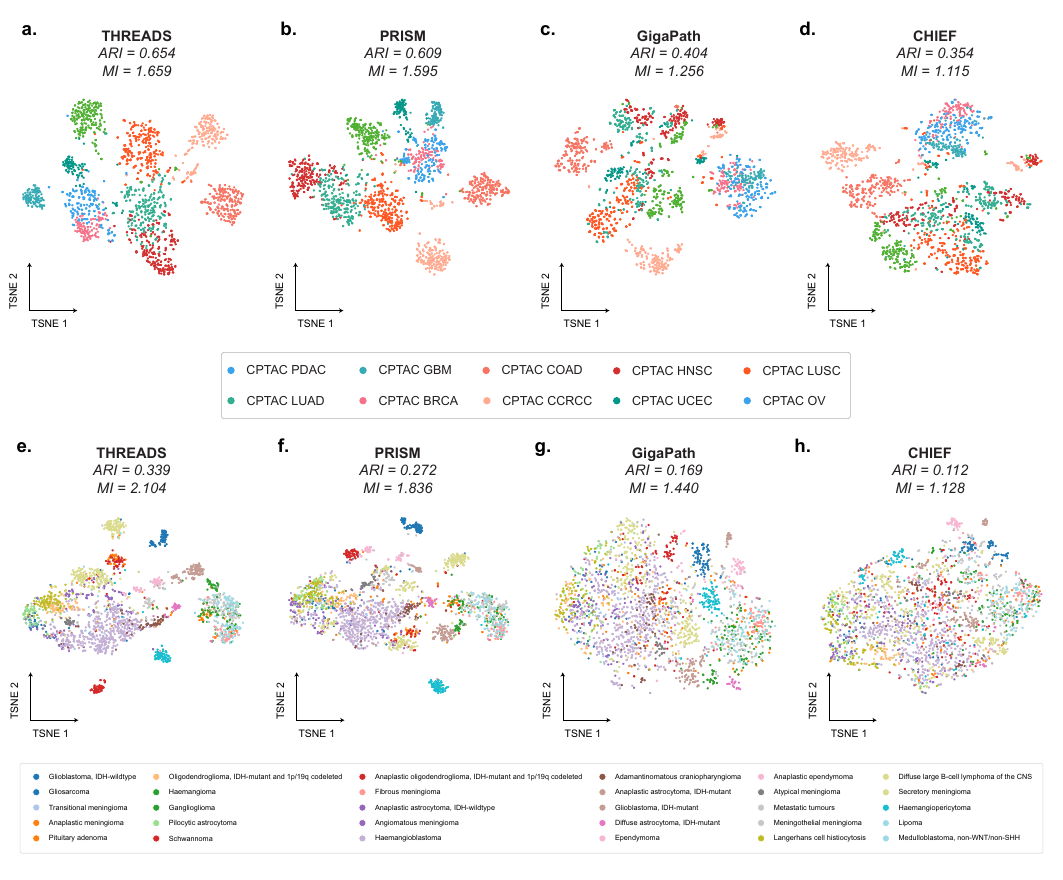}
\caption{\textbf{Clustering capabilities of \ours.}
2-dimensional tSNE representation of CPTAC cohort stratified by cancer type (n=10 cancer types) using \ours (\textbf{a.}), PRISM (\textbf{b.}), GigaPath (\textbf{c.}), and CHIEF (\textbf{d.}). 
2-dimensional tSNE representation of EBRAINS cohort stratified by tumor type (n=12 tumor types) using \ours (\textbf{e.}), PRISM (\textbf{f.}), GigaPath (\textbf{g.}), and CHIEF (\textbf{g.}). ARI: Adjusted random index; MI: Mutual information; tSNE: t-distributed stochastic neighbor embedding. 
}
\label{fig:edf2}
\end{figure}

\clearpage
\begin{figure}[t]
\centering
\includegraphics[width=1.0\textwidth]{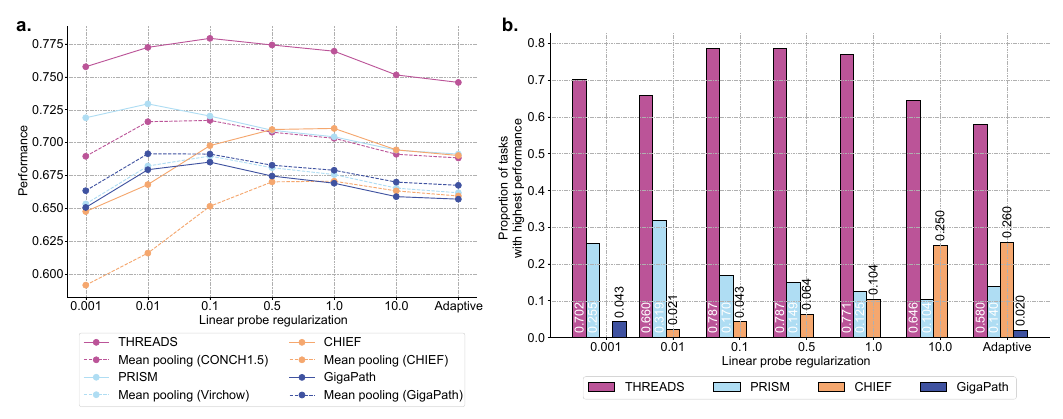}
\caption{\textbf{Impact of regularization in linear probing evaluation in our benchmark (54 tasks).}
\textbf{a.} Evolution of the average performance when varying the cost (inverse of regularization strength) in linear probing evaluation. 
\textbf{b.} Percentage of tasks where each baseline performs best based on the cost in linear probing evaluation. 
Adaptive regularization computes regularization cost by taking the embedding dimension times the number of classes normalized by 100\cite{kolesnikov2019revisiting}. 
}
\label{fig:edf5}
\end{figure}

\clearpage
\begin{figure}[t]
\centering
\includegraphics[width=\textwidth]{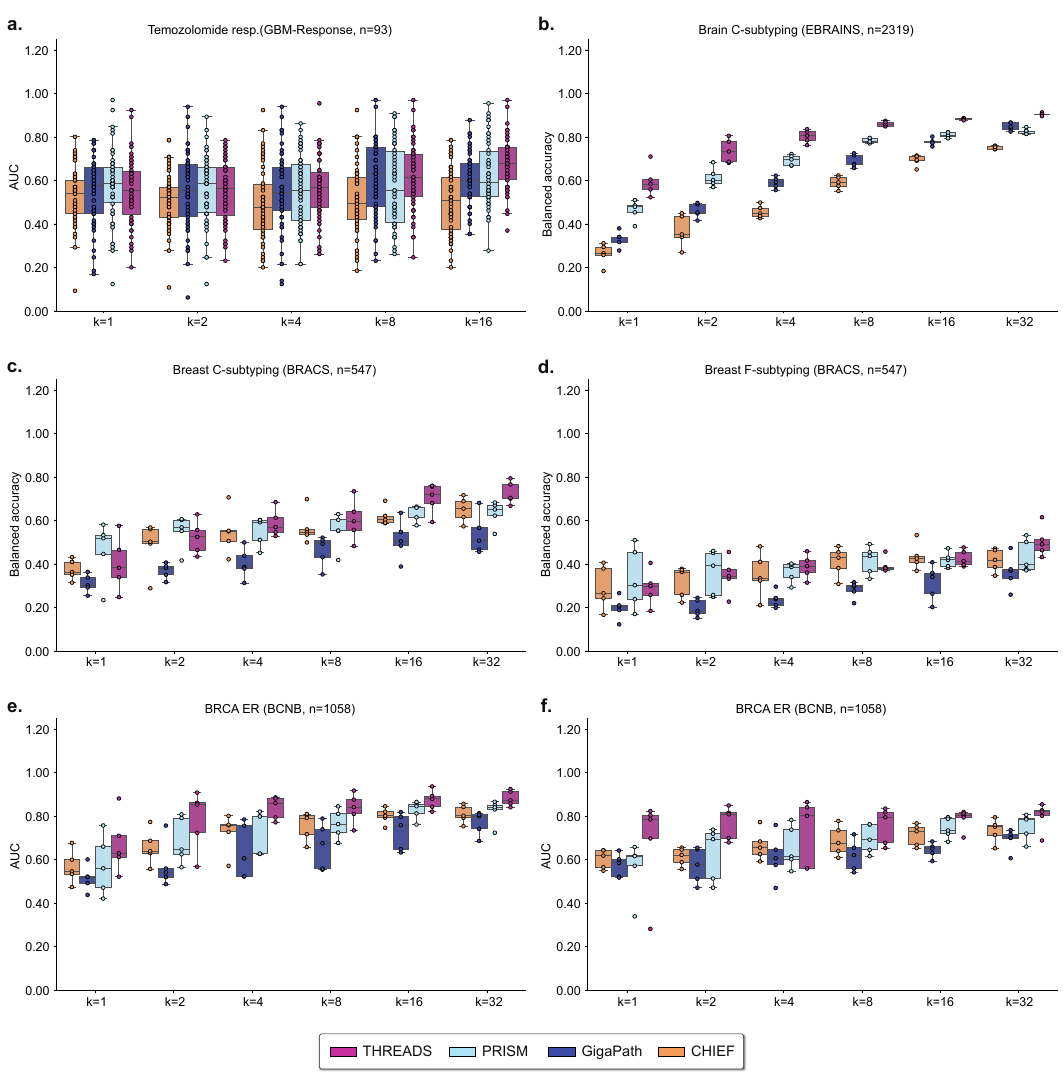}
\caption{\textbf{Few shot performance of \ours compared to baselines.} $k$ refers to the number of training samples per class. 
\textbf{a.} Temozolomide response prediction in glioblastoma. 
\textbf{b.} Coarse-grained brain tumor subtyping in EBRAINS dataset. 
\textbf{c.} Fine-grained breast tumor subtyping in BRACS dataset. 
\textbf{d.} Coarse-grained breast tumor subtyping in BRACS dataset.
\textbf{e.} Estrogen receptor status prediction in BCNB. 
\textbf{f.} Progesterone receptor status prediction in BCNB. 
Boxes indicate quartile values of model performance (n=5 runs), and whiskers extend to data points within 1.5-fold the interquartile range.}
\label{fig:edf3}
\end{figure}

\begin{figure}[h]
\centering
\includegraphics[width=0.85\textwidth]{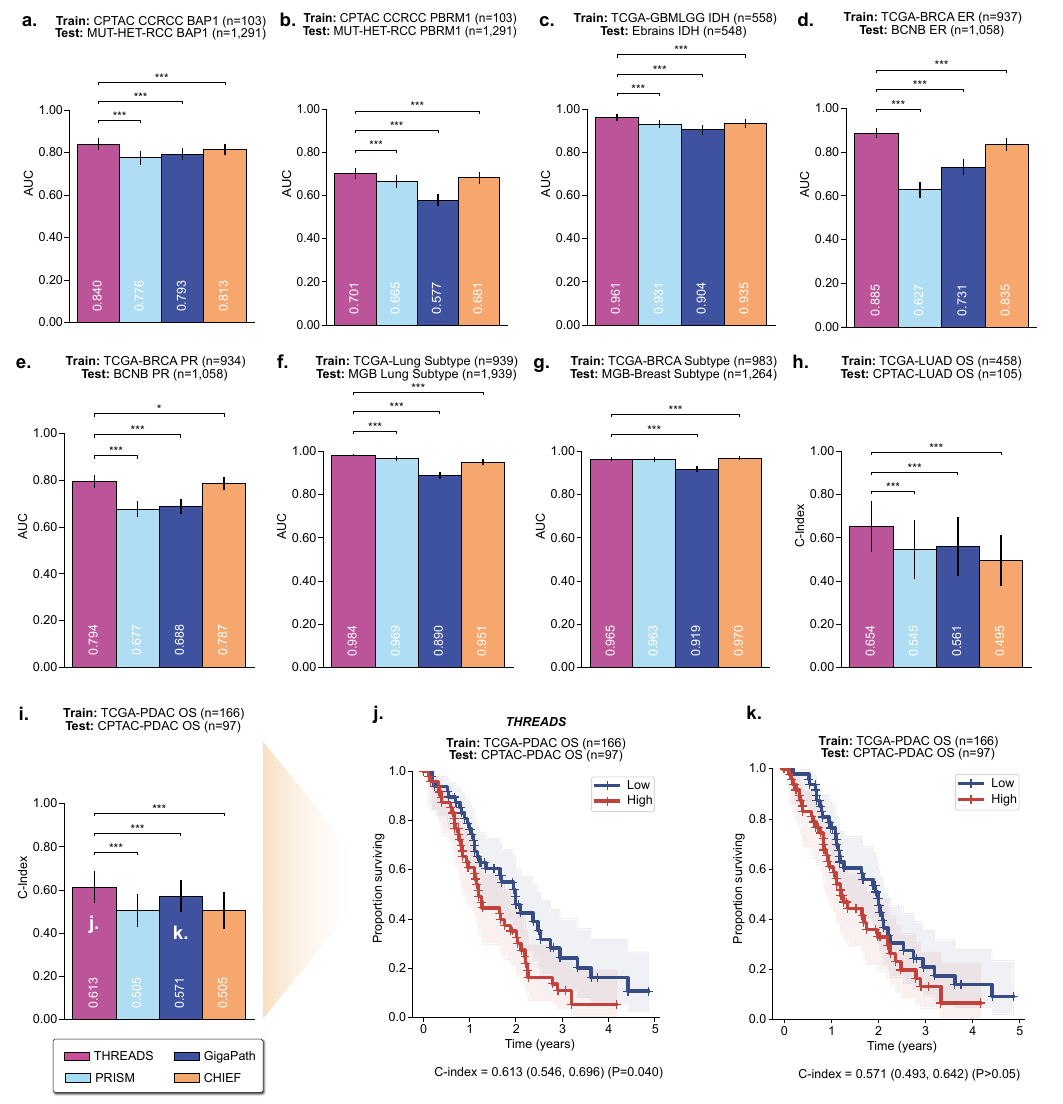}
\caption{\textbf{Transferability of \ours.} \ours and baselines are trained on one cohort and tested on an independent unseen test cohort. All test cohorts are independent of the \mbtg pretraining cohort. 
\textbf{a.} Clear cell renal cell carcinoma (ccRCC) BAP1 mutation prediction (CPTAC $\rightarrow$ MUT-HET-RCC). 
\textbf{b.} ccRCC PBRM1 mutation prediction (CPTAC $\rightarrow$ MUT-HET-RCC).
\textbf{c.} Glioblastoma and low-grade glioma (GBMLGG) IDH mutation prediction (TCGA $\rightarrow$ EBRAINS).
\textbf{d.} Invasive breast cancer (BRCA) estrogen receptor (ER) status prediction (TCGA $\rightarrow$ BCNB).
\textbf{e.} BRCA progesterone receptor (ER) status prediction (TCGA $\rightarrow$ BCNB).
\textbf{f.} BRCA subtype prediction (TCGA $\rightarrow$ MGB-Breast).
\textbf{g.} NSCLC (lung adenocarcinoma, LUAD and lung squamous cell carcinoma LUSC) subtyping prediction (TCGA $\rightarrow$ MGB-Lung).
\textbf{h.} LUAD overall survival prediction (TCGA $\rightarrow$ CPTAC). 
\textbf{i.} Pancreatic adenocarcinoma (PDAC) overall survival prediction (TCGA $\rightarrow$ CPTAC).
\textbf{j.} Kaplan Meier (KM) curve of \ours for PDAC overall survival prediction (TCGA $\rightarrow$ CPTAC).
\textbf{k.} KM curve for GigaPath overall survival prediction (TCGA $\rightarrow$ CPTAC).
Error bars represent 95\% confidence intervals and the centers correspond to computed values of each metric. In Kaplan-Meier curves, line shows value and shaded region shows 95\% confidence interval.
Task-wise P-values were determined using two-sided Tukey Honest Significance Difference tests accounting for multiple comparisons following a positive result (P$<$0.05) of a two-way ANOVA.
Statistical significance across multiple tasks (e.g., for each family) was assessed using a mixed-effects model. 
In Kaplan-Meier curves, P-values correspond to log-rank tests.
P$<$0.05: *, P$<$0.01: **, P$<$0.001: ***.
}
\label{fig:fig3}
\end{figure}

\clearpage
\begin{figure}[t]
\centering
\includegraphics[width=\textwidth]{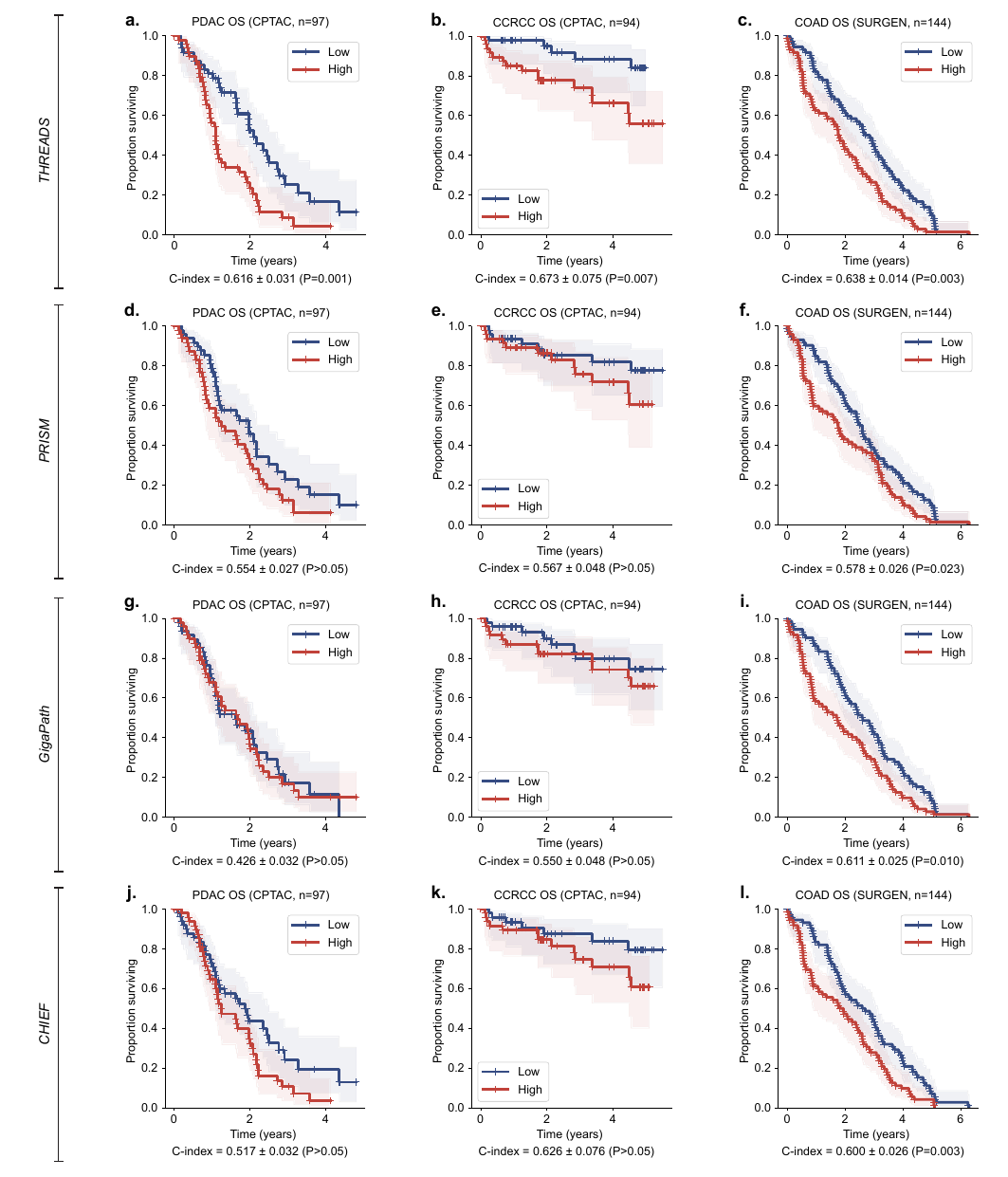}
\caption{\textbf{Survival analysis of \ours and baselines.}
Kaplan Meier survival plots of \ours (\textbf{a,b,c}), PRISM (\textbf{d,e,f}),  GigaPath (\textbf{g,h,i}), and CHIEF (\textbf{j,k,l}) tested on pancreatic adenocarcinoma (PDAC), clear cell renal cell carcinoma (ccRCC), and colon adenocarcima (COAD). 
The shaded region highlights 95\% confidence intervals.
P-values for Kaplan Meier curves were obtained using log-rank statistical testing.
}
\label{fig:edf4}
\end{figure}

\clearpage
\begin{nolinenumbers}
\setcounter{table}{0}
\renewcommand{\tablename}{Extended Data Table}
\begin{spacing}{0.9}

\begin{table}[h]
    \centering
    \caption{\textbf{Tissue Type Distribution in MBTG-47k.} $\ours$-pretraining consists of 47,171 WSIs across 40 major tissue types collected from Massachusetts General Hospital (MGH), Brigham \& Women's Hospital (BWH), The Cancer Genome Atlas (TCGA), and the Genotype-Tissue Expression (GTEx) consortium. Primary organs defined according to the highest level of the oncotree classification.}
    \scalebox{0.95}{
        \begin{tabular}{l|rrrrr}
            \toprule
            \multirow{2}{*}{Primary organ} & \multicolumn{5}{c}{Number of slides} \\
             & BWH & GTEx & MGH & TCGA & Total \\
            \midrule
            \midrule
            Adipose & 0 & 879 & 0 & 0 & 879 \\
            Adrenal Gland & 182 & 162 & 8 & 387 & 739 \\
            Ampulla Of Vater & 23 & 0 & 12 & 0 & 35 \\
            Artery & 0 & 882 & 0 & 0 & 882 \\
            Biliary Tract & 97 & 0 & 112 & 36 & 245 \\
            Bladder/Urinary Tract & 596 & 0 & 97 & 451 & 1144 \\
            Blood & 5 & 0 & 0 & 0 & 5 \\
            Bone & 419 & 0 & 18 & 4 & 441 \\
            Bowel & 2150 & 691 & 661 & 373 & 3875 \\
            Breast & 457 & 327 & 256 & 1124 & 2164 \\
            Cervix & 110 & 0 & 20 & 275 & 405 \\
            CNS/Brain & 2394 & 646 & 774 & 1080 & 4894 \\
            Esophagus/Stomach & 0 & 1202 & 276 & 0 & 1478 \\
            Eye & 20 & 0 & 24 & 80 & 124 \\
            Head And Neck & 549 & 129 & 449 & 470 & 1597 \\
            Heart & 0 & 620 & 0 & 1 & 621 \\
            Kidney & 508 & 64 & 114 & 885 & 1571 \\
            Liver & 1825 & 0 & 172 & 337 & 2334 \\
            Lung & 1264 & 383 & 1620 & 1042 & 4309 \\
            Lymph & 1740 & 0 & 0 & 138 & 1878 \\
            Lymphoid & 0 & 0 & 2 & 0 & 2 \\
            Muscle & 0 & 563 & 0 & 0 & 563 \\
            Ovary/Fallopian Tube & 820 & 116 & 173 & 71 & 1180 \\
            Pancreas & 587 & 201 & 425 & 196 & 1409 \\
            Penis & 6 & 0 & 0 & 0 & 6 \\
            Peripheral Nervous System & 16 & 436 & 0 & 12 & 464 \\
            Peritoneum & 469 & 0 & 27 & 2 & 498 \\
            Pleura & 613 & 0 & 0 & 86 & 699 \\
            Prostate & 474 & 165 & 127 & 447 & 1213 \\
            Skin & 686 & 988 & 542 & 322 & 2538 \\
            Soft Tissue & 1509 & 0 & 217 & 614 & 2340 \\
            Spleen & 0 & 158 & 2 & 2 & 162 \\
            Stomach & 1037 & 0 & 0 & 544 & 1581 \\
            Testis & 88 & 245 & 1 & 256 & 590 \\
            Thorax & 0 & 0 & 0 & 3 & 3 \\
            Thymus & 38 & 0 & 6 & 142 & 186 \\
            Thyroid & 401 & 465 & 377 & 518 & 1761 \\
            Unknown & 431 & 0 & 230 & 0 & 661 \\
            Uterus & 939 & 89 & 129 & 310 & 1467 \\
            Vulva/Vagina & 103 & 96 & 28 & 1 & 228 \\
            \midrule
            \midrule
            Total & 20556 & 9507 & 6899 & 10209 & 47171 \\
            \bottomrule
            \end{tabular}
            }
\label{tab:pretrain-counts}
\end{table}

\begin{table}[h]
    \centering
    \caption{\textbf{Summary of morphological subtyping prediction tasks.} All are WSI-level prediction tasks. \textbf{BA}: balanced accuracy, \textbf{AUC}: area under the receiver operating characteristic curve.}
    \scalebox{0.675}{
    \begin{tabular}{llccccccccc}
    \toprule
    & & & & & & & & \multicolumn{3}{c}{\textbf{\textit{k}=All Splits}} \\
    \cmidrule(lr){9-11}
    \textbf{Datasource} & \textbf{Subtyping Task} & \textbf{Organ} & \textbf{Unit} & \textbf{\# Patients} & \textbf{\# WSIs} & \textbf{\# Classes} & \textbf{Metric} & \textbf{Train:Test} & \textbf{Official?} & \textbf{\# Folds} \\
    \midrule
     MGB-BRCA & IDC vs. ILC & Breast & WSI & 1264 & 1264 & 2 & AUC & 1011:253 & No & 5 \\
    MGB-Lung & LUAD vs. LUSC & Lung & WSI & 1939 & 1939 & 2 & AUC & 1551:388 & No & 5 \\
    EBRAINS & Coarse & Brain & WSI & 2147 & 2319 & 12 & BA & 1746:573 & Yes & 1 \\
    EBRAINS & Fine & Brain & WSI & 2147 & 2319 & 30 & BA & 1746:573 & Yes & 1 \\
    BRACS & Coarse & Breast & WSI & 189 & 547 & 3 & BA & 396:151 & No & 5 \\
    BRACS & Fine & Breast & WSI & 189 & 547 & 7 & BA & 396:151 & No & 5 \\
    \bottomrule
    \end{tabular}
    }

    \label{tab:task_summary.morphological_subtyping}
\end{table}

\begin{table}[h]
    \centering
    \caption{\textbf{Summary of tumor grading prediction tasks.} Both are WSI-level prediction tasks. \textbf{QWK}: quadratic weighted kappa.}
    
    \scalebox{0.675}{
    \begin{tabular}{lccccccccc}
    \toprule
    & & & & & & & \multicolumn{3}{c}{\textbf{\textit{k}=All Splits}} \\
    \cmidrule(lr){8-10}
    \textbf{Datasource} & \textbf{Organ} & \textbf{Unit} & \textbf{\# Patients} & \textbf{\# WSIs} & \textbf{\# Classes} & \textbf{Metric} & \textbf{Train:Test} & \textbf{Official?} & \textbf{\# Folds} \\
    \midrule
    PANDA & Prostate & WSI & 9555 & 9555 & 6 & QWK & 7647:954 & Yes & 1 \\
    IMP & Colon & WSI & 5333 & 5333 & 3 & QWK & 4433:900 & Yes & 1 \\
    \bottomrule
    \end{tabular}
    }
    \label{tab:task_summary.tumor_grading}
\end{table}

\begin{table}[h]
    \centering
    \caption{\textbf{Summary of molecular subtyping tasks.} All are patient-level prediction tasks. \% Positive refers to percentage of samples with positive marker. \textbf{AUC}: area under the receiver operating characteristic curve.}



    \scalebox{0.675}{
    \begin{tabular}{lcccccccccc}
    \toprule
    & & & & & & & & \multicolumn{3}{c}{\textbf{\textit{k}=All Splits}} \\
    \cmidrule(lr){9-11}
    \textbf{Datasource} & \textbf{Marker} & \textbf{Organ} & \textbf{Unit} & \textbf{\# Patients} & \textbf{\# WSIs} & \textbf{\% Positive} & \textbf{Metric} & \textbf{Train:Test} & \textbf{Official?} & \textbf{\# Folds} \\
    \midrule
    MGB-BRCA & ER & Breast & Patient & 874 & 874 & 70.1\% & AUC & 699:175 & No & 5 \\
    MGB-BRCA & PR & Breast & Patient & 874 & 874 & 57.7\% & AUC & 699:175 & No & 5 \\
    MGB-BRCA & HER2 & Breast & Patient & 816 & 816 & 18.5\% & AUC & 652:164 & No & 5 \\
    BCNB & ER & Breast & Patient & 1058 & 1058 & 78.5\% & AUC & 846:212 & No & 5 \\
    BCNB & PR & Breast & Patient & 1058 & 1058 & 74.7\% & AUC & 846:212 & No & 5 \\
    BCNB & HER2 & Breast & Patient & 1058 & 1058 & 26.2\% & AUC & 846:212 & No & 5 \\
    MGB-Lung & TTF-1 & Lung & Patient & 488 & 488 & 67.0\% & AUC & 390:98 & No & 5 \\
    MGB-Lung & P40 & Lung & Patient & 185 & 185 & 38.9\% & AUC & 148:37 & No & 5 \\
    MGB-Lung & P63 & Lung & Patient & 153 & 153 & 52.9\% & AUC & 122:31 & No & 5 \\
    MGB-Lung & Napsin A & Lung & Patient & 126 & 126 & 52.4\% & AUC & 100:26 & No & 5 \\
    MGB-Lung & CDX-2 & Lung & Patient & 79 & 79 & 30.4\% & AUC & 63:16 & No & 5 \\
    MGB-Lung & CK5/6 & Lung & Patient & 58 & 58 & 50.0\% & AUC & 46:12 & No & 5 \\
    \bottomrule
    \end{tabular}
    }
    \label{tab:task_summary.molecular_subtyping}
\end{table}

\begin{table}[h]
    \centering
    \caption{\textbf{Summary of mutation prediction tasks.} All are patient-level prediction tasks. \textbf{AUC}: area under the receiver operating characteristic curve.}

    \scalebox{0.675}{
    \begin{tabular}{lcccccccccc}
    \toprule
    & & & & & & & & \multicolumn{3}{c}{\textbf{\textit{k}=All Splits}} \\
    \cmidrule(lr){9-11}
    \textbf{Datasource} & \textbf{Gene} & \textbf{Organ} & \textbf{Unit} & \textbf{\# Patients} & \textbf{\# WSIs} & \textbf{\% Mutated} & \textbf{Metric} & \textbf{Train:Test} & \textbf{Official?} & \textbf{\# Folds} \\
    \midrule
    CPTAC-BRCA & PIK3CA & Breast & Patient & 103 & 112 & 35.9\% & AUC & 83:20 & No & 50 \\
    CPTAC-BRCA & TP53 & Breast & Patient & 103 & 112 & 40.8\% & AUC & 83:20 & No & 50 \\
    CPTAC-CCRCC & BAP1 & Kidney & Patient & 103 & 245 & 16.5\% & AUC & 83:20 & No & 50 \\
    CPTAC-CCRCC & PBRM1 & Kidney & Patient & 103 & 245 & 45.6\% & AUC & 83:20 & No & 50 \\
    CPTAC-COAD & KRAS & Colon & Patient & 94 & 98 & 38.3\% & AUC & 76:18 & No & 50 \\
    CPTAC-COAD & TP53 & Colon & Patient & 94 & 98 & 66.0\% & AUC & 76:18 & No & 50 \\
    CPTAC-GBM & EGFR & Brain & Patient & 99 & 243 & 24.2\% & AUC & 80:19 & No & 50 \\
    CPTAC-GBM & TP53 & Brain & Patient & 99 & 243 & 32.3\% & AUC & 80:19 & No & 50 \\
    CPTAC-HNSC & CASP8 & Head \& Neck & Patient & 107 & 256 & 10.3\% & AUC & 86:21 & No & 50 \\
    CPTAC-LSCC & KEAP1 & Lung & Patient & 108 & 304 & 12.0\% & AUC & 87:21 & No & 50 \\
    CPTAC-LSCC & ARID1A & Lung & Patient & 108 & 304 & 12.0\% & AUC & 87:21 & No & 50 \\
    CPTAC-LUAD & EGFR & Lung & Patient & 108 & 324 & 36.1\% & AUC & 87:21 & No & 50 \\
    CPTAC-LUAD & STK11 & Lung & Patient & 108 & 324 & 16.7\% & AUC & 87:21 & No & 50 \\
    CPTAC-LUAD & TP53 & Lung & Patient & 108 & 324 & 59.3\% & AUC & 87:21 & No & 50 \\
    CPTAC-PDAC & SMAD4 & Pancreas & Patient & 105 & 242 & 19.0\% & AUC & 84:21 & No & 50 \\
    MUT-HET-RCC & BAP1 & Kidney & Patient & 1291 & 1291 & 12.5\% & AUC & 1032:259 & No & 5 \\
    MUT-HET-RCC & PBRM1 & Kidney & Patient & 1291 & 1291 & 51.8\% & AUC & 1032:259 & No & 5 \\
    MUT-HET-RCC & SETD2 & Kidney & Patient & 1291 & 1291 & 27.0\% & AUC & 1032:259 & No & 5 \\
    SURGEN & BRAF & Colon & Patient & 388 & 388 & 10.8\% & AUC & 310:78 & No & 5 \\
    SURGEN & RAS & Colon & Patient & 389 & 389 & 35.5\% & AUC & 311:78 & No & 5 \\
    SURGEN & MMR & Colon & Patient & 389 & 389 & 7.7\% & AUC & 311:78 & No & 5 \\
    \bottomrule
    \end{tabular}
    }
    \label{tab:task_summary.mutation_prediction}
\end{table}

\begin{table}[h]
    \centering
    \caption{\textbf{Summary of treatment response and assessment tasks.} Unit refers to whether a task is a patient-level or WSI-level prediction task. \% Positive refers to the fraction of cases either exhibiting positive (favorable) response or the criterion specified. Type refers to whether slides are resections, biopsies, or both. \textbf{AUC}: area under the receiver operating characteristic curve.}
    
    \scalebox{0.575}{
    \begin{tabular}{lllllrrllllrlr}
    \toprule
    & & & & & & & & & \multicolumn{3}{c}{\textbf{\textit{k}=All Splits}} \\
    \cmidrule(lr){10-12}
    \textbf{Datasource} & \textbf{Criterion} & \textbf{Organ} & \textbf{Type} & \textbf{Unit} & \textbf{\# Patients} & \textbf{\# WSIs} & \textbf{\% Positive} & \textbf{Metric} & \textbf{Train:Test} & \textbf{Official?} & \textbf{\# Folds} \\
    \midrule
    NADT-Prostate & Radiological response & Prostate & Resection & Patient & 36 & 449 & 41.7\% & AUC & 29:7 & No & 50 \\
    OV-Bevacizumab & Biomarker response & Ovary & Resection & Patient & 36 & 85 & 83.3\% & AUC & 29:7 & No & 50 \\
    GBM-Treatment & Clinical outcome & Brain & Biopsy & Patient & 93 & 347 & 73.1\% & AUC & 75:18 & No & 50 \\
    POST-NAT-BRCA & Lymphovascular invasion & Breast & Resection & WSI & 50 & 53 & 30.2\% & AUC & 43:10 & No & 50 \\
    MBC & Recist & Breast & Both & Patient & 76 & 97 & 46.1\% & QWK & 61:15 & No & 50 \\
    BOEHMK & PFS & Ovary & Both & Patient & 183 & 183 & -- & C-Index & 146:37 & No & 5 \\
    MBC & OS & Ovary & Biopsy & Patient & 75 & 96 & -- & C-Index & 60:15 & No & 5\\
    \bottomrule
    \end{tabular}
    }
    \label{tab:task_summary.treatment_response}
\end{table}

\begin{table}[h]
    \centering
    \caption{\textbf{Summary of survival prediction tasks.} All tasks are patient-level tasks. Censorship refers to incomplete observations due to end of study period or loss of follow-up. \textbf{OS}: duration of overall survival, \textbf{C-Index}: concordance index.}

    \scalebox{0.65}{
    \begin{tabular}{llllrrlllllr}
    \toprule
    & & & & & & & & & \multicolumn{3}{c}{\textbf{\textit{k}=All Splits}} \\
    \cmidrule(lr){10-12}
    \textbf{Datasource} & \textbf{Task} & \textbf{Organ} & \textbf{Unit} & \textbf{\# Patients} & \textbf{\# WSIs} & \textbf{\% Censored} & \textbf{Survival (days)} & \textbf{Metric} & \textbf{Train:Test} & \textbf{Official?} & \textbf{\# Folds} \\
    \midrule
    CPTAC-CCRCC & OS & Kidney & Patient & 94 & 218 & 78.7\% & 1064 ± 634 & C-Index & 75:19 & No & 5 \\
    CPTAC-HNSC & OS & Head \& Neck & Patient & 102 & 243 & 67.6\% & 833 ± 423 & C-Index & 81:21 & No & 5 \\
    CPTAC-LUAD & OS & Lung & Patient & 105 & 313 & 78.1\% & 753 ± 540 & C-Index & 84:21 & No & 5 \\
    CPTAC-PDA & OS & Pancreas & Patient & 97 & 227 & 26.8\% & 561 ± 379 & C-Index & 77:20 & No & 5 \\
    SURGEN & OS & Colon & Patient & 144 & 144 & 0.0\% & 854 ± 566 & C-Index & 115:29 & No & 5 \\
    SURGEN & Died within 5 years & Colon & Patient & 387 & 387 & -- & -- & AUC & 309:78 & No & 5 \\
    \bottomrule
    \end{tabular}
    }
    \label{tab:task_summary.survival_prediction}
\end{table}

\begin{table}[h]
    \centering
    \caption{\textbf{Publicly available datasets used for \ours evaluation.}}
        \begin{tabular}{l|l}
            \toprule
            Dataset & Link \\
            \midrule
            \midrule
            EBRAINS\cite{roetzer2022ebrains} & \href{https://doi.org/10.25493/WQ48-ZGX}{https://doi.org/10.25493/WQ48-ZGX} \\
            BRACS\cite{brancati2021bracs} & \href{https://www.bracs.icar.cnr.it/}{https://www.bracs.icar.cnr.it/} \\
            PANDA\cite{bulten2022artificial} & \href{https://panda.grand-challenge.org/data/}{https://panda.grand-challenge.org/data/} \\
            IMP\cite{neto2024interpretable} & \href{https://rdm.inesctec.pt/dataset/nis-2023-008}{https://rdm.inesctec.pt/dataset/nis-2023-008} \\
            BCNB\cite{xu2021predicting} & \href{https://bupt-ai-cz.github.io/BCNB/}{https://bupt-ai-cz.github.io/BCNB/} \\
            CPTAC-BRCA\cite{edwards2015cptac} & \href{https://www.cancerimagingarchive.net/collection/cptac-brca/}{https://www.cancerimagingarchive.net/collection/cptac-brca/} \\
            CPTAC-CCRCC\cite{edwards2015cptac} & \href{https://www.cancerimagingarchive.net/collection/cptac-ccrcc/}{https://www.cancerimagingarchive.net/collection/cptac-ccrcc/} \\
            CPTAC-COAD\cite{edwards2015cptac} & \href{https://www.cancerimagingarchive.net/collection/cptac-coad/}{https://www.cancerimagingarchive.net/collection/cptac-coad/} \\
            CPTAC-GBM\cite{edwards2015cptac} & \href{https://www.cancerimagingarchive.net/collection/cptac-gbm/}{https://www.cancerimagingarchive.net/collection/cptac-gbm/} \\
            CPTAC-HNSC\cite{edwards2015cptac} & \href{https://www.cancerimagingarchive.net/collection/cptac-hnsc/}{https://www.cancerimagingarchive.net/collection/cptac-hnsc/} \\
            CPTAC-LSCC\cite{edwards2015cptac} & \href{https://www.cancerimagingarchive.net/collection/cptac-lscc/}{https://www.cancerimagingarchive.net/collection/cptac-lscc/} \\
            CPTAC-LUAD\cite{edwards2015cptac} & \href{https://www.cancerimagingarchive.net/collection/cptac-luad/}{https://www.cancerimagingarchive.net/collection/cptac-luad/} \\
            CPTAC-PDAC\cite{edwards2015cptac} & \href{https://www.cancerimagingarchive.net/collection/cptac-pda/}{https://www.cancerimagingarchive.net/collection/cptac-pda/} \\
            MUT-HET-RCC\cite{} & \href{https://doi.org/10.25452/figshare.plus.c.5983795}{https://doi.org/10.25452/figshare.plus.c.5983795} \\
            OV-Bevacizumab\cite{wang2022weakly} & \href{https://www.nature.com/articles/s41597-022-01127-6}{https://www.nature.com/articles/s41597-022-01127-6} \\
            NADT-Prostate\cite{wilkinson2021nascent} & \href{https://www.sciencedirect.com/science/article/pii/S0302283821002074?via%3Dihub}{https://www.medrxiv.org/content/10.1101/2020.09.29.20199711v1.full} \\
            POST-NAT-BRCA & \href{https://onlinelibrary.wiley.com/doi/10.1002/cyto.a.23244}{https://onlinelibrary.wiley.com/doi/10.1002/cyto.a.23244} \\
            BOEHMK & \href{https://www.synapse.org/Synapse:syn25946117/wiki/611576}{https://www.synapse.org/Synapse:syn25946117/wiki/611576}  \\
            MBC &  \href{https://www.synapse.org/Synapse:syn59490671/wiki/628046}{https://www.synapse.org/Synapse:syn59490671/wiki/628046} \\
            SURGEN & \href{https://www.ebi.ac.uk/biostudies/bioimages/studies/S-BIAD1285}{https://www.ebi.ac.uk/biostudies/bioimages/studies/S-BIAD1285}  \\
            \bottomrule
            \end{tabular}
\label{tab:downstream-links}
\end{table}

\begin{table}[h]
\centering
\caption{\textbf{Performance comparison between $\ours$ and baselines on MGB Breast tasks.} Best in \textbf{bold}, second best is \underline{underlined}. \textbf{Subtype}: This is a slide-level classification task evaluated using AUROC, mean and standard error reported over 5-fold cross-validation. \textbf{ER}: This is a patient-level classification task evaluated using AUROC, mean and standard error reported over 5-fold cross-validation. \textbf{PR}: This is a patient-level classification task evaluated using AUROC, mean and standard error reported over 5-fold cross-validation. \textbf{HER2}: This is a patient-level classification task evaluated using AUROC, mean and standard error reported over 5-fold cross-validation. }
\scalebox{0.675}{\begin{tabular}{ll|cccc}
\toprule
 & \multirow{2}{*}{Model} & \multicolumn{4}{c}{Tasks} \\ 
\cmidrule(lr){3-6}
& & Subtype ($\uparrow$) (n=1264) & ER ($\uparrow$) (n=874) & PR ($\uparrow$) (n=874) & HER2 ($\uparrow$) (n=816) \\
\midrule
\multirow{8}{*}{\rotatebox{90}{\scriptsize{\textbf{Linear Probe}}}}
& \virchow\cite{vorontsov2024foundation} \mean & 0.971 ± 0.003 & 0.705 ± 0.016 & 0.672 ± 0.012 & 0.599 ± 0.020 \\
& \gigapath\cite{xu2024whole} \mean & 0.979 ± 0.002 & 0.711 ± 0.012 & 0.682 ± 0.018 & 0.575 ± 0.014 \\
& \chief\cite{wang2024chief} \mean & 0.955 ± 0.008 & 0.713 ± 0.024 & 0.721 ± 0.016 & 0.637 ± 0.019 \\
& \conch\cite{} \mean & 0.973 ± 0.005 & 0.728 ± 0.026 & 0.720 ± 0.012 & 0.656 ± 0.042 \\
& \prism\cite{shaikovski2024prism} & 0.985 ± 0.002 & 0.727 ± 0.022 & 0.639 ± 0.001 & \underline{0.710 ± 0.021} \\
& \gigapath\cite{xu2024whole} & 0.975 ± 0.003 & 0.708 ± 0.014 & 0.689 ± 0.020 & 0.597 ± 0.010 \\
& \chief\cite{wang2024chief} & 0.978 ± 0.004 & 0.749 ± 0.026 & 0.732 ± 0.020 & 0.696 ± 0.018 \\
& $\ours$ & 0.983 ± 0.004 & \textbf{0.784 ± 0.016} & \textbf{0.748 ± 0.021} & 0.694 ± 0.031 \\
\midrule
\multirow{5}{*}{\rotatebox{90}{\scriptsize{\textbf{Supervised}}}}
& ABMIL & 0.983 ± 0.003 & 0.700 ± 0.024 & 0.694 ± 0.005 & 0.648 ± 0.016 \\
& \gigapath\cite{xu2024whole} & \textbf{0.989 ± 0.001} & 0.747 ± 0.018 & 0.730 ± 0.016 & 0.663 ± 0.019 \\
& \chief\cite{wang2024chief} & 0.951 ± 0.005 & 0.696 ± 0.019 & 0.675 ± 0.017 & 0.567 ± 0.019 \\
& $\ours$ Random Init & \underline{0.986 ± 0.003} & \underline{0.771 ± 0.021} & \underline{0.740 ± 0.014} & 0.685 ± 0.041 \\
& $\ours$ & \underline{0.986 ± 0.002} & \underline{0.771 ± 0.025} & 0.739 ± 0.022 & \textbf{0.719 ± 0.030} \\
\midrule
\bottomrule
\end{tabular}
}\label{tab:allshot_mgb_brca}
\end{table}

\begin{table}[h]
\centering
\caption{\textbf{Performance comparison between $\ours$ and baselines on MGB Lung tasks.} Best in \textbf{bold}, second best is \underline{underlined}. \textbf{Subtype}: This is a slide-level classification task evaluated using AUROC, mean and standard error reported over 5-fold cross-validation. \textbf{TTF-1}: This is a patient-level classification task evaluated using AUROC, mean and standard error reported over 5-fold cross-validation. \textbf{P40}: This is a patient-level classification task evaluated using AUROC, mean and standard error reported over 5-fold cross-validation. \textbf{P63}: This is a patient-level classification task evaluated using AUROC, mean and standard error reported over 5-fold cross-validation. \textbf{Napsina}: This is a patient-level classification task evaluated using AUROC, mean and standard error reported over 5-fold cross-validation. \textbf{CDX-2}: This is a patient-level classification task evaluated using AUROC, mean and standard error reported over 5-fold cross-validation. \textbf{CK5-6}: This is a patient-level classification task evaluated using AUROC, mean and standard error reported over 5-fold cross-validation. }
\scalebox{0.59}{\begin{tabular}{ll|ccccccc}
\toprule
 & \multirow{2}{*}{Model} & \multicolumn{7}{c}{Tasks} \\ 
\cmidrule(lr){3-9}
& & Subtype ($\uparrow$) (n=1939) & TTF-1 ($\uparrow$) (n=488) & P40 ($\uparrow$) (n=185) & P63 ($\uparrow$) (n=153) & Napsina ($\uparrow$) (n=126) & CDX-2 ($\uparrow$) (n=79) & CK5-6 ($\uparrow$) (n=58) \\
\midrule
\multirow{8}{*}{\rotatebox{90}{\scriptsize{\textbf{Linear Probe}}}}
& \virchow\cite{vorontsov2024foundation} \mean & 0.960 ± 0.002 & 0.837 ± 0.017 & 0.750 ± 0.033 & 0.640 ± 0.036 & 0.678 ± 0.050 & 0.529 ± 0.050 & 0.742 ± 0.051 \\
& \gigapath\cite{xu2024whole} \mean & 0.969 ± 0.002 & 0.866 ± 0.008 & 0.739 ± 0.019 & 0.681 ± 0.027 & 0.693 ± 0.031 & 0.506 ± 0.092 & 0.697 ± 0.050 \\
& \chief\cite{wang2024chief} \mean & 0.950 ± 0.002 & 0.750 ± 0.018 & 0.627 ± 0.013 & 0.642 ± 0.029 & 0.593 ± 0.024 & 0.487 ± 0.074 & 0.634 ± 0.087 \\
& \conch\cite{} \mean & 0.969 ± 0.003 & 0.836 ± 0.009 & 0.867 ± 0.015 & 0.748 ± 0.034 & 0.773 ± 0.044 & 0.588 ± 0.073 & \underline{0.884 ± 0.047} \\
& \prism\cite{shaikovski2024prism} & 0.978 ± 0.005 & 0.858 ± 0.014 & 0.863 ± 0.017 & 0.809 ± 0.013 & 0.704 ± 0.015 & \textbf{0.738 ± 0.032} & 0.877 ± 0.022 \\
& \gigapath\cite{xu2024whole} & 0.957 ± 0.003 & 0.839 ± 0.013 & 0.696 ± 0.022 & 0.653 ± 0.034 & 0.634 ± 0.041 & 0.485 ± 0.093 & 0.670 ± 0.065 \\
& \chief\cite{wang2024chief} & 0.979 ± 0.003 & 0.822 ± 0.020 & 0.809 ± 0.039 & 0.792 ± 0.026 & 0.662 ± 0.022 & 0.558 ± 0.051 & 0.791 ± 0.049 \\
& $\ours$ & 0.982 ± 0.004 & \textbf{0.895 ± 0.011} & \underline{0.898 ± 0.027} & \textbf{0.879 ± 0.020} & 0.817 ± 0.033 & \underline{0.725 ± 0.062} & \textbf{0.933 ± 0.026} \\
\midrule
\multirow{5}{*}{\rotatebox{90}{\scriptsize{\textbf{Supervised}}}}
& ABMIL & 0.980 ± 0.002 & 0.866 ± 0.016 & 0.860 ± 0.027 & 0.811 ± 0.023 & \textbf{0.851 ± 0.013} & 0.599 ± 0.047 & 0.813 ± 0.047 \\
& \gigapath\cite{xu2024whole} & \textbf{0.988 ± 0.002} & 0.863 ± 0.014 & 0.642 ± 0.042 & 0.664 ± 0.031 & 0.624 ± 0.036 & 0.475 ± 0.041 & 0.676 ± 0.088 \\
& \chief\cite{wang2024chief} & 0.951 ± 0.009 & 0.682 ± 0.018 & 0.632 ± 0.049 & 0.525 ± 0.031 & 0.510 ± 0.074 & 0.498 ± 0.090 & 0.693 ± 0.068 \\
& $\ours$ Random Init & \underline{0.987 ± 0.002} & 0.882 ± 0.008 & 0.870 ± 0.023 & 0.796 ± 0.046 & 0.798 ± 0.020 & 0.610 ± 0.072 & 0.803 ± 0.061 \\
& $\ours$ & 0.985 ± 0.001 & \underline{0.887 ± 0.010} & \textbf{0.903 ± 0.025} & \underline{0.859 ± 0.026} & \underline{0.825 ± 0.018} & 0.672 ± 0.031 & 0.880 ± 0.031 \\
\midrule
\bottomrule
\end{tabular}
}\label{tab:allshot_mgb_lung}
\end{table}

\begin{table}[h]
\centering
\caption{\textbf{Performance comparison between $\ours$ and baselines on BCNB\cite{xu2021predicting} tasks.} Best in \textbf{bold}, second best is \underline{underlined}. \textbf{ER}: This is a patient-level classification task evaluated using AUROC, mean and standard error reported over 5-fold cross-validation. \textbf{PR}: This is a patient-level classification task evaluated using AUROC, mean and standard error reported over 5-fold cross-validation. \textbf{HER2}: This is a patient-level classification task evaluated using AUROC, mean and standard error reported over 5-fold cross-validation. }
\scalebox{0.675}{\begin{tabular}{ll|ccc}
\toprule
 & \multirow{2}{*}{Model} & \multicolumn{3}{c}{Tasks} \\ 
\cmidrule(lr){3-5}
& & ER ($\uparrow$) (n=1058) & PR ($\uparrow$) (n=1058) & HER2 ($\uparrow$) (n=1058) \\
\midrule
\multirow{8}{*}{\rotatebox{90}{\scriptsize{\textbf{Linear Probe}}}}
& \virchow\cite{vorontsov2024foundation} \mean & 0.888 ± 0.007 & 0.813 ± 0.010 & 0.707 ± 0.011 \\
& \gigapath\cite{xu2024whole} \mean & 0.901 ± 0.013 & 0.828 ± 0.014 & 0.719 ± 0.015 \\
& \chief\cite{wang2024chief} \mean & 0.852 ± 0.014 & 0.799 ± 0.021 & 0.703 ± 0.016 \\
& \conch\cite{} \mean & 0.889 ± 0.010 & 0.810 ± 0.022 & 0.737 ± 0.013 \\
& \prism\cite{shaikovski2024prism} & 0.892 ± 0.014 & 0.815 ± 0.019 & 0.711 ± 0.010 \\
& \gigapath\cite{xu2024whole} & 0.886 ± 0.016 & 0.811 ± 0.013 & 0.702 ± 0.016 \\
& \chief\cite{wang2024chief} & 0.883 ± 0.014 & 0.818 ± 0.017 & 0.719 ± 0.021 \\
& $\ours$ & 0.921 ± 0.009 & 0.837 ± 0.020 & 0.765 ± 0.008 \\
\midrule
\multirow{5}{*}{\rotatebox{90}{\scriptsize{\textbf{Supervised}}}}
& ABMIL & 0.877 ± 0.009 & 0.804 ± 0.017 & 0.737 ± 0.011 \\
& \gigapath\cite{xu2024whole} & \underline{0.925 ± 0.012} & \underline{0.856 ± 0.016} & 0.766 ± 0.016 \\
& \chief\cite{wang2024chief} & 0.786 ± 0.007 & 0.753 ± 0.013 & 0.658 ± 0.029 \\
& $\ours$ Random Init & \textbf{0.926 ± 0.010} & \textbf{0.859 ± 0.019} & \underline{0.780 ± 0.008} \\
& $\ours$ & 0.919 ± 0.011 & 0.848 ± 0.016 & \textbf{0.786 ± 0.007} \\
\midrule
\bottomrule
\end{tabular}
}\label{tab:allshot_bcnb}
\end{table}

\begin{table}[h]
\centering
\caption{\textbf{Performance comparison between $\ours$ and baselines on MUT-HET-RCC\cite{} tasks.} Best in \textbf{bold}, second best is \underline{underlined}. \textbf{BAP1 Mutation}: This is a patient-level classification task evaluated using AUROC, mean and standard error reported over 5-fold cross-validation. \textbf{PBRM1 Mutation}: This is a slide-level classification task evaluated using AUROC, mean and standard error reported over 5-fold cross-validation. \textbf{SETD2 Mutation}: This is a slide-level classification task evaluated using AUROC, mean and standard error reported over 5-fold cross-validation. }
\scalebox{0.675}{\begin{tabular}{ll|ccc}
\toprule
 & \multirow{2}{*}{Model} & \multicolumn{3}{c}{Tasks} \\ 
\cmidrule(lr){3-5}
& & BAP1 Mutation ($\uparrow$) (n=1291) & PBRM1 Mutation ($\uparrow$) (n=1291) & SETD2 Mutation ($\uparrow$) (n=1291) \\
\midrule
\multirow{8}{*}{\rotatebox{90}{\scriptsize{\textbf{Linear Probe}}}}
& \virchow\cite{vorontsov2024foundation} \mean & 0.879 ± 0.003 & 0.810 ± 0.008 & 0.712 ± 0.012 \\
& \gigapath\cite{xu2024whole} \mean & 0.880 ± 0.007 & 0.806 ± 0.007 & 0.701 ± 0.012 \\
& \chief\cite{wang2024chief} \mean & 0.850 ± 0.015 & 0.746 ± 0.003 & 0.722 ± 0.007 \\
& \conch\cite{} \mean & 0.846 ± 0.011 & 0.782 ± 0.005 & 0.728 ± 0.011 \\
& \prism\cite{shaikovski2024prism} & 0.848 ± 0.017 & 0.797 ± 0.008 & 0.734 ± 0.009 \\
& \gigapath\cite{xu2024whole} & 0.857 ± 0.007 & 0.799 ± 0.003 & 0.708 ± 0.013 \\
& \chief\cite{wang2024chief} & 0.852 ± 0.013 & 0.781 ± 0.011 & \underline{0.743 ± 0.012} \\
& $\ours$ & 0.873 ± 0.006 & \textbf{0.826 ± 0.003} & \textbf{0.756 ± 0.009} \\
\midrule
\multirow{5}{*}{\rotatebox{90}{\scriptsize{\textbf{Supervised}}}}
& ABMIL & 0.858 ± 0.015 & 0.778 ± 0.011 & 0.717 ± 0.008 \\
& \gigapath\cite{xu2024whole} & \underline{0.885 ± 0.005} & \underline{0.816 ± 0.011} & 0.727 ± 0.009 \\
& \chief\cite{wang2024chief} & 0.802 ± 0.025 & 0.697 ± 0.005 & 0.669 ± 0.019 \\
& $\ours$ Random Init & 0.868 ± 0.015 & 0.779 ± 0.005 & 0.729 ± 0.006 \\
& $\ours$ & \textbf{0.898 ± 0.005} & 0.807 ± 0.011 & 0.738 ± 0.007 \\
\midrule
\bottomrule
\end{tabular}
}\label{tab:allshot_mut-het-rcc}
\end{table}

\begin{table}[h]
\centering
\caption{\textbf{Performance comparison between $\ours$ and baselines on IMP\cite{neto2024interpretable} tasks.} Best in \textbf{bold}, second best is \underline{underlined}. \textbf{Grade}: This is a slide-level classification task evaluated using quadratic weighted kappa, mean and 95\% CI reported over 100 bootstraps of a single fold. }
\scalebox{0.675}{\begin{tabular}{ll|c}
\toprule
 & \multirow{2}{*}{Model} & \multicolumn{1}{c}{Tasks} \\ 
\cmidrule(lr){3-3}
& & Grade ($\uparrow$) (n=5333) \\
\midrule
\multirow{8}{*}{\rotatebox{90}{\scriptsize{\textbf{Linear Probe}}}}
& \virchow\cite{vorontsov2024foundation} \mean & 0.913 (0.894-0.929) \\
& \gigapath\cite{xu2024whole} \mean & 0.903 (0.884-0.921) \\
& \chief\cite{wang2024chief} \mean & 0.877 (0.847-0.908) \\
& \conch\cite{} \mean & 0.889 (0.862-0.912) \\
& \prism\cite{shaikovski2024prism} & 0.935 (0.919-0.951) \\
& \gigapath\cite{xu2024whole} & 0.903 (0.880-0.925) \\
& \chief\cite{wang2024chief} & 0.914 (0.893-0.937) \\
& $\ours$ & 0.919 (0.896-0.937) \\
\midrule
\multirow{5}{*}{\rotatebox{90}{\scriptsize{\textbf{Supervised}}}}
& ABMIL & 0.942 (0.926-0.957) \\
& \gigapath\cite{xu2024whole} & \textbf{0.956 (0.941-0.968)} \\
& \chief\cite{wang2024chief} & 0.917 (0.893-0.937) \\
& $\ours$ Random Init & 0.944 (0.922-0.964) \\
& $\ours$ & \underline{0.946 (0.929-0.963)} \\
\midrule
\bottomrule
\end{tabular}
}\label{tab:allshot_imp}
\end{table}

\begin{table}[h]
\centering
\caption{\textbf{Performance comparison between $\ours$ and baselines on PANDA\cite{bulten2022artificial} tasks.} Best in \textbf{bold}, second best is \underline{underlined}. \textbf{ISUP Grade}: This is a slide-level classification task evaluated using quadratic weighted kappa, mean and 95\% CI reported over 100 bootstraps of a single fold. }
\scalebox{0.675}{\begin{tabular}{ll|c}
\toprule
 & \multirow{2}{*}{Model} & \multicolumn{1}{c}{Tasks} \\ 
\cmidrule(lr){3-3}
& & ISUP Grade ($\uparrow$) (n=9555) \\
\midrule
\multirow{8}{*}{\rotatebox{90}{\scriptsize{\textbf{Linear Probe}}}}
& \virchow\cite{vorontsov2024foundation} \mean & 0.895 (0.875-0.913) \\
& \gigapath\cite{xu2024whole} \mean & 0.894 (0.871-0.915) \\
& \chief\cite{wang2024chief} \mean & 0.799 (0.765-0.827) \\
& \conch\cite{} \mean & 0.845 (0.814-0.864) \\
& \prism\cite{shaikovski2024prism} & 0.919 (0.901-0.935) \\
& \gigapath\cite{xu2024whole} & 0.873 (0.850-0.895) \\
& \chief\cite{wang2024chief} & 0.898 (0.878-0.918) \\
& $\ours$ & 0.915 (0.899-0.929) \\
\midrule
\multirow{5}{*}{\rotatebox{90}{\scriptsize{\textbf{Supervised}}}}
& ABMIL & \underline{0.932 (0.919-0.944)} \\
& \gigapath\cite{xu2024whole} & \textbf{0.959 (0.951-0.967)} \\
& \chief\cite{wang2024chief} & 0.798 (0.765-0.830) \\
& $\ours$ Random Init & 0.926 (0.913-0.938) \\
& $\ours$ & 0.930 (0.917-0.943) \\
\midrule
\bottomrule
\end{tabular}
}\label{tab:allshot_panda}
\end{table}

\begin{table}[h]
\centering
\caption{\textbf{Performance comparison between $\ours$ and baselines on CPTAC-BRCA\cite{edwards2015cptac} tasks.} Best in \textbf{bold}, second best is \underline{underlined}. \textbf{PIK3CA Mutation}: This is a patient-level classification task evaluated using AUROC, mean and standard error reported over 50-fold Monte Carlo. \textbf{TP53 Mutation}: This is a patient-level classification task evaluated using AUROC, mean and standard error reported over 50-fold Monte Carlo. }
\scalebox{0.675}{\begin{tabular}{ll|cc}
\toprule
 & \multirow{2}{*}{Model} & \multicolumn{2}{c}{Tasks} \\ 
\cmidrule(lr){3-4}
& & PIK3CA Mutation ($\uparrow$) (n=103) & TP53 Mutation ($\uparrow$) (n=103) \\
\midrule
\multirow{8}{*}{\rotatebox{90}{\scriptsize{\textbf{Linear Probe}}}}
& \virchow\cite{vorontsov2024foundation} \mean & 0.569 ± 0.015 & 0.766 ± 0.012 \\
& \gigapath\cite{xu2024whole} \mean & 0.554 ± 0.016 & 0.754 ± 0.012 \\
& \chief\cite{wang2024chief} \mean & \underline{0.615 ± 0.015} & 0.786 ± 0.012 \\
& \conch\cite{} \mean & 0.513 ± 0.017 & 0.796 ± 0.011 \\
& \prism\cite{shaikovski2024prism} & 0.575 ± 0.014 & 0.787 ± 0.013 \\
& \gigapath\cite{xu2024whole} & 0.531 ± 0.017 & 0.749 ± 0.012 \\
& \chief\cite{wang2024chief} & \textbf{0.647 ± 0.013} & 0.832 ± 0.012 \\
& $\ours$ & 0.571 ± 0.017 & \textbf{0.876 ± 0.012} \\
\midrule
\multirow{5}{*}{\rotatebox{90}{\scriptsize{\textbf{Supervised}}}}
& ABMIL & 0.531 ± 0.017 & 0.791 ± 0.012 \\
& \gigapath\cite{xu2024whole} & 0.570 ± 0.016 & 0.775 ± 0.014 \\
& \chief\cite{wang2024chief} & 0.504 ± 0.023 & 0.620 ± 0.020 \\
& $\ours$ Random Init & 0.578 ± 0.016 & 0.833 ± 0.012 \\
& $\ours$ & 0.611 ± 0.015 & \underline{0.846 ± 0.012} \\
\midrule
\bottomrule
\end{tabular}
}\label{tab:allshot_cptac_brca}
\end{table}

\begin{table}[h]
\centering
\caption{\textbf{Performance comparison between $\ours$ and baselines on CPTAC-CCRCC\cite{edwards2015cptac} tasks.} Best in \textbf{bold}, second best is \underline{underlined}. \textbf{BAP1 Mutation}: This is a patient-level classification task evaluated using AUROC, mean and standard error reported over 50-fold Monte Carlo. \textbf{PBRM1 Mutation}: This is a patient-level classification task evaluated using AUROC, mean and standard error reported over 50-fold Monte Carlo. }
\scalebox{0.675}{\begin{tabular}{ll|cc}
\toprule
 & \multirow{2}{*}{Model} & \multicolumn{2}{c}{Tasks} \\ 
\cmidrule(lr){3-4}
& & BAP1 Mutation ($\uparrow$) (n=103) & PBRM1 Mutation ($\uparrow$) (n=103) \\
\midrule
\multirow{8}{*}{\rotatebox{90}{\scriptsize{\textbf{Linear Probe}}}}
& \virchow\cite{vorontsov2024foundation} \mean & 0.624 ± 0.022 & 0.501 ± 0.016 \\
& \gigapath\cite{xu2024whole} \mean & 0.688 ± 0.023 & 0.482 ± 0.015 \\
& \chief\cite{wang2024chief} \mean & 0.717 ± 0.018 & 0.457 ± 0.016 \\
& \conch\cite{} \mean & 0.655 ± 0.016 & 0.518 ± 0.015 \\
& \prism\cite{shaikovski2024prism} & 0.664 ± 0.022 & 0.598 ± 0.019 \\
& \gigapath\cite{xu2024whole} & 0.670 ± 0.023 & 0.466 ± 0.016 \\
& \chief\cite{wang2024chief} & 0.745 ± 0.019 & 0.522 ± 0.015 \\
& $\ours$ & \textbf{0.809 ± 0.014} & \textbf{0.668 ± 0.016} \\
\midrule
\multirow{5}{*}{\rotatebox{90}{\scriptsize{\textbf{Supervised}}}}
& ABMIL & 0.729 ± 0.020 & 0.576 ± 0.018 \\
& \gigapath\cite{xu2024whole} & 0.691 ± 0.020 & 0.512 ± 0.017 \\
& \chief\cite{wang2024chief} & 0.641 ± 0.024 & 0.514 ± 0.018 \\
& $\ours$ Random Init & 0.772 ± 0.015 & 0.595 ± 0.018 \\
& $\ours$ & \underline{0.802 ± 0.015} & \underline{0.629 ± 0.017} \\
\midrule
\bottomrule
\end{tabular}
}\label{tab:allshot_cptac_ccrcc}
\end{table}

\begin{table}[h]
\centering
\caption{\textbf{Performance comparison between $\ours$ and baselines on CPTAC-COAD\cite{edwards2015cptac} tasks.} Best in \textbf{bold}, second best is \underline{underlined}. \textbf{KRAS Mutation}: This is a patient-level classification task evaluated using AUROC, mean and standard error reported over 50-fold Monte Carlo. \textbf{TP53 Mutation}: This is a patient-level classification task evaluated using AUROC, mean and standard error reported over 50-fold Monte Carlo. }
\scalebox{0.675}{\begin{tabular}{ll|cc}
\toprule
 & \multirow{2}{*}{Model} & \multicolumn{2}{c}{Tasks} \\ 
\cmidrule(lr){3-4}
& & KRAS Mutation ($\uparrow$) (n=94) & TP53 Mutation ($\uparrow$) (n=94) \\
\midrule
\multirow{8}{*}{\rotatebox{90}{\scriptsize{\textbf{Linear Probe}}}}
& \virchow\cite{vorontsov2024foundation} \mean & 0.661 ± 0.015 & 0.684 ± 0.017 \\
& \gigapath\cite{xu2024whole} \mean & 0.642 ± 0.015 & 0.656 ± 0.018 \\
& \chief\cite{wang2024chief} \mean & 0.652 ± 0.015 & 0.649 ± 0.019 \\
& \conch\cite{} \mean & \textbf{0.742 ± 0.016} & 0.690 ± 0.018 \\
& \prism\cite{shaikovski2024prism} & 0.554 ± 0.015 & 0.578 ± 0.019 \\
& \gigapath\cite{xu2024whole} & 0.654 ± 0.013 & 0.642 ± 0.016 \\
& \chief\cite{wang2024chief} & 0.649 ± 0.016 & 0.659 ± 0.018 \\
& $\ours$ & \underline{0.704 ± 0.014} & \underline{0.742 ± 0.016} \\
\midrule
\multirow{5}{*}{\rotatebox{90}{\scriptsize{\textbf{Supervised}}}}
& ABMIL & 0.623 ± 0.017 & 0.730 ± 0.016 \\
& \gigapath\cite{xu2024whole} & 0.622 ± 0.015 & 0.648 ± 0.017 \\
& \chief\cite{wang2024chief} & 0.531 ± 0.017 & 0.526 ± 0.019 \\
& $\ours$ Random Init & 0.696 ± 0.015 & 0.698 ± 0.018 \\
& $\ours$ & 0.670 ± 0.016 & \textbf{0.785 ± 0.014} \\
\midrule
\bottomrule
\end{tabular}
}\label{tab:allshot_cptac_coad}
\end{table}

\begin{table}[h]
\centering
\caption{\textbf{Performance comparison between $\ours$ and baselines on CPTAC-GBM\cite{edwards2015cptac} tasks.} Best in \textbf{bold}, second best is \underline{underlined}. \textbf{EGFR Mutation}: This is a patient-level classification task evaluated using AUROC, mean and standard error reported over 50-fold Monte Carlo. \textbf{TP53 Mutation}: This is a patient-level classification task evaluated using AUROC, mean and standard error reported over 50-fold Monte Carlo. }
\scalebox{0.675}{\begin{tabular}{ll|cc}
\toprule
 & \multirow{2}{*}{Model} & \multicolumn{2}{c}{Tasks} \\ 
\cmidrule(lr){3-4}
& & EGFR Mutation ($\uparrow$) (n=99) & TP53 Mutation ($\uparrow$) (n=99) \\
\midrule
\multirow{8}{*}{\rotatebox{90}{\scriptsize{\textbf{Linear Probe}}}}
& \virchow\cite{vorontsov2024foundation} \mean & 0.509 ± 0.017 & 0.816 ± 0.012 \\
& \gigapath\cite{xu2024whole} \mean & 0.623 ± 0.016 & 0.805 ± 0.012 \\
& \chief\cite{wang2024chief} \mean & 0.662 ± 0.015 & 0.849 ± 0.011 \\
& \conch\cite{} \mean & 0.639 ± 0.014 & 0.828 ± 0.014 \\
& \prism\cite{shaikovski2024prism} & 0.619 ± 0.015 & 0.858 ± 0.015 \\
& \gigapath\cite{xu2024whole} & 0.634 ± 0.016 & 0.785 ± 0.014 \\
& \chief\cite{wang2024chief} & 0.743 ± 0.015 & \underline{0.862 ± 0.010} \\
& $\ours$ & \underline{0.782 ± 0.011} & 0.842 ± 0.013 \\
\midrule
\multirow{5}{*}{\rotatebox{90}{\scriptsize{\textbf{Supervised}}}}
& ABMIL & 0.713 ± 0.016 & 0.836 ± 0.013 \\
& \gigapath\cite{xu2024whole} & 0.624 ± 0.014 & 0.698 ± 0.015 \\
& \chief\cite{wang2024chief} & 0.480 ± 0.020 & 0.519 ± 0.021 \\
& $\ours$ Random Init & 0.674 ± 0.012 & 0.832 ± 0.016 \\
& $\ours$ & \textbf{0.791 ± 0.010} & \textbf{0.864 ± 0.012} \\
\midrule
\bottomrule
\end{tabular}
}\label{tab:allshot_cptac_gbm}
\end{table}

\begin{table}[h]
\centering
\caption{\textbf{Performance comparison between $\ours$ and baselines on CPTAC-HNSC\cite{edwards2015cptac} tasks.} Best in \textbf{bold}, second best is \underline{underlined}. \textbf{CASP8 Mutation}: This is a patient-level classification task evaluated using AUROC, mean and standard error reported over 50-fold Monte Carlo. }
\scalebox{0.675}{\begin{tabular}{ll|c}
\toprule
 & \multirow{2}{*}{Model} & \multicolumn{1}{c}{Tasks} \\ 
\cmidrule(lr){3-3}
& & CASP8 Mutation ($\uparrow$) (n=107) \\
\midrule
\multirow{8}{*}{\rotatebox{90}{\scriptsize{\textbf{Linear Probe}}}}
& \virchow\cite{vorontsov2024foundation} \mean & 0.397 ± 0.025 \\
& \gigapath\cite{xu2024whole} \mean & 0.497 ± 0.028 \\
& \chief\cite{wang2024chief} \mean & 0.482 ± 0.025 \\
& \conch\cite{} \mean & 0.614 ± 0.028 \\
& \prism\cite{shaikovski2024prism} & 0.601 ± 0.027 \\
& \gigapath\cite{xu2024whole} & 0.474 ± 0.025 \\
& \chief\cite{wang2024chief} & 0.493 ± 0.027 \\
& $\ours$ & \textbf{0.754 ± 0.019} \\
\midrule
\multirow{5}{*}{\rotatebox{90}{\scriptsize{\textbf{Supervised}}}}
& ABMIL & 0.681 ± 0.029 \\
& \gigapath\cite{xu2024whole} & 0.619 ± 0.029 \\
& \chief\cite{wang2024chief} & 0.558 ± 0.035 \\
& $\ours$ Random Init & 0.673 ± 0.022 \\
& $\ours$ & \underline{0.736 ± 0.023} \\
\midrule
\bottomrule
\end{tabular}
}\label{tab:allshot_cptac_hnsc}
\end{table}

\begin{table}[h]
\centering
\caption{\textbf{Performance comparison between $\ours$ and baselines on CPTAC-LSCC\cite{edwards2015cptac} tasks.} Best in \textbf{bold}, second best is \underline{underlined}. \textbf{KEAP1 Mutation}: This is a patient-level classification task evaluated using AUROC, mean and standard error reported over 50-fold Monte Carlo. \textbf{ARID1A Mutation}: This is a patient-level classification task evaluated using AUROC, mean and standard error reported over 50-fold Monte Carlo. }
\scalebox{0.675}{\begin{tabular}{ll|cc}
\toprule
 & \multirow{2}{*}{Model} & \multicolumn{2}{c}{Tasks} \\ 
\cmidrule(lr){3-4}
& & KEAP1 Mutation ($\uparrow$) (n=108) & ARID1A Mutation ($\uparrow$) (n=108) \\
\midrule
\multirow{8}{*}{\rotatebox{90}{\scriptsize{\textbf{Linear Probe}}}}
& \virchow\cite{vorontsov2024foundation} \mean & 0.671 ± 0.020 & 0.417 ± 0.020 \\
& \gigapath\cite{xu2024whole} \mean & 0.640 ± 0.017 & 0.437 ± 0.019 \\
& \chief\cite{wang2024chief} \mean & \underline{0.676 ± 0.016} & 0.524 ± 0.022 \\
& \conch\cite{} \mean & 0.574 ± 0.018 & 0.411 ± 0.020 \\
& \prism\cite{shaikovski2024prism} & 0.492 ± 0.018 & 0.449 ± 0.020 \\
& \gigapath\cite{xu2024whole} & 0.663 ± 0.018 & 0.465 ± 0.016 \\
& \chief\cite{wang2024chief} & 0.656 ± 0.016 & 0.535 ± 0.024 \\
& $\ours$ & \textbf{0.685 ± 0.019} & \textbf{0.658 ± 0.023} \\
\midrule
\multirow{5}{*}{\rotatebox{90}{\scriptsize{\textbf{Supervised}}}}
& ABMIL & 0.459 ± 0.023 & 0.432 ± 0.023 \\
& \gigapath\cite{xu2024whole} & 0.614 ± 0.022 & \underline{0.539 ± 0.022} \\
& \chief\cite{wang2024chief} & 0.446 ± 0.026 & 0.467 ± 0.024 \\
& $\ours$ Random Init & 0.629 ± 0.018 & 0.487 ± 0.020 \\
& $\ours$ & 0.608 ± 0.017 & 0.514 ± 0.022 \\
\midrule
\bottomrule
\end{tabular}
}\label{tab:allshot_cptac_lscc}
\end{table}

\begin{table}[h]
\centering
\caption{\textbf{Performance comparison between $\ours$ and baselines on CPTAC-LUAD\cite{edwards2015cptac} tasks.} Best in \textbf{bold}, second best is \underline{underlined}. \textbf{EGFR Mutation}: This is a patient-level classification task evaluated using AUROC, mean and standard error reported over 50-fold Monte Carlo. \textbf{STK11 Mutation}: This is a patient-level classification task evaluated using AUROC, mean and standard error reported over 50-fold Monte Carlo. \textbf{TP53 Mutation}: This is a patient-level classification task evaluated using AUROC, mean and standard error reported over 50-fold Monte Carlo. }
\scalebox{0.675}{\begin{tabular}{ll|ccc}
\toprule
 & \multirow{2}{*}{Model} & \multicolumn{3}{c}{Tasks} \\ 
\cmidrule(lr){3-5}
& & EGFR Mutation ($\uparrow$) (n=108) & STK11 Mutation ($\uparrow$) (n=108) & TP53 Mutation ($\uparrow$) (n=108) \\
\midrule
\multirow{8}{*}{\rotatebox{90}{\scriptsize{\textbf{Linear Probe}}}}
& \virchow\cite{vorontsov2024foundation} \mean & 0.791 ± 0.015 & 0.864 ± 0.013 & 0.734 ± 0.014 \\
& \gigapath\cite{xu2024whole} \mean & 0.782 ± 0.014 & 0.852 ± 0.013 & 0.715 ± 0.014 \\
& \chief\cite{wang2024chief} \mean & 0.722 ± 0.017 & 0.765 ± 0.019 & 0.663 ± 0.014 \\
& \conch\cite{} \mean & 0.789 ± 0.014 & 0.824 ± 0.018 & 0.682 ± 0.015 \\
& \prism\cite{shaikovski2024prism} & \underline{0.809 ± 0.013} & 0.854 ± 0.015 & \textbf{0.755 ± 0.013} \\
& \gigapath\cite{xu2024whole} & 0.791 ± 0.014 & 0.822 ± 0.015 & 0.737 ± 0.013 \\
& \chief\cite{wang2024chief} & 0.717 ± 0.017 & 0.824 ± 0.015 & 0.704 ± 0.014 \\
& $\ours$ & \textbf{0.822 ± 0.011} & \underline{0.889 ± 0.015} & \underline{0.752 ± 0.014} \\
\midrule
\multirow{5}{*}{\rotatebox{90}{\scriptsize{\textbf{Supervised}}}}
& ABMIL & 0.748 ± 0.014 & 0.856 ± 0.017 & 0.686 ± 0.015 \\
& \gigapath\cite{xu2024whole} & 0.749 ± 0.017 & 0.791 ± 0.019 & 0.699 ± 0.015 \\
& \chief\cite{wang2024chief} & 0.495 ± 0.025 & 0.517 ± 0.024 & 0.524 ± 0.017 \\
& $\ours$ Random Init & 0.805 ± 0.015 & 0.862 ± 0.016 & 0.704 ± 0.015 \\
& $\ours$ & 0.798 ± 0.014 & \textbf{0.891 ± 0.011} & \underline{0.752 ± 0.014} \\
\midrule
\bottomrule
\end{tabular}
}\label{tab:allshot_cptac_luad}
\end{table}

\begin{table}[h]
\centering
\caption{\textbf{Performance comparison between $\ours$ and baselines on CPTAC-PDAC\cite{edwards2015cptac} tasks.} Best in \textbf{bold}, second best is \underline{underlined}. \textbf{SMAD4 Mutation}: This is a patient-level classification task evaluated using AUROC, mean and standard error reported over 50-fold Monte Carlo. }
\scalebox{0.675}{\begin{tabular}{ll|c}
\toprule
 & \multirow{2}{*}{Model} & \multicolumn{1}{c}{Tasks} \\ 
\cmidrule(lr){3-3}
& & SMAD4 Mutation ($\uparrow$) (n=105) \\
\midrule
\multirow{8}{*}{\rotatebox{90}{\scriptsize{\textbf{Linear Probe}}}}
& \virchow\cite{vorontsov2024foundation} \mean & 0.488 ± 0.020 \\
& \gigapath\cite{xu2024whole} \mean & 0.418 ± 0.020 \\
& \chief\cite{wang2024chief} \mean & 0.439 ± 0.018 \\
& \conch\cite{} \mean & 0.576 ± 0.019 \\
& \prism\cite{shaikovski2024prism} & 0.523 ± 0.022 \\
& \gigapath\cite{xu2024whole} & 0.423 ± 0.021 \\
& \chief\cite{wang2024chief} & 0.393 ± 0.018 \\
& $\ours$ & \underline{0.578 ± 0.024} \\
\midrule
\multirow{5}{*}{\rotatebox{90}{\scriptsize{\textbf{Supervised}}}}
& ABMIL & 0.512 ± 0.019 \\
& \gigapath\cite{xu2024whole} & 0.340 ± 0.018 \\
& \chief\cite{wang2024chief} & 0.478 ± 0.022 \\
& $\ours$ Random Init & \textbf{0.598 ± 0.020} \\
& $\ours$ & 0.576 ± 0.022 \\
\midrule
\bottomrule
\end{tabular}
}\label{tab:allshot_cptac_pda}
\end{table}

\begin{table}[h]
\centering
\caption{\textbf{Performance comparison between $\ours$ and baselines on BRACS\cite{brancati2021bracs} tasks.} Best in \textbf{bold}, second best is \underline{underlined}. \textbf{Fine Subtype}: This is a slide-level classification task evaluated using balanced accuracy, mean and standard error reported over 5-fold cross-validation. \textbf{Coarse Subtype}: This is a slide-level classification task evaluated using balanced accuracy, mean and standard error reported over 5-fold cross-validation. }
\scalebox{0.675}{\begin{tabular}{ll|cc}
\toprule
 & \multirow{2}{*}{Model} & \multicolumn{2}{c}{Tasks} \\ 
\cmidrule(lr){3-4}
& & Fine Subtype ($\uparrow$) (n=547) & Coarse Subtype ($\uparrow$) (n=547) \\
\midrule
\multirow{8}{*}{\rotatebox{90}{\scriptsize{\textbf{Linear Probe}}}}
& \virchow\cite{vorontsov2024foundation} \mean & 0.348 ± 0.015 & 0.618 ± 0.028 \\
& \gigapath\cite{xu2024whole} \mean & 0.318 ± 0.016 & 0.569 ± 0.031 \\
& \chief\cite{wang2024chief} \mean & 0.350 ± 0.029 & 0.621 ± 0.030 \\
& \conch\cite{} \mean & 0.378 ± 0.036 & 0.620 ± 0.036 \\
& \prism\cite{shaikovski2024prism} & 0.419 ± 0.014 & 0.659 ± 0.019 \\
& \gigapath\cite{xu2024whole} & 0.342 ± 0.019 & 0.606 ± 0.037 \\
& \chief\cite{wang2024chief} & 0.452 ± 0.015 & 0.704 ± 0.023 \\
& $\ours$ & \underline{0.481 ± 0.015} & 0.716 ± 0.019 \\
\midrule
\multirow{5}{*}{\rotatebox{90}{\scriptsize{\textbf{Supervised}}}}
& ABMIL & 0.478 ± 0.038 & \textbf{0.740 ± 0.023} \\
& \gigapath\cite{xu2024whole} & 0.444 ± 0.020 & 0.678 ± 0.018 \\
& \chief\cite{wang2024chief} & 0.228 ± 0.007 & 0.495 ± 0.039 \\
& $\ours$ Random Init & 0.456 ± 0.020 & 0.715 ± 0.020 \\
& $\ours$ & \textbf{0.492 ± 0.013} & \underline{0.731 ± 0.012} \\
\midrule
\bottomrule
\end{tabular}
}\label{tab:allshot_bracs}
\end{table}

\begin{table}[h]
\centering
\caption{\textbf{Performance comparison between $\ours$ and baselines on EBRAINS\cite{roetzer2022ebrains} tasks.} Best in \textbf{bold}, second best is \underline{underlined}. \textbf{Fine Subtype}: This is a slide-level classification task evaluated using balanced accuracy, mean and 95\% CI reported over 100 bootstraps of a single fold. \textbf{Coarse Subtype}: This is a slide-level classification task evaluated using balanced accuracy, mean and 95\% CI reported over 100 bootstraps of a single fold. }
\scalebox{0.675}{\begin{tabular}{ll|cc}
\toprule
 & \multirow{2}{*}{Model} & \multicolumn{2}{c}{Tasks} \\ 
\cmidrule(lr){3-4}
& & Fine Subtype ($\uparrow$) (n=2319) & Coarse Subtype ($\uparrow$) (n=2319) \\
\midrule
\multirow{8}{*}{\rotatebox{90}{\scriptsize{\textbf{Linear Probe}}}}
& \virchow\cite{vorontsov2024foundation} \mean & 0.708 (0.666-0.756) & 0.873 (0.841-0.913) \\
& \gigapath\cite{xu2024whole} \mean & 0.740 (0.700-0.776) & \underline{0.903 (0.871-0.941)} \\
& \chief\cite{wang2024chief} \mean & 0.637 (0.594-0.671) & 0.785 (0.730-0.832) \\
& \conch\cite{} \mean & 0.693 (0.646-0.733) & 0.870 (0.836-0.910) \\
& \prism\cite{shaikovski2024prism} & 0.721 (0.686-0.766) & 0.871 (0.823-0.908) \\
& \gigapath\cite{xu2024whole} & 0.722 (0.679-0.757) & 0.887 (0.849-0.928) \\
& \chief\cite{wang2024chief} & 0.660 (0.621-0.704) & 0.840 (0.793-0.893) \\
& $\ours$ & \underline{0.742 (0.707-0.775)} & \textbf{0.916 (0.873-0.951)} \\
\midrule
\multirow{5}{*}{\rotatebox{90}{\scriptsize{\textbf{Supervised}}}}
& ABMIL & 0.720 (0.684-0.758) & 0.900 (0.865-0.929) \\
& \gigapath\cite{xu2024whole} & \textbf{0.751 (0.708-0.787)} & 0.890 (0.856-0.923) \\
& \chief\cite{wang2024chief} & 0.033 (0.033-0.033) & 0.156 (0.150-0.163) \\
& $\ours$ Random Init & 0.737 (0.697-0.774) & 0.898 (0.858-0.933) \\
& $\ours$ & 0.734 (0.692-0.769) & 0.881 (0.838-0.920) \\
\midrule
\bottomrule
\end{tabular}
}\label{tab:allshot_ebrains}
\end{table}

\begin{table}[h]
\centering
\caption{\textbf{Performance comparison between $\ours$ and baselines on OV-Bevacizumab\cite{wang2022weakly} tasks.} Best in \textbf{bold}, second best is \underline{underlined}. \textbf{Response}: This is a patient-level classification task evaluated using AUROC, mean and standard error reported over 50-fold Monte Carlo. }
\scalebox{0.675}{\begin{tabular}{ll|c}
\toprule
 & \multirow{2}{*}{Model} & \multicolumn{1}{c}{Tasks} \\ 
\cmidrule(lr){3-3}
& & Response ($\uparrow$) (n=36) \\
\midrule
\multirow{8}{*}{\rotatebox{90}{\scriptsize{\textbf{Linear Probe}}}}
& \virchow\cite{vorontsov2024foundation} \mean & 0.607 ± 0.033 \\
& \gigapath\cite{xu2024whole} \mean & 0.523 ± 0.034 \\
& \chief\cite{wang2024chief} \mean & 0.667 ± 0.035 \\
& \conch\cite{} \mean & 0.623 ± 0.030 \\
& \prism\cite{shaikovski2024prism} & 0.433 ± 0.034 \\
& \gigapath\cite{xu2024whole} & 0.613 ± 0.040 \\
& \chief\cite{wang2024chief} & 0.620 ± 0.041 \\
& $\ours$ & \textbf{0.863 ± 0.021} \\
\midrule
\multirow{5}{*}{\rotatebox{90}{\scriptsize{\textbf{Supervised}}}}
& ABMIL & 0.553 ± 0.034 \\
& \gigapath\cite{xu2024whole} & 0.593 ± 0.043 \\
& \chief\cite{wang2024chief} & 0.453 ± 0.042 \\
& $\ours$ Random Init & 0.757 ± 0.027 \\
& $\ours$ & \underline{0.793 ± 0.029} \\
\midrule
\bottomrule
\end{tabular}
}\label{tab:allshot_ovarian}
\end{table}

\begin{table}[h]
\centering
\caption{\textbf{Performance comparison between $\ours$ and baselines on NADT Prostate\cite{wilkinson2021nascent} tasks.} Best in \textbf{bold}, second best is \underline{underlined}. \textbf{Response}: This is a patient-level classification task evaluated using AUROC, mean and standard error reported over 50-fold Monte Carlo. }
\scalebox{0.675}{\begin{tabular}{ll|c}
\toprule
 & \multirow{2}{*}{Model} & \multicolumn{1}{c}{Tasks} \\ 
\cmidrule(lr){3-3}
& & Response ($\uparrow$) (n=36) \\
\midrule
\multirow{8}{*}{\rotatebox{90}{\scriptsize{\textbf{Linear Probe}}}}
& \virchow\cite{vorontsov2024foundation} \mean & 0.612 ± 0.027 \\
& \gigapath\cite{xu2024whole} \mean & 0.585 ± 0.028 \\
& \chief\cite{wang2024chief} \mean & 0.643 ± 0.024 \\
& \conch\cite{} \mean & 0.620 ± 0.025 \\
& \prism\cite{shaikovski2024prism} & \underline{0.723 ± 0.022} \\
& \gigapath\cite{xu2024whole} & 0.552 ± 0.024 \\
& \chief\cite{wang2024chief} & 0.685 ± 0.033 \\
& $\ours$ & \textbf{0.730 ± 0.022} \\
\midrule
\multirow{5}{*}{\rotatebox{90}{\scriptsize{\textbf{Supervised}}}}
& ABMIL & 0.577 ± 0.029 \\
& \gigapath\cite{xu2024whole} & 0.450 ± 0.036 \\
& \chief\cite{wang2024chief} & 0.540 ± 0.031 \\
& $\ours$ Random Init & 0.702 ± 0.022 \\
& $\ours$ & 0.695 ± 0.023 \\
\midrule
\bottomrule
\end{tabular}
}\label{tab:allshot_nadt}
\end{table}

\begin{table}[h]
\centering
\caption{\textbf{Performance comparison between $\ours$ and baselines on GBM-Treatment tasks.} Best in \textbf{bold}, second best is \underline{underlined}. \textbf{Response}: This is a patient-level classification task evaluated using AUROC, mean and standard error reported over 50-fold Monte Carlo. }
\scalebox{0.675}{\begin{tabular}{ll|c}
\toprule
 & \multirow{2}{*}{Model} & \multicolumn{1}{c}{Tasks} \\ 
\cmidrule(lr){3-3}
& & Response ($\uparrow$) (n=93) \\
\midrule
\multirow{8}{*}{\rotatebox{90}{\scriptsize{\textbf{Linear Probe}}}}
& \virchow\cite{vorontsov2024foundation} \mean & 0.580 ± 0.018 \\
& \gigapath\cite{xu2024whole} \mean & 0.605 ± 0.017 \\
& \chief\cite{wang2024chief} \mean & 0.570 ± 0.017 \\
& \conch\cite{} \mean & 0.703 ± 0.018 \\
& \prism\cite{shaikovski2024prism} & 0.622 ± 0.022 \\
& \gigapath\cite{xu2024whole} & 0.642 ± 0.014 \\
& \chief\cite{wang2024chief} & 0.575 ± 0.019 \\
& $\ours$ & \textbf{0.741 ± 0.016} \\
\midrule
\multirow{5}{*}{\rotatebox{90}{\scriptsize{\textbf{Supervised}}}}
& ABMIL & 0.677 ± 0.016 \\
& \gigapath\cite{xu2024whole} & 0.680 ± 0.018 \\
& \chief\cite{wang2024chief} & 0.550 ± 0.021 \\
& $\ours$ Random Init & 0.655 ± 0.017 \\
& $\ours$ & \underline{0.705 ± 0.015} \\
\midrule
\bottomrule
\end{tabular}
}\label{tab:allshot_gbm_rt_tmz}
\end{table}

\begin{table}[h]
\centering
\caption{\textbf{Performance comparison between $\ours$ and baselines on Post-NAT-BRCA\cite{} tasks.} Best in \textbf{bold}, second best is \underline{underlined}. \textbf{Lymphovascular Invasion}: This is a slide-level classification task evaluated using AUROC, mean and standard error reported over 50-fold Monte Carlo. }
\scalebox{0.675}{\begin{tabular}{ll|c}
\toprule
 & \multirow{2}{*}{Model} & \multicolumn{1}{c}{Tasks} \\ 
\cmidrule(lr){3-3}
& & Lymphovascular Invasion ($\uparrow$) (n=53) \\
\midrule
\multirow{8}{*}{\rotatebox{90}{\scriptsize{\textbf{Linear Probe}}}}
& \virchow\cite{vorontsov2024foundation} \mean & 0.612 ± 0.028 \\
& \gigapath\cite{xu2024whole} \mean & 0.603 ± 0.025 \\
& \chief\cite{wang2024chief} \mean & 0.497 ± 0.023 \\
& \conch\cite{} \mean & 0.654 ± 0.022 \\
& \prism\cite{shaikovski2024prism} & 0.540 ± 0.029 \\
& \gigapath\cite{xu2024whole} & 0.582 ± 0.024 \\
& \chief\cite{wang2024chief} & 0.542 ± 0.030 \\
& $\ours$ & \textbf{0.701 ± 0.027} \\
\midrule
\multirow{5}{*}{\rotatebox{90}{\scriptsize{\textbf{Supervised}}}}
& ABMIL & 0.530 ± 0.029 \\
& \gigapath\cite{xu2024whole} & 0.607 ± 0.023 \\
& \chief\cite{wang2024chief} & 0.495 ± 0.028 \\
& $\ours$ Random Init & 0.600 ± 0.025 \\
& $\ours$ & \underline{0.662 ± 0.024} \\
\midrule
\bottomrule
\end{tabular}
}\label{tab:allshot_natbrca}
\end{table}

\begin{table}[h]
\centering
\caption{\textbf{Performance comparison between $\ours$ and baselines on SURGEN\cite{} tasks.} Best in \textbf{bold}, second best is \underline{underlined}. \textbf{BRAF Mutation}: This is a patient-level classification task evaluated using AUROC, mean and standard error reported over 5-fold cross-validation. \textbf{RAS Mutation}: This is a patient-level classification task evaluated using AUROC, mean and standard error reported over 5-fold cross-validation. \textbf{MMR Loss}: This is a patient-level classification task evaluated using AUROC, mean and standard error reported over 5-fold cross-validation. \textbf{Death in 5 Years}: This is a patient-level classification task evaluated using AUROC, mean and standard error reported over 5-fold cross-validation. }
\scalebox{0.675}{\begin{tabular}{ll|cccc}
\toprule
 & \multirow{2}{*}{Model} & \multicolumn{4}{c}{Tasks} \\ 
\cmidrule(lr){3-6}
& & BRAF Mutation ($\uparrow$) (n=388) & RAS Mutation ($\uparrow$) (n=389) & MMR Loss ($\uparrow$) (n=389) & Death in 5 Years ($\uparrow$) (n=387) \\
\midrule
\multirow{8}{*}{\rotatebox{90}{\scriptsize{\textbf{Linear Probe}}}}
& \virchow\cite{vorontsov2024foundation} \mean & 0.693 ± 0.037 & 0.586 ± 0.030 & 0.880 ± 0.017 & 0.642 ± 0.021 \\
& \gigapath\cite{xu2024whole} \mean & 0.694 ± 0.033 & 0.605 ± 0.015 & 0.877 ± 0.017 & 0.663 ± 0.016 \\
& \chief\cite{wang2024chief} \mean & 0.670 ± 0.033 & 0.627 ± 0.036 & 0.840 ± 0.020 & 0.671 ± 0.012 \\
& \conch\cite{} \mean & 0.650 ± 0.025 & 0.629 ± 0.028 & 0.796 ± 0.037 & 0.662 ± 0.023 \\
& \prism\cite{shaikovski2024prism} & \underline{0.740 ± 0.034} & 0.632 ± 0.014 & 0.883 ± 0.010 & 0.635 ± 0.023 \\
& \gigapath\cite{xu2024whole} & 0.680 ± 0.026 & 0.622 ± 0.022 & 0.860 ± 0.015 & 0.632 ± 0.013 \\
& \chief\cite{wang2024chief} & 0.736 ± 0.043 & \underline{0.652 ± 0.018} & 0.830 ± 0.031 & 0.683 ± 0.020 \\
& $\ours$ & 0.727 ± 0.042 & 0.633 ± 0.027 & \textbf{0.910 ± 0.023} & 0.685 ± 0.017 \\
\midrule
\multirow{5}{*}{\rotatebox{90}{\scriptsize{\textbf{Supervised}}}}
& ABMIL & 0.692 ± 0.039 & 0.629 ± 0.021 & 0.893 ± 0.028 & \underline{0.698 ± 0.024} \\
& \gigapath\cite{xu2024whole} & 0.665 ± 0.020 & 0.634 ± 0.015 & 0.779 ± 0.027 & 0.668 ± 0.022 \\
& \chief\cite{wang2024chief} & 0.725 ± 0.013 & 0.540 ± 0.024 & 0.749 ± 0.050 & 0.612 ± 0.013 \\
& $\ours$ Random Init & 0.697 ± 0.023 & 0.629 ± 0.022 & \underline{0.896 ± 0.016} & 0.687 ± 0.028 \\
& $\ours$ & \textbf{0.754 ± 0.037} & \textbf{0.676 ± 0.026} & 0.885 ± 0.032 & \textbf{0.715 ± 0.024} \\
\midrule
\bottomrule
\end{tabular}
}\label{tab:allshot_sr386_}
\end{table}

\begin{table}[h]
\centering
\caption{\textbf{Performance comparison between $\ours$ and baselines on MBC\cite{} tasks.} Best in \textbf{bold}, second best is \underline{underlined}. \textbf{Recist}: This is a patient-level classification task evaluated using quadratic weighted kappa, mean and standard error reported over 50-fold Monte Carlo. }
\scalebox{0.675}{\begin{tabular}{ll|c}
\toprule
 & \multirow{2}{*}{Model} & \multicolumn{1}{c}{Tasks} \\ 
\cmidrule(lr){3-3}
& & Recist ($\uparrow$) (n=76) \\
\midrule
\multirow{8}{*}{\rotatebox{90}{\scriptsize{\textbf{Linear Probe}}}}
& \virchow\cite{vorontsov2024foundation} \mean & 0.051 ± 0.028 \\
& \gigapath\cite{xu2024whole} \mean & 0.131 ± 0.034 \\
& \chief\cite{wang2024chief} \mean & 0.204 ± 0.038 \\
& \conch\cite{} \mean & 0.253 ± 0.038 \\
& \prism\cite{shaikovski2024prism} & 0.206 ± 0.034 \\
& \gigapath\cite{xu2024whole} & 0.016 ± 0.032 \\
& \chief\cite{wang2024chief} & 0.230 ± 0.033 \\
& $\ours$ & 0.258 ± 0.037 \\
\midrule
\multirow{5}{*}{\rotatebox{90}{\scriptsize{\textbf{Supervised}}}}
& ABMIL & \underline{0.268 ± 0.036} \\
& \gigapath\cite{xu2024whole} & 0.111 ± 0.032 \\
& \chief\cite{wang2024chief} & 0.015 ± 0.027 \\
& $\ours$ Random Init & 0.209 ± 0.027 \\
& $\ours$ & \textbf{0.326 ± 0.030} \\
\midrule
\bottomrule
\end{tabular}
}\label{tab:allshot_mbc_}
\end{table}

\begin{table}[h]
\centering
\caption{\textbf{Survival prediction on CPTAC-PDAC\cite{edwards2015cptac}.} Best in \textbf{bold}, second best is \underline{underlined}. $\alpha$ is the cost parameter for CoxNet, set empirically to allow for convergence. \textbf{Overall Survival}: This is a patient-level survival task evaluated using C-index, mean and standard error reported over 5-fold cross-validation. }
\scalebox{0.675}{\begin{tabular}{ll|c|c}
\toprule
 & \multirow{2}{*}{Model} & \multirow{2}{*}{$\alpha$} & \multicolumn{1}{c}{Tasks} \\ 
\cmidrule(lr){4-4}
& & & Overall Survival ($\uparrow$) (n=97) \\
\midrule
\multirow{8}{*}{\rotatebox{90}{\scriptsize{\textbf{CoxNet}}}}
& \virchow\cite{vorontsov2024foundation} \mean & 0.07 & 0.502 ± 0.051 \\
& \gigapath\cite{xu2024whole} \mean & 0.07 & 0.510 ± 0.043 \\
& \chief\cite{wang2024chief} \mean & 0.07 & 0.508 ± 0.042 \\
& \conch\cite{} \mean & 0.07 & 0.569 ± 0.018 \\
& \prism\cite{shaikovski2024prism} & 0.07 & 0.554 ± 0.027 \\
& \gigapath\cite{xu2024whole} & 0.07 & 0.426 ± 0.032 \\
& \chief\cite{wang2024chief} & 0.07 & 0.517 ± 0.032 \\
& $\ours$ & 0.07 & \textbf{0.616 ± 0.031} \\
\midrule
\multirow{5}{*}{\rotatebox{90}{\scriptsize{\textbf{Supervised}}}}
& ABMIL & -- & \underline{0.611 ± 0.042} \\
& \gigapath\cite{xu2024whole} & -- & 0.410 ± 0.013 \\
& \chief\cite{wang2024chief} & -- & 0.489 ± 0.046 \\
& $\ours$ Random Init & -- & 0.582 ± 0.037 \\
& $\ours$ & -- & 0.577 ± 0.043 \\
\midrule
\bottomrule
\end{tabular}
}\label{tab:survival_cptac_pda}
\end{table}

\begin{table}[h]
\centering
\caption{\textbf{Survival prediction on CPTAC-LUAD\cite{edwards2015cptac}.} Best in \textbf{bold}, second best is \underline{underlined}. $\alpha$ is the cost parameter for CoxNet, set empirically to allow for convergence. \textbf{Overall Survival}: This is a patient-level survival task evaluated using C-index, mean and standard error reported over 5-fold cross-validation. }
\scalebox{0.675}{\begin{tabular}{ll|c|c}
\toprule
 & \multirow{2}{*}{Model} & \multirow{2}{*}{$\alpha$} & \multicolumn{1}{c}{Tasks} \\ 
\cmidrule(lr){4-4}
& & & Overall Survival ($\uparrow$) (n=105) \\
\midrule
\multirow{8}{*}{\rotatebox{90}{\scriptsize{\textbf{CoxNet}}}}
& \virchow\cite{vorontsov2024foundation} \mean & 0.07 & 0.443 ± 0.072 \\
& \gigapath\cite{xu2024whole} \mean & 0.07 & 0.363 ± 0.041 \\
& \chief\cite{wang2024chief} \mean & 0.07 & 0.483 ± 0.009 \\
& \conch\cite{} \mean & 0.07 & 0.579 ± 0.060 \\
& \prism\cite{shaikovski2024prism} & 0.07 & \textbf{0.614 ± 0.032} \\
& \gigapath\cite{xu2024whole} & 0.07 & 0.469 ± 0.010 \\
& \chief\cite{wang2024chief} & 0.07 & 0.462 ± 0.030 \\
& $\ours$ & 0.07 & \underline{0.613 ± 0.054} \\
\midrule
\multirow{5}{*}{\rotatebox{90}{\scriptsize{\textbf{Supervised}}}}
& ABMIL & -- & 0.524 ± 0.069 \\
& \gigapath\cite{xu2024whole} & -- & 0.462 ± 0.062 \\
& \chief\cite{wang2024chief} & -- & 0.518 ± 0.050 \\
& $\ours$ Random Init & -- & 0.576 ± 0.046 \\
& $\ours$ & -- & 0.535 ± 0.074 \\
\midrule
\bottomrule
\end{tabular}
}\label{tab:survival_cptac_luad}
\end{table}

\begin{table}[h]
\centering
\caption{\textbf{Survival prediction on CPTAC-CCRCC\cite{edwards2015cptac}.} Best in \textbf{bold}, second best is \underline{underlined}. $\alpha$ is the cost parameter for CoxNet, set empirically to allow for convergence. \textbf{Overall Survival}: This is a patient-level survival task evaluated using C-index, mean and standard error reported over 5-fold cross-validation. }
\scalebox{0.675}{\begin{tabular}{ll|c|c}
\toprule
 & \multirow{2}{*}{Model} & \multirow{2}{*}{$\alpha$} & \multicolumn{1}{c}{Tasks} \\ 
\cmidrule(lr){4-4}
& & & Overall Survival ($\uparrow$) (n=94) \\
\midrule
\multirow{8}{*}{\rotatebox{90}{\scriptsize{\textbf{CoxNet}}}}
& \virchow\cite{vorontsov2024foundation} \mean & 0.07 & 0.554 ± 0.093 \\
& \gigapath\cite{xu2024whole} \mean & 0.07 & \underline{0.675 ± 0.063} \\
& \chief\cite{wang2024chief} \mean & 0.07 & 0.463 ± 0.034 \\
& \conch\cite{} \mean & 0.07 & 0.555 ± 0.091 \\
& \prism\cite{shaikovski2024prism} & 0.07 & 0.567 ± 0.048 \\
& \gigapath\cite{xu2024whole} & 0.07 & 0.550 ± 0.048 \\
& \chief\cite{wang2024chief} & 0.01 & 0.626 ± 0.076 \\
& $\ours$ & 0.07 & 0.673 ± 0.075 \\
\midrule
\multirow{5}{*}{\rotatebox{90}{\scriptsize{\textbf{Supervised}}}}
& ABMIL & -- & \textbf{0.693 ± 0.043} \\
& \gigapath\cite{xu2024whole} & -- & 0.527 ± 0.079 \\
& \chief\cite{wang2024chief} & -- & 0.369 ± 0.039 \\
& $\ours$ Random Init & -- & 0.529 ± 0.101 \\
& $\ours$ & -- & 0.495 ± 0.090 \\
\midrule
\bottomrule
\end{tabular}
}\label{tab:survival_cptac_ccrcc}
\end{table}

\begin{table}[h]
\centering
\caption{\textbf{Survival prediction on CPTAC-HNSC\cite{edwards2015cptac}.} Best in \textbf{bold}, second best is \underline{underlined}. $\alpha$ is the cost parameter for CoxNet, set empirically to allow for convergence. \textbf{Overall Survival}: This is a patient-level survival task evaluated using C-index, mean and standard error reported over 5-fold cross-validation. }
\scalebox{0.675}{\begin{tabular}{ll|c|c}
\toprule
 & \multirow{2}{*}{Model} & \multirow{2}{*}{$\alpha$} & \multicolumn{1}{c}{Tasks} \\ 
\cmidrule(lr){4-4}
& & & Overall Survival ($\uparrow$) (n=102) \\
\midrule
\multirow{8}{*}{\rotatebox{90}{\scriptsize{\textbf{CoxNet}}}}
& \virchow\cite{vorontsov2024foundation} \mean & 0.07 & 0.569 ± 0.049 \\
& \gigapath\cite{xu2024whole} \mean & 0.07 & \underline{0.590 ± 0.056} \\
& \chief\cite{wang2024chief} \mean & 0.07 & 0.480 ± 0.021 \\
& \conch\cite{} \mean & 0.07 & 0.568 ± 0.053 \\
& \prism\cite{shaikovski2024prism} & 0.07 & 0.587 ± 0.066 \\
& \gigapath\cite{xu2024whole} & 0.07 & 0.562 ± 0.063 \\
& \chief\cite{wang2024chief} & 0.07 & 0.474 ± 0.011 \\
& $\ours$ & 0.07 & \textbf{0.631 ± 0.076} \\
\midrule
\multirow{5}{*}{\rotatebox{90}{\scriptsize{\textbf{Supervised}}}}
& ABMIL & -- & 0.495 ± 0.024 \\
& \gigapath\cite{xu2024whole} & -- & 0.531 ± 0.064 \\
& \chief\cite{wang2024chief} & -- & 0.489 ± 0.047 \\
& $\ours$ Random Init & -- & 0.550 ± 0.014 \\
& $\ours$ & -- & 0.514 ± 0.011 \\
\midrule
\bottomrule
\end{tabular}
}\label{tab:survival_cptac_hnsc}
\end{table}

\begin{table}[h]
\centering
\caption{\textbf{Survival prediction on SURGEN\cite{}.} Best in \textbf{bold}, second best is \underline{underlined}. $\alpha$ is the cost parameter for CoxNet, set empirically to allow for convergence. \textbf{Overall Survival}: This is a patient-level survival task evaluated using C-index, mean and standard error reported over 5-fold cross-validation. }
\scalebox{0.675}{\begin{tabular}{ll|c|c}
\toprule
 & \multirow{2}{*}{Model} & \multirow{2}{*}{$\alpha$} & \multicolumn{1}{c}{Tasks} \\ 
\cmidrule(lr){4-4}
& & & Overall Survival ($\uparrow$) (n=144) \\
\midrule
\multirow{8}{*}{\rotatebox{90}{\scriptsize{\textbf{CoxNet}}}}
& \virchow\cite{vorontsov2024foundation} \mean & 0.07 & 0.593 ± 0.025 \\
& \gigapath\cite{xu2024whole} \mean & 0.07 & 0.606 ± 0.027 \\
& \chief\cite{wang2024chief} \mean & 0.07 & 0.606 ± 0.023 \\
& \conch\cite{} \mean & 0.07 & 0.625 ± 0.022 \\
& \prism\cite{shaikovski2024prism} & 0.07 & 0.578 ± 0.026 \\
& \gigapath\cite{xu2024whole} & 0.07 & 0.611 ± 0.025 \\
& \chief\cite{wang2024chief} & 0.07 & 0.600 ± 0.026 \\
& $\ours$ & 0.07 & \textbf{0.638 ± 0.014} \\
\midrule
\multirow{5}{*}{\rotatebox{90}{\scriptsize{\textbf{Supervised}}}}
& ABMIL & -- & 0.587 ± 0.026 \\
& \gigapath\cite{xu2024whole} & -- & 0.612 ± 0.022 \\
& \chief\cite{wang2024chief} & -- & 0.531 ± 0.037 \\
& $\ours$ Random Init & -- & \underline{0.632 ± 0.022} \\
& $\ours$ & -- & 0.613 ± 0.034 \\
\midrule
\bottomrule
\end{tabular}
}\label{tab:survival_sr386_}
\end{table}

\begin{table}[h]
\centering
\caption{\textbf{Survival prediction on MBC\cite{}.} Best in \textbf{bold}, second best is \underline{underlined}. $\alpha$ is the cost parameter for CoxNet, set empirically to allow for convergence. \textbf{Overall Survival}: This is a patient-level survival task evaluated using C-index, mean and standard error reported over 5-fold cross-validation. }
\scalebox{0.675}{\begin{tabular}{ll|c|c}
\toprule
 & \multirow{2}{*}{Model} & \multirow{2}{*}{$\alpha$} & \multicolumn{1}{c}{Tasks} \\ 
\cmidrule(lr){4-4}
& & & Overall Survival ($\uparrow$) (n=75) \\
\midrule
\multirow{8}{*}{\rotatebox{90}{\scriptsize{\textbf{CoxNet}}}}
& \virchow\cite{vorontsov2024foundation} \mean & 0.07 & 0.517 ± 0.031 \\
& \gigapath\cite{xu2024whole} \mean & 0.07 & 0.472 ± 0.024 \\
& \chief\cite{wang2024chief} \mean & 0.07 & 0.441 ± 0.040 \\
& \conch\cite{} \mean & 0.07 & 0.510 ± 0.052 \\
& \prism\cite{shaikovski2024prism} & 0.07 & 0.511 ± 0.038 \\
& \gigapath\cite{xu2024whole} & 0.07 & 0.440 ± 0.030 \\
& \chief\cite{wang2024chief} & 0.07 & 0.460 ± 0.046 \\
& $\ours$ & 0.07 & \underline{0.550 ± 0.027} \\
\midrule
\multirow{5}{*}{\rotatebox{90}{\scriptsize{\textbf{Supervised}}}}
& ABMIL & -- & 0.519 ± 0.043 \\
& \gigapath\cite{xu2024whole} & -- & 0.433 ± 0.014 \\
& \chief\cite{wang2024chief} & -- & 0.529 ± 0.064 \\
& $\ours$ Random Init & -- & 0.512 ± 0.034 \\
& $\ours$ & -- & \textbf{0.608 ± 0.030} \\
\midrule
\bottomrule
\end{tabular}
}\label{tab:survival_mbc_}
\end{table}

\begin{table}[h]
\centering
\caption{\textbf{Survival prediction on BOEHMK\cite{}.} Best in \textbf{bold}, second best is \underline{underlined}. $\alpha$ is the cost parameter for CoxNet, set empirically to allow for convergence. \gigapath-supervised could not be evaluated due to a bug in PyTorch autocast (https://github.com/pytorch/pytorch/issues/81876). \textbf{Progression Free Survival}: This is a patient-level survival task evaluated using C-index, mean and standard error reported over 5-fold cross-validation. }
\scalebox{0.675}{\begin{tabular}{ll|c|c}
\toprule
 & \multirow{2}{*}{Model} & \multirow{2}{*}{$\alpha$} & \multicolumn{1}{c}{Tasks} \\ 
\cmidrule(lr){4-4}
& & & Progression Free Survival ($\uparrow$) (n=183) \\
\midrule
\multirow{8}{*}{\rotatebox{90}{\scriptsize{\textbf{CoxNet}}}}
& \virchow\cite{vorontsov2024foundation} \mean & 0.01 & 0.513 ± 0.017 \\
& \gigapath\cite{xu2024whole} \mean & 0.01 & 0.487 ± 0.016 \\
& \chief\cite{wang2024chief} \mean & 0.01 & 0.480 ± 0.031 \\
& \conch\cite{} \mean & 0.01 & 0.536 ± 0.019 \\
& \prism\cite{shaikovski2024prism} & 0.02 & 0.500 ± 0.019 \\
& \gigapath\cite{xu2024whole} & 0.01 & 0.536 ± 0.043 \\
& \chief\cite{wang2024chief} & 0.01 & 0.520 ± 0.031 \\
& $\ours$ & 0.01 & 0.541 ± 0.013 \\
\midrule
\multirow{5}{*}{\rotatebox{90}{\scriptsize{\textbf{Supervised}}}}
& ABMIL & -- & 0.523 ± 0.036 \\
& \gigapath\cite{xu2024whole} & -- & --- \\
& \chief\cite{wang2024chief} & -- & 0.517 ± 0.033 \\
& $\ours$ Random Init & -- & \underline{0.553 ± 0.058} \\
& $\ours$ & -- & \textbf{0.575 ± 0.049} \\
\midrule
\bottomrule
\end{tabular}
}\label{tab:survival_boehmk_}
\end{table}

\begin{table}[h]
    \centering
    \caption{\textbf{Effect of increasing linear probe cost on our benchmark.} Lower cost corresponds to stronger regularization. Adaptive cost\cite{kolesnikov2019revisiting} is computed as $\frac{\texttt{embedding\_dim} \times \texttt{num\_classes}}{100}$.}
    \scalebox{0.75}{
        \begin{tabular}{l|c|c}
            \toprule
            Model & Cost & Benchmark (without survival) \\
            \midrule
            \midrule
            
            \virchow\cite{vorontsov2024foundation} \mean & \multirow{8}{*}{0.001} & 0.653 \\
            \gigapath\cite{xu2024whole} \mean &  & 0.663 \\
            \chief\cite{wang2024chief} \mean &  & 0.592 \\
            \conch \mean &  & 0.690 \\
            \prism\cite{shaikovski2024prism} &  & \underline{0.719} \\
            \gigapath\cite{xu2024whole} &  & 0.651 \\
            \chief\cite{wang2024chief} &  & 0.648 \\
            \ours\cite{} &  & \textbf{0.758} \\
            \midrule

            \virchow\cite{vorontsov2024foundation} \mean & \multirow{8}{*}{0.01} & 0.682 \\
            \gigapath\cite{xu2024whole} \mean &  & 0.691 \\
            \chief\cite{wang2024chief} \mean &  & 0.616 \\
            \conch \mean &  & 0.716 \\
            \prism\cite{shaikovski2024prism} &  & \underline{0.729} \\
            \gigapath\cite{xu2024whole} &  & 0.679 \\
            \chief\cite{wang2024chief} &  & 0.668 \\
            \ours\cite{} &  & \textbf{0.772} \\
            \midrule

            \virchow\cite{vorontsov2024foundation} \mean & \multirow{8}{*}{0.1} & 0.690 \\
            \gigapath\cite{xu2024whole} \mean &  & 0.691 \\
            \chief\cite{wang2024chief} \mean &  & 0.652 \\
            \conch \mean &  & 0.717 \\
            \prism\cite{shaikovski2024prism} &  & \underline{0.720} \\
            \gigapath\cite{xu2024whole} &  & 0.685 \\
            \chief\cite{wang2024chief} &  & 0.698 \\
            \ours\cite{} &  & \textbf{0.779} \\
            \midrule

            \virchow\cite{vorontsov2024foundation} \mean & \multirow{8}{*}{0.5} & 0.681 \\
            \gigapath\cite{xu2024whole} \mean &  & 0.683 \\
            \chief\cite{wang2024chief} \mean &  & 0.670 \\
            \conch \mean &  & 0.708 \\
            \prism\cite{shaikovski2024prism} &  & \underline{0.709} \\
            \gigapath\cite{xu2024whole} &  & 0.675 \\
            \chief\cite{wang2024chief} &  & 0.710 \\
            \ours\cite{} &  & \textbf{0.774} \\
            \midrule

            \virchow\cite{vorontsov2024foundation} \mean & \multirow{8}{*}{1.0} & 0.676 \\
            \gigapath\cite{xu2024whole} \mean &  & 0.679 \\
            \chief\cite{wang2024chief} \mean &  & 0.671 \\
            \conch \mean &  & 0.703 \\
            \prism\cite{shaikovski2024prism} &  & 0.704 \\
            \gigapath\cite{xu2024whole} &  & 0.669 \\
            \chief\cite{wang2024chief} &  & \underline{0.711} \\
            \ours\cite{} &  & \textbf{0.769} \\
            \midrule

            \virchow\cite{vorontsov2024foundation} \mean & \multirow{8}{*}{10.0} & 0.665 \\
            \gigapath\cite{xu2024whole} \mean &  & 0.670 \\
            \chief\cite{wang2024chief} \mean &  & 0.663 \\
            \conch \mean &  & 0.691 \\
            \prism\cite{shaikovski2024prism} &  & \underline{0.694} \\
            \gigapath\cite{xu2024whole} &  & 0.659 \\
            \chief\cite{wang2024chief} &  & \underline{0.694} \\
            \ours\cite{} &  & \textbf{0.751 }\\
            \midrule

            \virchow\cite{vorontsov2024foundation} \mean & \multirow{8}{*}{Adaptive} & 0.662 \\
            \gigapath\cite{xu2024whole} \mean &  & 0.668 \\
            \chief\cite{wang2024chief} \mean &  & 0.659 \\
            \conch \mean &  & 0.688 \\
            \prism\cite{shaikovski2024prism} &  & \underline{0.691} \\
            \gigapath\cite{xu2024whole} &  & 0.657 \\
            \chief\cite{wang2024chief} &  & 0.690 \\
            \ours\cite{} &  & \textbf{0.746} \\
            \midrule
            
            \bottomrule
            \end{tabular}
            }
\label{tab:cost-scale-linprobe}
\end{table}

\begin{table}[h]
    \centering
    \caption{\textbf{Generalizability experiments using linear probe (for classification) and CoxNet (for survival).} All samples from the train dataset are used for training, and all samples from the test dataset are used for testing. All tasks are binary classification. CI: 95\% confidence interval over 100 bootstraps.}
    \scalebox{0.7}{
    \begin{tabular}{ccccc}
        \toprule
        Train Dataset & Test Dataset & Task & Method & Mean AUC (CI) \\ 
        \midrule
        \midrule

		\multirow{8}{*}{CPTAC-CCRCC\cite{edwards2015cptac}} & \multirow{8}{*}{MUT-HET-RCC\cite{}} & \multirow{4}{*}{BAP1 Mutation} & $\prism$ & 0.776 (0.745-0.809) \\ 
		&  &  & $\gigapath$ & 0.793 (0.765-0.824) \\ 
		&  &  & $\chief$ & \underline{0.813 (0.788-0.840)} \\ 
		&  &  & $\ours$ & \textbf{0.840 (0.809-0.866)} \\ 
		\cline{3-5}
		&  & \multirow{4}{*}{PBRM1 Mutation} & $\prism$ & 0.665 (0.636-0.695) \\ 
		&  &  & $\gigapath$ & 0.577 (0.548-0.603) \\ 
		&  &  & $\chief$ & \underline{0.681 (0.654-0.710)} \\ 
		&  &  & $\ours$ & \textbf{0.701 (0.676-0.729)} \\ 
		\midrule
		\multirow{8}{*}{TCGA-BRCA\cite{}} & \multirow{8}{*}{BCNB\cite{xu2021predicting}} & \multirow{4}{*}{ER} & $\prism$ & 0.627 (0.590-0.663) \\ 
		&  &  & $\gigapath$ & 0.731 (0.698-0.770) \\ 
		&  &  & $\chief$ & \underline{0.835 (0.804-0.864)} \\ 
		&  &  & $\ours$ & \textbf{0.885 (0.859-0.906)} \\ 
		\cline{3-5}
		&  & \multirow{4}{*}{PR} & $\prism$ & 0.677 (0.638-0.708) \\ 
		&  &  & $\gigapath$ & 0.688 (0.654-0.716) \\ 
		&  &  & $\chief$ & \underline{0.787 (0.760-0.813)} \\ 
		&  &  & $\ours$ & \textbf{0.794 (0.767-0.822)} \\ 
		\midrule
  
		\multirow{4}{*}{TCGA-GBMLGG\cite{}} & \multirow{4}{*}{EBRAINS\cite{roetzer2022ebrains}} & \multirow{4}{*}{IDH Status} & $\prism$ & 0.931 (0.909-0.947) \\ 
		&  &  & $\gigapath$ & 0.904 (0.882-0.926) \\ 
		&  &  & $\chief$ & \underline{0.935 (0.913-0.954)} \\ 
		&  &  & $\ours$ & \textbf{0.961 (0.947-0.975)} \\ 
		\midrule
		
		\multirow{4}{*}{TCGA-NSCLC\cite{}} & \multirow{4}{*}{MGB Lung} & \multirow{4}{*}{Subtype} & $\prism$ & \underline{0.969 (0.957-0.981)} \\ 
		&  &  & $\gigapath$ & 0.890 (0.871-0.905) \\ 
		&  &  & $\chief$ & 0.951 (0.937-0.962) \\ 
		&  &  & $\ours$ & \textbf{0.984 (0.978-0.989)} \\ 
		\midrule

            \multirow{4}{*}{TCGA-BRCA\cite{}} & \multirow{4}{*}{MGB Breast} & \multirow{4}{*}{Subtype} & $\prism$ & 0.963 (0.950-0.973) \\ 
		&  &  & $\gigapath$ & 0.919 (0.905-0.934) \\ 
		&  &  & $\chief$ & \textbf{0.970 (0.962-0.978)} \\ 
		&  &  & $\ours$ & \underline{0.965 (0.956-0.973)} \\ 
		\midrule
  
            \multirow{4}{*}{TCGA-LUAD\cite{}} & \multirow{4}{*}{CPTAC-LUAD\cite{edwards2015cptac}} & \multirow{4}{*}{Overall Survival} & $\prism$ & 0.545 (0.407-0.681) \\ 
		&  &  & $\gigapath$ & \underline{0.561 (0.413-0.685)} \\ 
		&  &  & $\chief$ & 0.495 (0.369-0.607) \\ 
		&  &  & $\ours$ & \textbf{0.654 (0.511-0.748)} \\ 
		\midrule
  
		\multirow{4}{*}{TCGA-PDAC\cite{}} & \multirow{4}{*}{CPTAC-PDAC\cite{edwards2015cptac}} & \multirow{4}{*}{Overall Survival} & $\prism$ & 0.505 (0.432-0.587) \\ 
		&  &  & $\gigapath$ & \underline{0.571 (0.493-0.642)} \\ 
		&  &  & $\chief$ & 0.505 (0.419-0.592) \\ 
		&  &  & $\ours$ & \textbf{0.613 (0.546-0.696)} \\ 

        \bottomrule
    \end{tabular}
    }
    \label{tab:generalizability}
\end{table}

\begin{table}[h]
\centering
\caption{\textbf{Performance comparison between $\ours$ and baselines on BCNB\cite{xu2021predicting} tasks in a few-shot setting.} Best in \textbf{bold}, second best is \underline{underlined}. \textbf{ER}: This is a patient-level classification task evaluated using AUROC, mean and standard error reported over 5-fold cross-validation. \textbf{PR}: This is a patient-level classification task evaluated using AUROC, mean and standard error reported over 5-fold cross-validation. \textbf{HER2}: This is a patient-level classification task evaluated using AUROC, mean and standard error reported over 5-fold cross-validation. }
\scalebox{0.675}{\begin{tabular}{ll|c|ccc}
\toprule
 & \multirow{2}{*}{Model} & \multirow{2}{*}{\# Shots} & \multicolumn{3}{c}{Tasks} \\ 
\cmidrule(lr){4-6}
& & & ER  (n=1058) & PR  (n=1058) & HER2  (n=1058) \\
\midrule
& \virchow\cite{vorontsov2024foundation} \mean & 1 & 0.520 ± 0.025 & 0.555 ± 0.012 & 0.540 ± 0.011 \\
& \gigapath\cite{xu2024whole} \mean & 1 & 0.505 ± 0.014 & 0.576 ± 0.018 & 0.542 ± 0.012 \\
& \chief\cite{wang2024chief} \mean & 1 & 0.485 ± 0.016 & 0.569 ± 0.010 & 0.493 ± 0.017 \\
& \conch\cite{} \mean & 1 & 0.520 ± 0.082 & \underline{0.627 ± 0.035} & \underline{0.568 ± 0.031} \\
& \prism\cite{shaikovski2024prism} & 1 & \underline{0.574 ± 0.055} & 0.560 ± 0.051 & \textbf{0.624 ± 0.023} \\
& \gigapath\cite{xu2024whole} & 1 & 0.514 ± 0.024 & 0.574 ± 0.021 & 0.523 ± 0.010 \\
& \chief\cite{wang2024chief} & 1 & 0.566 ± 0.031 & 0.603 ± 0.018 & 0.488 ± 0.020 \\
& $\ours$ & 1 & \textbf{0.671 ± 0.054} & \textbf{0.678 ± 0.091} & 0.543 ± 0.062 \\
\midrule
& \virchow\cite{vorontsov2024foundation} \mean & 2 & 0.512 ± 0.035 & 0.546 ± 0.023 & 0.541 ± 0.011 \\
& \gigapath\cite{xu2024whole} \mean & 2 & 0.583 ± 0.046 & 0.594 ± 0.031 & 0.531 ± 0.019 \\
& \chief\cite{wang2024chief} \mean & 2 & 0.560 ± 0.038 & 0.557 ± 0.026 & 0.487 ± 0.046 \\
& \conch\cite{} \mean & 2 & 0.602 ± 0.073 & \underline{0.675 ± 0.020} & \underline{0.574 ± 0.020} \\
& \prism\cite{shaikovski2024prism} & 2 & \underline{0.687 ± 0.043} & 0.627 ± 0.050 & \textbf{0.580 ± 0.039} \\
& \gigapath\cite{xu2024whole} & 2 & 0.573 ± 0.042 & 0.572 ± 0.032 & 0.525 ± 0.027 \\
& \chief\cite{wang2024chief} & 2 & 0.657 ± 0.032 & 0.614 ± 0.016 & 0.497 ± 0.030 \\
& $\ours$ & 2 & \textbf{0.783 ± 0.055} & \textbf{0.770 ± 0.030} & \textbf{0.580 ± 0.037} \\
\midrule
& \virchow\cite{vorontsov2024foundation} \mean & 4 & 0.568 ± 0.044 & 0.587 ± 0.035 & 0.523 ± 0.012 \\
& \gigapath\cite{xu2024whole} \mean & 4 & 0.668 ± 0.051 & 0.629 ± 0.039 & 0.545 ± 0.031 \\
& \chief\cite{wang2024chief} \mean & 4 & 0.585 ± 0.036 & 0.599 ± 0.039 & 0.524 ± 0.023 \\
& \conch\cite{} \mean & 4 & 0.696 ± 0.068 & \underline{0.713 ± 0.040} & 0.565 ± 0.034 \\
& \prism\cite{shaikovski2024prism} & 4 & \underline{0.734 ± 0.039} & 0.657 ± 0.040 & \underline{0.643 ± 0.019} \\
& \gigapath\cite{xu2024whole} & 4 & 0.638 ± 0.050 & 0.611 ± 0.042 & 0.526 ± 0.019 \\
& \chief\cite{wang2024chief} & 4 & 0.724 ± 0.036 & 0.666 ± 0.028 & 0.578 ± 0.017 \\
& $\ours$ & 4 & \textbf{0.838 ± 0.021} & \textbf{0.725 ± 0.061} & \textbf{0.651 ± 0.024} \\
\midrule
& \virchow\cite{vorontsov2024foundation} \mean & 8 & 0.601 ± 0.036 & 0.592 ± 0.024 & 0.555 ± 0.022 \\
& \gigapath\cite{xu2024whole} \mean & 8 & 0.686 ± 0.046 & 0.651 ± 0.023 & 0.594 ± 0.032 \\
& \chief\cite{wang2024chief} \mean & 8 & 0.623 ± 0.034 & 0.618 ± 0.023 & 0.540 ± 0.023 \\
& \conch\cite{} \mean & 8 & 0.716 ± 0.060 & \underline{0.721 ± 0.037} & 0.608 ± 0.033 \\
& \prism\cite{shaikovski2024prism} & 8 & \underline{0.764 ± 0.027} & 0.695 ± 0.027 & \underline{0.655 ± 0.017} \\
& \gigapath\cite{xu2024whole} & 8 & 0.663 ± 0.042 & 0.619 ± 0.028 & 0.567 ± 0.027 \\
& \chief\cite{wang2024chief} & 8 & 0.756 ± 0.027 & 0.687 ± 0.028 & 0.606 ± 0.026 \\
& $\ours$ & 8 & \textbf{0.836 ± 0.028} & \textbf{0.756 ± 0.034} & \textbf{0.666 ± 0.012} \\
\midrule
& \virchow\cite{vorontsov2024foundation} \mean & 16 & 0.683 ± 0.030 & 0.641 ± 0.019 & 0.591 ± 0.024 \\
& \gigapath\cite{xu2024whole} \mean & 16 & 0.762 ± 0.021 & 0.698 ± 0.015 & 0.605 ± 0.029 \\
& \chief\cite{wang2024chief} \mean & 16 & 0.678 ± 0.021 & 0.635 ± 0.019 & 0.561 ± 0.026 \\
& \conch\cite{} \mean & 16 & 0.786 ± 0.032 & 0.723 ± 0.025 & 0.626 ± 0.030 \\
& \prism\cite{shaikovski2024prism} & 16 & \underline{0.828 ± 0.017} & \underline{0.744 ± 0.019} & \underline{0.656 ± 0.030} \\
& \gigapath\cite{xu2024whole} & 16 & 0.731 ± 0.034 & 0.643 ± 0.014 & 0.599 ± 0.026 \\
& \chief\cite{wang2024chief} & 16 & 0.801 ± 0.015 & 0.713 ± 0.020 & 0.645 ± 0.026 \\
& $\ours$ & 16 & \textbf{0.874 ± 0.018} & \textbf{0.786 ± 0.019} & \textbf{0.684 ± 0.023} \\
\midrule
& \virchow\cite{vorontsov2024foundation} \mean & 32 & 0.757 ± 0.031 & 0.688 ± 0.023 & 0.614 ± 0.020 \\
& \gigapath\cite{xu2024whole} \mean & 32 & 0.787 ± 0.021 & 0.728 ± 0.025 & 0.612 ± 0.024 \\
& \chief\cite{wang2024chief} \mean & 32 & 0.720 ± 0.023 & 0.667 ± 0.024 & 0.598 ± 0.029 \\
& \conch\cite{} \mean & 32 & 0.810 ± 0.024 & 0.744 ± 0.021 & 0.632 ± 0.019 \\
& \prism\cite{shaikovski2024prism} & 32 & \underline{0.822 ± 0.023} & \underline{0.752 ± 0.025} & \underline{0.654 ± 0.017} \\
& \gigapath\cite{xu2024whole} & 32 & 0.769 ± 0.022 & 0.692 ± 0.020 & 0.619 ± 0.023 \\
& \chief\cite{wang2024chief} & 32 & 0.809 ± 0.016 & 0.734 ± 0.022 & 0.646 ± 0.026 \\
& $\ours$ & 32 & \textbf{0.882 ± 0.014} & \textbf{0.798 ± 0.026} & \textbf{0.694 ± 0.020} \\
\midrule
\midrule
\bottomrule
\end{tabular}
}\label{tab:fewshot_bcnb}
\end{table}

\begin{table}[h]
\centering
\caption{\textbf{Performance comparison between $\ours$ and baselines on BRACS\cite{brancati2021bracs} tasks in a few-shot setting.} Best in \textbf{bold}, second best is \underline{underlined}. \textbf{Fine Subtype}: This is a slide-level classification task evaluated using balanced accuracy, mean and standard error reported over 5-fold cross-validation. \textbf{Coarse Subtype}: This is a slide-level classification task evaluated using balanced accuracy, mean and standard error reported over 5-fold cross-validation. }
\scalebox{0.675}{\begin{tabular}{ll|c|cc}
\toprule
 & \multirow{2}{*}{Model} & \multirow{2}{*}{\# Shots} & \multicolumn{2}{c}{Tasks} \\ 
\cmidrule(lr){4-5}
& & & Fine Subtype  (n=547) & Coarse Subtype  (n=547) \\
\midrule
& \virchow\cite{vorontsov2024foundation} \mean & 1 & 0.229 ± 0.024 & 0.335 ± 0.020 \\
& \gigapath\cite{xu2024whole} \mean & 1 & 0.201 ± 0.014 & 0.328 ± 0.021 \\
& \chief\cite{wang2024chief} \mean & 1 & 0.216 ± 0.021 & 0.333 ± 0.024 \\
& \conch\cite{} \mean & 1 & 0.241 ± 0.024 & 0.308 ± 0.024 \\
& \prism\cite{shaikovski2024prism} & 1 & \textbf{0.334 ± 0.057} & \textbf{0.462 ± 0.055} \\
& \gigapath\cite{xu2024whole} & 1 & 0.198 ± 0.021 & 0.309 ± 0.017 \\
& \chief\cite{wang2024chief} & 1 & \underline{0.291 ± 0.040} & 0.373 ± 0.019 \\
& $\ours$ & 1 & \underline{0.291 ± 0.032} & \underline{0.402 ± 0.050} \\
\midrule
& \virchow\cite{vorontsov2024foundation} \mean & 2 & 0.204 ± 0.023 & 0.369 ± 0.017 \\
& \gigapath\cite{xu2024whole} \mean & 2 & 0.227 ± 0.022 & 0.369 ± 0.015 \\
& \chief\cite{wang2024chief} \mean & 2 & 0.254 ± 0.023 & 0.350 ± 0.017 \\
& \conch\cite{} \mean & 2 & 0.231 ± 0.015 & 0.398 ± 0.022 \\
& \prism\cite{shaikovski2024prism} & 2 & \textbf{0.361 ± 0.041} & \textbf{0.549 ± 0.031} \\
& \gigapath\cite{xu2024whole} & 2 & 0.196 ± 0.016 & 0.369 ± 0.014 \\
& \chief\cite{wang2024chief} & 2 & 0.319 ± 0.029 & 0.482 ± 0.045 \\
& $\ours$ & 2 & \underline{0.346 ± 0.033} & \underline{0.521 ± 0.031} \\
\midrule
& \virchow\cite{vorontsov2024foundation} \mean & 4 & 0.258 ± 0.023 & 0.430 ± 0.021 \\
& \gigapath\cite{xu2024whole} \mean & 4 & 0.234 ± 0.018 & 0.413 ± 0.023 \\
& \chief\cite{wang2024chief} \mean & 4 & 0.255 ± 0.027 & 0.410 ± 0.022 \\
& \conch\cite{} \mean & 4 & 0.259 ± 0.026 & 0.492 ± 0.018 \\
& \prism\cite{shaikovski2024prism} & 4 & \underline{0.363 ± 0.019} & \underline{0.551 ± 0.027} \\
& \gigapath\cite{xu2024whole} & 4 & 0.237 ± 0.015 & 0.402 ± 0.028 \\
& \chief\cite{wang2024chief} & 4 & 0.351 ± 0.041 & 0.547 ± 0.041 \\
& $\ours$ & 4 & \textbf{0.387 ± 0.022} & \textbf{0.588 ± 0.025} \\
\midrule
& \virchow\cite{vorontsov2024foundation} \mean & 8 & 0.315 ± 0.033 & 0.459 ± 0.015 \\
& \gigapath\cite{xu2024whole} \mean & 8 & 0.285 ± 0.020 & 0.473 ± 0.024 \\
& \chief\cite{wang2024chief} \mean & 8 & 0.276 ± 0.021 & 0.450 ± 0.022 \\
& \conch\cite{} \mean & 8 & 0.290 ± 0.022 & 0.523 ± 0.051 \\
& \prism\cite{shaikovski2024prism} & 8 & \textbf{0.416 ± 0.026} & 0.551 ± 0.033 \\
& \gigapath\cite{xu2024whole} & 8 & 0.281 ± 0.015 & 0.461 ± 0.028 \\
& \chief\cite{wang2024chief} & 8 & \underline{0.411 ± 0.027} & \underline{0.567 ± 0.031} \\
& $\ours$ & 8 & 0.394 ± 0.014 & \textbf{0.602 ± 0.038} \\
\midrule
& \virchow\cite{vorontsov2024foundation} \mean & 16 & 0.368 ± 0.036 & 0.480 ± 0.017 \\
& \gigapath\cite{xu2024whole} \mean & 16 & 0.314 ± 0.033 & 0.494 ± 0.038 \\
& \chief\cite{wang2024chief} \mean & 16 & 0.343 ± 0.033 & 0.526 ± 0.011 \\
& \conch\cite{} \mean & 16 & 0.318 ± 0.041 & 0.539 ± 0.038 \\
& \prism\cite{shaikovski2024prism} & 16 & 0.418 ± 0.014 & \underline{0.635 ± 0.015} \\
& \gigapath\cite{xu2024whole} & 16 & 0.314 ± 0.033 & 0.512 ± 0.036 \\
& \chief\cite{wang2024chief} & 16 & \textbf{0.434 ± 0.024} & 0.618 ± 0.017 \\
& $\ours$ & 16 & \underline{0.425 ± 0.015} & \textbf{0.701 ± 0.027} \\
\midrule
& \virchow\cite{vorontsov2024foundation} \mean & 32 & 0.349 ± 0.020 & 0.527 ± 0.009 \\
& \gigapath\cite{xu2024whole} \mean & 32 & 0.353 ± 0.025 & 0.505 ± 0.032 \\
& \chief\cite{wang2024chief} \mean & 32 & 0.335 ± 0.041 & 0.544 ± 0.019 \\
& \conch\cite{} \mean & 32 & 0.361 ± 0.042 & 0.607 ± 0.045 \\
& \prism\cite{shaikovski2024prism} & 32 & \underline{0.435 ± 0.030} & 0.632 ± 0.023 \\
& \gigapath\cite{xu2024whole} & 32 & 0.362 ± 0.031 & 0.535 ± 0.037 \\
& \chief\cite{wang2024chief} & 32 & 0.417 ± 0.021 & \underline{0.649 ± 0.023} \\
& $\ours$ & 32 & \textbf{0.502 ± 0.028} & \textbf{0.726 ± 0.021} \\
\midrule
\midrule
\bottomrule
\end{tabular}
}\label{tab:fewshot_bracs}
\end{table}

\begin{table}[h]
\centering
\caption{\textbf{Performance comparison between $\ours$ and baselines on EBRAINS\cite{roetzer2022ebrains} tasks in a few-shot setting.} Best in \textbf{bold}, second best is \underline{underlined}. \textbf{Fine Subtype}: This is a slide-level classification task evaluated using balanced accuracy, mean and 95\% CI reported over 5 bootstraps of the official train split. \textbf{Coarse Subtype}: This is a slide-level classification task evaluated using balanced accuracy, mean and 95\% CI reported over 5 bootstraps of the official train split. }
\scalebox{0.675}{\begin{tabular}{ll|c|cc}
\toprule
 & \multirow{2}{*}{Model} & \multirow{2}{*}{\# Shots} & \multicolumn{2}{c}{Tasks} \\ 
\cmidrule(lr){4-5}
& & & Fine Subtype  (n=2319) & Coarse Subtype  (n=2319) \\
\midrule
& \virchow\cite{vorontsov2024foundation} \mean & 1 & 0.303 ± 0.017 & 0.368 ± 0.013 \\
& \gigapath\cite{xu2024whole} \mean & 1 & 0.304 ± 0.016 & 0.347 ± 0.015 \\
& \chief\cite{wang2024chief} \mean & 1 & 0.215 ± 0.006 & 0.273 ± 0.008 \\
& \conch\cite{} \mean & 1 & 0.361 ± 0.012 & 0.414 ± 0.021 \\
& \prism\cite{shaikovski2024prism} & 1 & \underline{0.406 ± 0.017} & \underline{0.463 ± 0.018} \\
& \gigapath\cite{xu2024whole} & 1 & 0.293 ± 0.012 & 0.331 ± 0.015 \\
& \chief\cite{wang2024chief} & 1 & 0.234 ± 0.016 & 0.262 ± 0.020 \\
& $\ours$ & 1 & \textbf{0.503 ± 0.008} & \textbf{0.595 ± 0.028} \\
\midrule
& \virchow\cite{vorontsov2024foundation} \mean & 2 & 0.380 ± 0.010 & 0.472 ± 0.016 \\
& \gigapath\cite{xu2024whole} \mean & 2 & 0.413 ± 0.005 & 0.490 ± 0.020 \\
& \chief\cite{wang2024chief} \mean & 2 & 0.282 ± 0.007 & 0.373 ± 0.015 \\
& \conch\cite{} \mean & 2 & 0.449 ± 0.015 & 0.537 ± 0.008 \\
& \prism\cite{shaikovski2024prism} & 2 & \underline{0.505 ± 0.008} & \underline{0.614 ± 0.018} \\
& \gigapath\cite{xu2024whole} & 2 & 0.395 ± 0.006 & 0.460 ± 0.013 \\
& \chief\cite{wang2024chief} & 2 & 0.311 ± 0.020 & 0.369 ± 0.030 \\
& $\ours$ & 2 & \textbf{0.570 ± 0.004} & \textbf{0.736 ± 0.022} \\
\midrule
& \virchow\cite{vorontsov2024foundation} \mean & 4 & 0.489 ± 0.006 & 0.612 ± 0.009 \\
& \gigapath\cite{xu2024whole} \mean & 4 & 0.522 ± 0.009 & 0.625 ± 0.009 \\
& \chief\cite{wang2024chief} \mean & 4 & 0.371 ± 0.009 & 0.450 ± 0.014 \\
& \conch\cite{} \mean & 4 & 0.525 ± 0.009 & 0.631 ± 0.020 \\
& \prism\cite{shaikovski2024prism} & 4 & \underline{0.565 ± 0.009} & \underline{0.693 ± 0.010} \\
& \gigapath\cite{xu2024whole} & 4 & 0.490 ± 0.008 & 0.590 ± 0.010 \\
& \chief\cite{wang2024chief} & 4 & 0.402 ± 0.012 & 0.456 ± 0.012 \\
& $\ours$ & 4 & \textbf{0.635 ± 0.007} & \textbf{0.802 ± 0.012} \\
\midrule
& \virchow\cite{vorontsov2024foundation} \mean & 8 & 0.593 ± 0.013 & 0.703 ± 0.010 \\
& \gigapath\cite{xu2024whole} \mean & 8 & 0.598 ± 0.008 & 0.719 ± 0.010 \\
& \chief\cite{wang2024chief} \mean & 8 & 0.457 ± 0.006 & 0.544 ± 0.014 \\
& \conch\cite{} \mean & 8 & 0.592 ± 0.004 & 0.748 ± 0.010 \\
& \prism\cite{shaikovski2024prism} & 8 & \underline{0.603 ± 0.006} & \underline{0.782 ± 0.004} \\
& \gigapath\cite{xu2024whole} & 8 & 0.592 ± 0.007 & 0.689 ± 0.012 \\
& \chief\cite{wang2024chief} & 8 & 0.494 ± 0.007 & 0.590 ± 0.012 \\
& $\ours$ & 8 & \textbf{0.683 ± 0.008} & \textbf{0.859 ± 0.004} \\
\midrule
& \virchow\cite{vorontsov2024foundation} \mean & 16 & 0.644 ± 0.005 & 0.793 ± 0.005 \\
& \gigapath\cite{xu2024whole} \mean & 16 & \underline{0.674 ± 0.007} & 0.800 ± 0.006 \\
& \chief\cite{wang2024chief} \mean & 16 & 0.541 ± 0.007 & 0.618 ± 0.007 \\
& \conch\cite{} \mean & 16 & 0.641 ± 0.006 & 0.801 ± 0.004 \\
& \prism\cite{shaikovski2024prism} & 16 & 0.643 ± 0.008 & \underline{0.809 ± 0.005} \\
& \gigapath\cite{xu2024whole} & 16 & 0.661 ± 0.007 & 0.777 ± 0.007 \\
& \chief\cite{wang2024chief} & 16 & 0.566 ± 0.006 & 0.692 ± 0.010 \\
& $\ours$ & 16 & \textbf{0.712 ± 0.004} & \textbf{0.882 ± 0.002} \\
\midrule
& \virchow\cite{vorontsov2024foundation} \mean & 32 & --- & 0.841 ± 0.005 \\
& \gigapath\cite{xu2024whole} \mean & 32 & --- & \underline{0.866 ± 0.005} \\
& \chief\cite{wang2024chief} \mean & 32 & --- & 0.700 ± 0.006 \\
& \conch\cite{} \mean & 32 & --- & 0.832 ± 0.004 \\
& \prism\cite{shaikovski2024prism} & 32 & --- & 0.826 ± 0.005 \\
& \gigapath\cite{xu2024whole} & 32 & --- & 0.847 ± 0.007 \\
& \chief\cite{wang2024chief} & 32 & --- & 0.751 ± 0.003 \\
& $\ours$ & 32 & --- & \textbf{0.905 ± 0.002} \\
\midrule
\midrule
\bottomrule
\end{tabular}
}\label{tab:fewshot_ebrains}
\end{table}

\begin{table}[h]
\centering
\caption{\textbf{Performance comparison between $\ours$ and baselines on GBM-Treatment tasks in a few-shot setting.} Best in \textbf{bold}, second best is \underline{underlined}. \textbf{Response}: This is a patient-level classification task evaluated using AUROC, mean and standard error reported over 50-fold Monte Carlo. }
\scalebox{0.675}{\begin{tabular}{ll|c|c}
\toprule
 & \multirow{2}{*}{Model} & \multirow{2}{*}{\# Shots} & \multicolumn{1}{c}{Tasks} \\ 
\cmidrule(lr){4-4}
& & & Response  (n=93) \\
\midrule
& \virchow\cite{vorontsov2024foundation} \mean & 1 & \textbf{0.583 ± 0.021} \\
& \gigapath\cite{xu2024whole} \mean & 1 & 0.519 ± 0.020 \\
& \chief\cite{wang2024chief} \mean & 1 & 0.526 ± 0.022 \\
& \conch\cite{} \mean & 1 & \underline{0.578 ± 0.021} \\
& \prism\cite{shaikovski2024prism} & 1 & 0.573 ± 0.023 \\
& \gigapath\cite{xu2024whole} & 1 & 0.550 ± 0.022 \\
& \chief\cite{wang2024chief} & 1 & 0.525 ± 0.019 \\
& $\ours$ & 1 & 0.551 ± 0.021 \\
\midrule
& \virchow\cite{vorontsov2024foundation} \mean & 2 & \underline{0.555 ± 0.026} \\
& \gigapath\cite{xu2024whole} \mean & 2 & 0.530 ± 0.023 \\
& \chief\cite{wang2024chief} \mean & 2 & 0.542 ± 0.022 \\
& \conch\cite{} \mean & 2 & 0.554 ± 0.021 \\
& \prism\cite{shaikovski2024prism} & 2 & \textbf{0.558 ± 0.022} \\
& \gigapath\cite{xu2024whole} & 2 & 0.551 ± 0.025 \\
& \chief\cite{wang2024chief} & 2 & 0.509 ± 0.019 \\
& $\ours$ & 2 & 0.548 ± 0.020 \\
\midrule
& \virchow\cite{vorontsov2024foundation} \mean & 4 & 0.556 ± 0.020 \\
& \gigapath\cite{xu2024whole} \mean & 4 & 0.539 ± 0.022 \\
& \chief\cite{wang2024chief} \mean & 4 & 0.528 ± 0.019 \\
& \conch\cite{} \mean & 4 & \textbf{0.606 ± 0.019} \\
& \prism\cite{shaikovski2024prism} & 4 & 0.554 ± 0.023 \\
& \gigapath\cite{xu2024whole} & 4 & 0.553 ± 0.025 \\
& \chief\cite{wang2024chief} & 4 & 0.489 ± 0.023 \\
& $\ours$ & 4 & \underline{0.566 ± 0.020} \\
\midrule
& \virchow\cite{vorontsov2024foundation} \mean & 8 & 0.591 ± 0.020 \\
& \gigapath\cite{xu2024whole} \mean & 8 & 0.581 ± 0.023 \\
& \chief\cite{wang2024chief} \mean & 8 & 0.534 ± 0.022 \\
& \conch\cite{} \mean & 8 & \textbf{0.634 ± 0.019} \\
& \prism\cite{shaikovski2024prism} & 8 & 0.574 ± 0.026 \\
& \gigapath\cite{xu2024whole} & 8 & 0.613 ± 0.024 \\
& \chief\cite{wang2024chief} & 8 & 0.511 ± 0.023 \\
& $\ours$ & 8 & \underline{0.618 ± 0.022} \\
\midrule
& \virchow\cite{vorontsov2024foundation} \mean & 16 & 0.572 ± 0.017 \\
& \gigapath\cite{xu2024whole} \mean & 16 & 0.572 ± 0.019 \\
& \chief\cite{wang2024chief} \mean & 16 & 0.527 ± 0.019 \\
& \conch\cite{} \mean & 16 & \underline{0.638 ± 0.020} \\
& \prism\cite{shaikovski2024prism} & 16 & 0.630 ± 0.021 \\
& \gigapath\cite{xu2024whole} & 16 & 0.608 ± 0.017 \\
& \chief\cite{wang2024chief} & 16 & 0.503 ± 0.021 \\
& $\ours$ & 16 & \textbf{0.676 ± 0.017} \\
\midrule
\bottomrule
\end{tabular}
}\label{tab:fewshot_gbm_rt_tmz}
\end{table}

\begin{table}[h]
\centering
\caption{\textbf{Retrieval performance on EBRAINS fine (30 classes) and coarse (12 classes) subtyping.} Best in \textbf{bold}, second best is \underline{underlined}. \textbf{mAP@K}: mean average precision using top-k retrieved examples. CI: 95\% confidence interval.}
\scalebox{0.9}{
\begin{tabular}{lllll}
\toprule
Model & Subtyping Task & mAP@1 (CI) & mAP@5 (CI) & mAP@10 (CI) \\
\midrule
\virchow\cite{vorontsov2024foundation} \mean & Fine & 0.593 (0.555-0.627) & 0.453 (0.425-0.480) & 0.393 (0.366-0.417) \\
\gigapath\cite{xu2024whole} \mean & Fine & 0.581 (0.536-0.626) & 0.456 (0.427-0.490) & 0.395 (0.367-0.425) \\
\chief\cite{wang2024chief} \mean & Fine & 0.463 (0.425-0.504) & 0.317 (0.288-0.340) & 0.260 (0.235-0.279) \\
\conch\cite{} \mean & Fine & 0.645 (0.610-0.675) & 0.514 (0.488-0.540) & 0.463 (0.438-0.487) \\
\midrule
\prism\cite{shaikovski2024prism} & Fine & \underline{0.648 (0.615-0.686)} & \underline{0.534 (0.506-0.560)} & \underline{0.488 (0.463-0.514)} \\
\gigapath\cite{xu2024whole} & Fine & 0.607 (0.572-0.640) & 0.454 (0.426-0.482) & 0.383 (0.355-0.411) \\
\chief\cite{wang2024chief} & Fine & 0.502 (0.469-0.541) & 0.359 (0.333-0.387) & 0.297 (0.272-0.320) \\
\midrule
$\ours$ & Fine & \textbf{0.706 (0.667-0.739)} & \textbf{0.606 (0.573-0.633)} & \textbf{0.568 (0.538-0.597)} \\
\midrule
\midrule
\virchow\cite{vorontsov2024foundation} \mean & Coarse & 0.797 (0.759-0.830) & 0.702 (0.670-0.726) & 0.648 (0.618-0.670) \\
\gigapath\cite{xu2024whole} \mean & Coarse & 0.782 (0.741-0.820) & 0.698 (0.665-0.724) & 0.642 (0.611-0.670) \\
\chief\cite{wang2024chief} \mean & Coarse & 0.659 (0.619-0.696) & 0.534 (0.502-0.563) & 0.483 (0.455-0.511) \\
\conch\cite{} \mean & Coarse & \underline{0.857 (0.829-0.880)} & \underline{0.774 (0.746-0.796)} & 0.731 (0.704-0.753) \\
\midrule
\prism\cite{shaikovski2024prism} & Coarse & 0.845 (0.818-0.874) & 0.772 (0.746-0.793) & \underline{0.738 (0.713-0.760)} \\
\gigapath\cite{xu2024whole} & Coarse & 0.798 (0.763-0.828) & 0.696 (0.667-0.720) & 0.635 (0.608-0.659) \\
\chief\cite{wang2024chief} & Coarse & 0.697 (0.661-0.740) & 0.582 (0.551-0.613) & 0.526 (0.497-0.555) \\
\midrule
$\ours$ & Coarse & \textbf{0.904 (0.883-0.928)} & \textbf{0.854 (0.834-0.873)} & \textbf{0.831 (0.809-0.849)} \\
\bottomrule
\end{tabular}

}
\label{tab:retrieval-ebrains}
\end{table}

\begin{table}[h]
\centering
\caption{\textbf{Tissue type retrieval (10 classes) performance using CPTAC.} Best in \textbf{bold}, second best is \underline{underlined}. \textbf{mAP@K}: mean average precision using top-k retrieved examples. CI: 95\% confidence interval.}
\begin{tabular}{llll}
\toprule
Model & mAP@1 (CI) & mAP@5 (CI) & mAP@10 (CI) \\
\midrule
\virchow\cite{} \mean  & 0.935 ± 0.004 & 0.818 ± 0.003 & 0.754 ± 0.004 \\
\gigapath\cite{} \mean & 0.943 ± 0.005 & 0.831 ± 0.007 & 0.753 ± 0.005 \\
\chief\cite{} \mean & 0.910 ± 0.003 & 0.778 ± 0.002 & 0.698 ± 0.003 \\
\conch\cite{} \mean & \underline{0.955 ± 0.002} & \underline{0.890 ± 0.004} & 0.848 ± 0.004 \\
\midrule
\prism\cite{} & 0.944 ± 0.003 & 0.883 ± 0.004 & \underline{0.850 ± 0.005} \\
\gigapath\cite{} & 0.943 ± 0.006 & 0.830 ± 0.006 & 0.749 ± 0.005 \\
\chief\cite{} & 0.916 ± 0.002 & 0.801 ± 0.006 & 0.734 ± 0.007 \\
\midrule
$\ours$ & \textbf{0.967 ± 0.003} & \textbf{0.900 ± 0.004} & \textbf{0.861 ± 0.006} \\
\bottomrule
\end{tabular}
\label{tab:retrieval-cptac_organ}
\end{table}
\begin{table}[h]
    \centering
    \caption{\textbf{Prompting experiments.} All samples from the train dataset are used for training, and all samples from the test dataset are used for testing. Best in \textbf{bold}, second best is \underline{underlined}. CI: 95\% confidence interval.}
    \scalebox{0.9}{
    \begin{tabular}{ccccc}
        \toprule
        Train Dataset & Test Dataset & Task & Method & Mean AUC (CI) \\ 
        \midrule

		
		\multirow{8}{*}{TCGA-BRCA\cite{}} & \multirow{8}{*}{MGB Breast} & \multirow{2}{*}{ER} & $\conch \mean$ & \underline{0.702 (0.673-0.744)} \\ 
		&  &  & $\ours$ Molecular & \textbf{0.779 (0.747-0.810)} \\ 
		\cline{3-5}
		&  & \multirow{2}{*}{HER2} & $\conch \mean$ & \underline{0.580 (0.522-0.623)} \\ 
		&  &  & $\ours$ Molecular & \textbf{0.713 (0.668-0.764)} \\ 
		\cline{3-5}
		&  & \multirow{2}{*}{PR} & $\conch \mean$ & \underline{0.662 (0.620-0.694)} \\ 
		&  &  & $\ours$ Molecular & \textbf{0.736 (0.698-0.771)} \\ 
		\cline{3-5}
		&  & \multirow{2}{*}{Subtype} & $\conch \mean$ & \textbf{0.916 (0.901-0.934)} \\ 
		&  &  & $\ours$ Molecular & \underline{0.882 (0.858-0.905)} \\ 
		\midrule
		\multirow{2}{*}{TCGA-GBMLGG\cite{}} & \multirow{2}{*}{EBRAINS\cite{roetzer2022ebrains}} & \multirow{2}{*}{IDH Status} & $\conch \mean$ & \underline{0.910 (0.892-0.927)} \\ 
		&  &  & $\ours$ Molecular & \textbf{0.960 (0.947-0.972)} \\ 
		\midrule
		\multirow{2}{*}{TCGA-NSCLC\cite{}} & \multirow{2}{*}{MGB Lung} & \multirow{2}{*}{Subtype} & $\conch \mean$ & \underline{0.889 (0.870-0.905)} \\ 
		&  &  & $\ours$ Molecular & \textbf{0.984 (0.974-0.990)} \\ 
		\midrule
  \multirow{2}{*}{TCGA-CCRCC\cite{}} & \multirow{2}{*}{CPTAC-CCRCC\cite{edwards2015cptac}} & \multirow{2}{*}{Overall Survival} & $\conch \mean$ & \underline{0.581 (0.497-0.653)} \\ 
		&  &  & $\ours$ Molecular & \textbf{0.687 (0.591-0.762)} \\ 
		\midrule
		\multirow{2}{*}{TCGA-PDAC\cite{}} & \multirow{2}{*}{CPTAC-PDAC\cite{edwards2015cptac}} & \multirow{2}{*}{Overall Survival} & $\conch \mean$ & \underline{0.531 (0.481-0.584)} \\ 
		&  &  & $\ours$ Molecular & \textbf{0.589 (0.543-0.635)} \\ 

        \bottomrule
    \end{tabular}
    }
    \label{tab:prompting]}
\end{table}

\begin{table}[h]
    \centering
    \caption{\textbf{Impact of pretraining size on performance.} We train $\ours$ on pretraining datasets of increasing size and report the average linear probe performance of each family of tasks (Table~\ref{tab:task_summary.morphological_subtyping}, \ref{tab:task_summary.tumor_grading}, \ref{tab:task_summary.molecular_subtyping}, \ref{tab:task_summary.mutation_prediction}, \ref{tab:task_summary.treatment_response}, \ref{tab:task_summary.survival_prediction}). Full benchmark refers to our proposed pan-tissue benchmark.}
    \scalebox{0.85}{
        \begin{tabular}{c|ccccc}
            \toprule
            \% of pretraining data & Clinical subtyping & IHC & Mutation & Treatment response and & Full benchmark \\
            (number of WSIs) & and grading & prediction & prediction & survival prediction & performance \\
            \midrule
            \midrule
            \conch \mean & 0.779 & 0.770 & 0.680 & 0.574 & 0.689 \\
            \midrule
            1 (470) & 0.806 & 0.791 & 0.716 & 0.583 & 0.714 \\
            5 (2356) & 0.829 & 0.804 & 0.727 & 0.585 & 0.725 \\
            25 (11791) & \textbf{0.836} & 0.802 & 0.727 & 0.613 & 0.732 \\
            50 (23584) & \textbf{0.836} & \underline{0.814} & 0.737 & 0.614 & 0.739 \\
            75 (35377) & 0.828 & 0.815 & \underline{0.752} & \underline{0.629} & \underline{0.747} \\
            100 (47171) & \underline{0.832} & \textbf{0.825} & \textbf{0.755} & \textbf{0.635} & \textbf{0.753} \\
            \bottomrule
            \end{tabular}
            }
\label{tab:data-scale-linprobe}
\end{table}

\begin{table}[h]
    \centering
    \caption{\textbf{Impact of model size on performance.} Number of pretraining heads in $\ours$ (1, 2, 4 heads) compared against $\prism$ and $\gigapath$. Evaluations done using linear probing on Full benchmark. \resnet \mean is ResNet50 model pretrained on ImageNet (IN)}
    \scalebox{1.0}{
        \begin{tabular}{l|cc}
            \toprule
            Model & Number of parameters & Full benchmark \\
            name & in WSI encoder (million) & performance \\
            \midrule
            \midrule
            \conch \mean & N/A & 0.689 \\
            \virchow\cite{vorontsov2024foundation} \mean & N/A & 0.662 \\
            \gigapath\cite{xu2024whole} \mean & N/A & 0.663 \\
            \chief\cite{wang2024chief} (Patch) \mean & N/A & 0.651 \\
            \resnet\cite{russakovsky2015imagenet} \mean & N/A & 0.560 \\
            \midrule
            \ours (1 head) & 5.0 & 0.734 \\
            \ours (2 heads) & 11.3 & \textbf{0.753} \\
            \ours (4 heads) & 19.7 & \underline{0.743} \\
            \ours (6 heads) & 28.1 & \underline{0.743} \\
            \ours (ViT) & 16.1 & 0.690 \\
            \midrule
            \prism\cite{shaikovski2024prism} & 45.0 & 0.690 \\
            \gigapath\cite{xu2024whole} & 85.1 & 0.654 \\
            \chief\cite{wang2024chief} (Slide) & 1.2 & 0.686 \\
            \bottomrule
            \end{tabular}
            }
\label{tab:model-scale-linprobe}
\end{table}


\begin{table}[h]
  \centering
    \caption{\textbf{\conch hyperparameters}. \conch was initialized with UNI (Vision Transformer Large). \textit{Batch size} refers to the total batch size across GPUs. Effective batch size used for optimization is \textit{batch size} $\times$ \textit{gradient accumulation steps}. Learning rate is increased from zero linearly to the \textit{peak learning rate} over the course of \textit{warmup steps} and decays back to zero following the \textit{learning rate scheduler}. The model was trained for 20 epochs with 1.26 million image / caption pairs where the maximum sequence length for captions is set to 128. Non-squared images are first padded to square and then resized to 448 $\times$ 448.}
  \begin{tabular}{@{}p{7.5cm}|p{3cm}}
    \toprule
    Hyperparameter & Values \\
    \midrule
    \midrule
    Image size & 448 $\times$ 448 \\
    Automatic mixed precision & FP16 \\
    Batch size & 256 \\
    Gradient accumulation steps & 3 \\
    Learning rate scheduler & Cosine \\
    Warmup steps & 250 \\
    Peak learning rate & 1e-4 \\
    AdamW $\beta$ & (0.9, 0.999) \\
    AdamW $\epsilon$ & 1e-8 \\
    Weight decay & 0.2 \\
    Softmax temperature & Learned \\
    Epochs & 20 \\
    \bottomrule
  \end{tabular}
  \label{tab:conch_hyperparams}
\end{table}

\begin{table}[h]
    \centering
    \caption{\textbf{CLIP hyperparameters in $\ours$ pretraining.}}
    \scalebox{0.95}{
        \begin{tabular}{l|l}
            \toprule
            Hyperparameter & Value \\
            \midrule
            \midrule
            GPU & 4$\times$ 80GB A100 \\
            Batch size per GPU & 300 \\
            Patches sampled during training & 512 \\
            AdamW $\beta$ & (0.9, 0.999) \\
            WSI embedder learning rate & 0.00005 \\
            RNA embedder learning rate & 0.00005 \\
            DNA embedder learning rate & 0.00005 \\
            WSI embedder weight decay & 0.0001 \\
            RNA embedder weight decay & 0.0001 \\
            DNA embedder weight decay & 0.0001 \\
            Learning rate schedule & Cosine \\
            Learning rate (start) & 0 \\
            Learning rate (post warmup) & 1e-5\\
            Learning rate (final) & 1e-8 \\
            Warmup epochs & 5 \\
            Max epochs & 101 \\
            \textsc{infoNCE} Temperature & 0.07 \\
            Automatic mixed precision & bfloaft16 \\
            Distributed Data Parallel Backend & GLOO \\
            Early stopping criteria & SmoothRank \cite{garrido2023rankme} \\
            \bottomrule
            \end{tabular}
            }
\label{tab:ssl-hparams}
\end{table}

\begin{table}[h]
    \centering
    \caption{\textbf{$\ours$ architectural hyperparameters.} WSI: wsi encoder; RNA: scGPT RNA encoder; DNA: DNA encoder}
    \scalebox{0.95}{
        \begin{tabular}{l|l|l}
            \toprule
            & Hyperparameter & Value \\
            \midrule
            \midrule
            \multirow{7}{*}{\rotatebox{90}{WSI}} & Pre-attention hidden dimension & 1024 \\
            & Pre-attention hidden layers & 2 \\
            & Pre-attention droput & 0.1 \\
            & Attention heads & 2 \\
            & Head activation & GeLU \\
            & Head dropout & 0.1 \\
            & Patch embedding dimension & 768 \\
            \midrule
            \multirow{6}{*}{\rotatebox{90}{RNA}} & Encoder & scGPT \cite{cui2024scgpt} \\
            & Normalization & $\log_2(\text{Transcripts per million})$ \\
            & scGPT data binning & None \\
            & scGPT pretrained weights & Pancancer \\
            & scGPT number of genes to sample & 1199 \\
            & scGPT hidden dim & 512 \\
            \midrule
            \multirow{5}{*}{\rotatebox{90}{DNA}} & Encoder & MLP \\
            & Hidden layers & 2 \\
            & Hidden dimension & 1024 \\
            & Activation & ReLU \\
            & Dropout & 0.2 \\
            \bottomrule
        \end{tabular}
    }
    \label{tab:arch-hparams}
\end{table}

\begin{table}[h!]
    \centering
    \caption{\textbf{Label breakdown for the MGB-BRCA morphological subtyping task.} Each sample corresponds to a WSI.}
    \begin{tabular}{lr}
    \toprule
    Grade & \# Samples \\
    \midrule
    Invasive ductal carcinoma (IDC) & 981 \\
    Invasive lobular carcinoma (ILC) & 283 \\
    \bottomrule
    \end{tabular}
    \label{tab:label_counts_mgb_brca_subtype}
\end{table}

\clearpage
\begin{table}[h!]
    \centering
    \caption{\textbf{Label breakdown for the MGB-Lung morphological subtyping task.} Each sample corresponds to a WSI.}
    \begin{tabular}{lr}
    \toprule
    Grade & \# Samples \\
    \midrule
    Lung adenocarcinoma (LUAD) & 1616 \\
    Lung squamous cell carcinoma (LUSC) & 325 \\
    \bottomrule
    \end{tabular}
    \label{tab:label_counts_mgb_lung_subtype}
\end{table}

\begin{table}[h!]
    \centering
    \caption{\textbf{Label breakdown for the IMP tumor grading task.} Each sample corresponds to a WSI.}
    \begin{tabular}{lr}
    \toprule
    Grade & \# Samples \\
    \midrule
    0 (non-neoplastic) & 847 \\
    1 (low-grade) & 2847 \\
    2 (high-grade) & 1639 \\
    \bottomrule
    \end{tabular}
    \label{tab:label_counts_imp_grade}
\end{table}

\begin{table}[h!]
    \centering
    \caption{\textbf{Label breakdown for the PANDA tumor grading task.} Each sample corresponds to a WSI.}
    \begin{tabular}{lr}
    \toprule
    ISUP Grade & \# Samples \\
    \midrule
    0 & 2603 \\
    1 & 2399 \\
    2 & 1209 \\
    4 & 1124 \\
    3 & 1118 \\
    5 & 1102 \\
    \bottomrule
    \end{tabular}
    \label{tab:label_counts_panda_isup_grade}
\end{table}

\begin{table}[h!]
    \centering
    \caption{\textbf{Label breakdown for the BRACS coarse-grained morphological subtyping task.} Each sample corresponds to a WSI.}
    \begin{tabular}{lr}
    \toprule
    Label & \# Samples \\
    \midrule
    Benign tumor & 265 \\
    Malignant tumor & 193 \\
    Atypical tumor & 89 \\
    \bottomrule
    \end{tabular}
    \label{tab:label_counts_bracs_slidelevel_coarse}
\end{table}

\begin{table}[h!]
    \centering
    \caption{\textbf{Label breakdown for the BRACS fine-grained morphological subtyping task.} Each sample corresponds to a WSI.}
    \begin{tabular}{lr}
    \toprule
    Label & \# Samples \\
    \midrule
    Pathological benign & 147 \\
    Invasive carcinoma & 132 \\
    Usual ductal hyperplasia & 74 \\
    Ductal carcinoma \textit{in situ} & 61 \\
    Atypical ductal hyperplasia & 48 \\
    Normal & 44 \\
    Flat epithelial atypia & 41 \\
    \bottomrule
    \end{tabular}
    \label{tab:label_counts_bracs_slidelevel_fine}
\end{table}

\begin{table}[h!]
    \centering
    \caption{\textbf{Label breakdown for the EBRAINS coarse-grained diagnosis task.} Each sample corresponds to a WSI.}
    \begin{tabular}{lr}
    \toprule
    Label & \# Samples \\
    \midrule
    Adult-type diffuse gliomas & 837 \\
    Meningiomas & 430 \\
    Mesenchymal, non-meningothelial tumours involving the CNS & 190 \\
    Tumours of the sellar region & 184 \\
    Circumscribed astrocytic gliomas  & 173 \\
    Ependymal Tumours & 96 \\
    Haematolymphoid tumours involving the CNS & 91 \\
    Glioneuronal and neuronal tumours & 88 \\
    Cranial and paraspinal nerve tumours & 81 \\
    Paediatric-type diffuse low-grade gliomas  & 70 \\
    Metastatic tumours & 47 \\
    Embryonal Tumors & 32 \\
    \bottomrule
    \end{tabular}
    \label{tab:label_counts_ebrains_diagnosis_group}
\end{table}

\begin{table}[!ht]
    \centering
    \caption{\textbf{Label breakdown for the EBRAINS fine-grained diagnosis task.} Each sample corresponds to a WSI.}
    \begin{tabular}{lr}
    \toprule
    Label & \# Samples \\
    \midrule
    Glioblastoma, IDH-wildtype & 474 \\
    Pilocytic astrocytoma & 173 \\
    Meningothelial meningioma & 104 \\
    Pituitary adenoma & 99 \\
    Anaplastic oligodendroglioma, IDH-mutant and 1p/19q codeleted & 91 \\
    Ganglioglioma & 88 \\
    Haemangioblastoma & 88 \\
    Adamantinomatous craniopharyngioma & 85 \\
    Oligodendroglioma, IDH-mutant and 1p/19q codeleted & 85 \\
    Atypical meningioma & 83 \\
    Schwannoma & 81 \\
    Diffuse astrocytoma, IDH-mutant & 70 \\
    Transitional meningioma & 68 \\
    Diffuse large B-cell lymphoma of the CNS & 59 \\
    Gliosarcoma & 59 \\
    Fibrous meningioma & 57 \\
    Anaplastic ependymoma & 50 \\
    Anaplastic astrocytoma, IDH-wildtype & 47 \\
    Metastatic tumours & 47 \\
    Anaplastic astrocytoma, IDH-mutant & 47 \\
    Ependymoma & 46 \\
    Anaplastic meningioma & 46 \\
    Secretory meningioma & 41 \\
    Lipoma & 38 \\
    Haemangiopericytoma & 34 \\
    Glioblastoma, IDH-mutant & 34 \\
    Medulloblastoma, non-WNT/non-SHH & 32 \\
    Langerhans cell histiocytosis & 32 \\
    Angiomatous meningioma & 31 \\
    Haemangioma & 30 \\
    \bottomrule
    \end{tabular}
    \label{tab:label_counts_ebrains_diagnosis}
\end{table}

\end{spacing}
\end{nolinenumbers}
\clearpage

\newpage
\begin{nolinenumbers}
\Heading{References}
\begin{spacing}{0.9}
\bibliographystyle{naturemag}
\bibliography{main}

\begin{thebibliography}{10}
\expandafter\ifx\csname url\endcsname\relax
  \def\url#1{\texttt{#1}}\fi
\expandafter\ifx\csname urlprefix\endcsname\relax\def\urlprefix{URL }\fi
\providecommand{\bibinfo}[2]{#2}
\providecommand{\eprint}[2][]{\url{#2}}

\bibitem{song2023artificial}
\bibinfo{author}{Song, A.~H.} \emph{et~al.}
\newblock \bibinfo{title}{Artificial intelligence for digital and computational pathology}.
\newblock \emph{\bibinfo{journal}{Nature Reviews Bioengineering}}  (\bibinfo{year}{2023}).
\newblock \urlprefix\url{https://doi.org/10.1038/s44222-023-00096-8}.

\bibitem{van2021deep}
\bibinfo{author}{Van~der Laak, J.}, \bibinfo{author}{Litjens, G.} \& \bibinfo{author}{Ciompi, F.}
\newblock \bibinfo{title}{Deep learning in histopathology: the path to the clinic}.
\newblock \emph{\bibinfo{journal}{Nature Medicine}} \textbf{\bibinfo{volume}{27}}, \bibinfo{pages}{775--784} (\bibinfo{year}{2021}).

\bibitem{bulten2022artificial}
\bibinfo{author}{Bulten, W.} \emph{et~al.}
\newblock \bibinfo{title}{Artificial intelligence for diagnosis and gleason grading of prostate cancer: the panda challenge}.
\newblock \emph{\bibinfo{journal}{Nature medicine}} \textbf{\bibinfo{volume}{28}}, \bibinfo{pages}{154--163} (\bibinfo{year}{2022}).

\bibitem{bejnordi2017diagnostic}
\bibinfo{author}{Bejnordi, B.~E.} \emph{et~al.}
\newblock \bibinfo{title}{Diagnostic assessment of deep learning algorithms for detection of lymph node metastases in women with breast cancer}.
\newblock \emph{\bibinfo{journal}{JAMA}} \textbf{\bibinfo{volume}{318}}, \bibinfo{pages}{2199--2210} (\bibinfo{year}{2017}).

\bibitem{chen2024towards}
\bibinfo{author}{Chen, R.~J.} \emph{et~al.}
\newblock \bibinfo{title}{Towards a general-purpose foundation model for computational pathology}.
\newblock \emph{\bibinfo{journal}{Nature Medicine}}  (\bibinfo{year}{2024}).

\bibitem{xu2024whole}
\bibinfo{author}{Xu, H.} \emph{et~al.}
\newblock \bibinfo{title}{A whole-slide foundation model for digital pathology from real-world data}.
\newblock \emph{\bibinfo{journal}{Nature}} \bibinfo{pages}{1--8} (\bibinfo{year}{2024}).

\bibitem{vorontsov2024foundation}
\bibinfo{author}{Vorontsov, E.} \emph{et~al.}
\newblock \bibinfo{title}{A foundation model for clinical-grade computational pathology and rare cancers detection}.
\newblock \emph{\bibinfo{journal}{Nature Medicine}} \bibinfo{pages}{1--12} (\bibinfo{year}{2024}).

\bibitem{chen2022scaling}
\bibinfo{author}{Chen, R.~J.} \emph{et~al.}
\newblock \bibinfo{title}{Scaling vision transformers to gigapixel images via hierarchical self-supervised learning}.
\newblock In \emph{\bibinfo{booktitle}{Proceedings of the IEEE/CVF Conference on Computer Vision and Pattern Recognition}} (\bibinfo{year}{2022}).

\bibitem{jaume2024transcriptomics}
\bibinfo{author}{Jaume, G.} \emph{et~al.}
\newblock \bibinfo{title}{Transcriptomics-guided slide representation learning in computational pathology}.
\newblock In \emph{\bibinfo{booktitle}{Proceedings of the IEEE/CVF Conference on Computer Vision and Pattern Recognition}}, \bibinfo{pages}{9632--9644} (\bibinfo{year}{2024}).

\bibitem{jaume2024multistain}
\bibinfo{author}{Jaume, G.} \emph{et~al.}
\newblock \bibinfo{title}{Multistain pretraining for slide representation learning in pathology}.
\newblock In \emph{\bibinfo{booktitle}{Computer Vision -- European Conference on Computer Vision 2024}}, \bibinfo{pages}{19--37} (\bibinfo{publisher}{Springer Nature Switzerland}, \bibinfo{address}{Cham}, \bibinfo{year}{2025}).

\bibitem{wang2024chief}
\bibinfo{author}{Wang, X.} \emph{et~al.}
\newblock \bibinfo{title}{{A pathology foundation model for cancer diagnosis and prognosis prediction}}.
\newblock \emph{\bibinfo{journal}{Nature}} \bibinfo{pages}{1--9} (\bibinfo{year}{2024}).

\bibitem{gtex2015genotype}
\bibinfo{author}{Consortium, G.} \emph{et~al.}
\newblock \bibinfo{title}{The genotype-tissue expression (gtex) pilot analysis: multitissue gene regulation in humans}.
\newblock \emph{\bibinfo{journal}{Science}} \textbf{\bibinfo{volume}{348}}, \bibinfo{pages}{648--660} (\bibinfo{year}{2015}).

\bibitem{shaikovski2024prism}
\bibinfo{author}{Shaikovski, G.} \emph{et~al.}
\newblock \bibinfo{title}{Prism: A multi-modal generative foundation model for slide-level histopathology}.
\newblock \emph{\bibinfo{journal}{arXiv preprint arXiv:2405.10254}}  (\bibinfo{year}{2024}).

\bibitem{van2008visualizing}
\bibinfo{author}{Van~der Maaten, L.} \& \bibinfo{author}{Hinton, G.}
\newblock \bibinfo{title}{Visualizing data using t-sne.}
\newblock \emph{\bibinfo{journal}{Journal of machine learning research}} \textbf{\bibinfo{volume}{9}} (\bibinfo{year}{2008}).

\bibitem{lu2024towards}
\bibinfo{author}{Lu, M.} \emph{et~al.}
\newblock \bibinfo{title}{Towards a visual-language foundation model for computational pathology}.
\newblock \emph{\bibinfo{journal}{Nature Medicine}}  (\bibinfo{year}{2024}).

\bibitem{dosovitskiy2021image}
\bibinfo{author}{Dosovitskiy, A.} \emph{et~al.}
\newblock \bibinfo{title}{An image is worth 16x16 words: Transformers for image recognition at scale}.
\newblock In \emph{\bibinfo{booktitle}{International Conference on Learning Representations}} (\bibinfo{year}{2021}).

\bibitem{vaswani2017attention}
\bibinfo{author}{Vaswani, A.} \emph{et~al.}
\newblock \bibinfo{title}{Attention is all you need}.
\newblock \emph{\bibinfo{journal}{Advances in neural information processing systems}} \textbf{\bibinfo{volume}{30}} (\bibinfo{year}{2017}).

\bibitem{cui2024scgpt}
\bibinfo{author}{Cui, H.} \emph{et~al.}
\newblock \bibinfo{title}{{scGPT:} {Toward} building a foundation model for single-cell multi-omics using generative ai}.
\newblock \emph{\bibinfo{journal}{Nature Methods}} \bibinfo{pages}{1--11} (\bibinfo{year}{2024}).

\bibitem{chen2020pathomic}
\bibinfo{author}{Chen, R.~J.} \emph{et~al.}
\newblock \bibinfo{title}{Pathomic fusion: an integrated framework for fusing histopathology and genomic features for cancer diagnosis and prognosis}.
\newblock \emph{\bibinfo{journal}{IEEE Transactions on Medical Imaging}}  (\bibinfo{year}{2020}).

\bibitem{jaegle2021perceiver}
\bibinfo{author}{Jaegle, A.} \emph{et~al.}
\newblock \bibinfo{title}{Perceiver: General perception with iterative attention}.
\newblock In \emph{\bibinfo{booktitle}{International conference on machine learning}}, \bibinfo{pages}{4651--4664} (\bibinfo{organization}{PMLR}, \bibinfo{year}{2021}).

\bibitem{oquab2023dinov2}
\bibinfo{author}{Oquab, M.} \emph{et~al.}
\newblock \bibinfo{title}{{DINO}v2: Learning robust visual features without supervision}.
\newblock \emph{\bibinfo{journal}{Transactions on Machine Learning Research}}  (\bibinfo{year}{2024}).

\bibitem{ding2023longnet}
\bibinfo{author}{Ding, J.} \emph{et~al.}
\newblock \bibinfo{title}{Longnet: Scaling transformers to 1,000,000,000 tokens}.
\newblock \emph{\bibinfo{journal}{arXiv preprint arXiv:2307.02486}}  (\bibinfo{year}{2023}).

\bibitem{wang2021transpath}
\bibinfo{author}{Wang, X.} \emph{et~al.}
\newblock \bibinfo{title}{Transpath: Transformer-based self-supervised learning for histopathological image classification}.
\newblock In \emph{\bibinfo{booktitle}{International Conference on Medical Image Computing and Computer-Assisted Intervention}} (\bibinfo{year}{2021}).

\bibitem{wang2022transformer}
\bibinfo{author}{Wang, X.} \emph{et~al.}
\newblock \bibinfo{title}{Transformer-based unsupervised contrastive learning for histopathological image classification}.
\newblock \emph{\bibinfo{journal}{Medical Image Analysis}} \textbf{\bibinfo{volume}{81}}, \bibinfo{pages}{102559} (\bibinfo{year}{2022}).

\bibitem{ilse2018attention}
\bibinfo{author}{Ilse, M.}, \bibinfo{author}{Tomczak, J.} \& \bibinfo{author}{Welling, M.}
\newblock \bibinfo{title}{Attention-based deep multiple instance learning}.
\newblock In \emph{\bibinfo{booktitle}{Proceedings of the 35th International Conference on Machine Learning}}, \bibinfo{pages}{2132--2141} (\bibinfo{year}{2018}).

\bibitem{oliveira2023cad}
\bibinfo{author}{Oliveira, S.~P.} \emph{et~al.}
\newblock \bibinfo{title}{{A CAD system for automatic dysplasia grading on H{\&}E cervical whole-slide images}}.
\newblock \emph{\bibinfo{journal}{Sci. Rep.}} \textbf{\bibinfo{volume}{13}}, \bibinfo{pages}{1--12} (\bibinfo{year}{2023}).

\bibitem{roetzer2022ebrains}
\bibinfo{author}{Roetzer-Pejrimovsky, T.} \emph{et~al.}
\newblock \bibinfo{title}{{The Digital Brain Tumour Atlas, an open histopathology resource}}.
\newblock \emph{\bibinfo{journal}{Scientific Data}} \textbf{\bibinfo{volume}{9}}, \bibinfo{pages}{1--6} (\bibinfo{year}{2022}).

\bibitem{brancati2021bracs}
\bibinfo{author}{Brancati, N.} \emph{et~al.}
\newblock \bibinfo{title}{{{BRACS}: {A} Dataset for BReAst Carcinoma Subtyping in H\&amp;E Histology Images}}.
\newblock \emph{\bibinfo{journal}{Database}} \textbf{\bibinfo{volume}{2022}} (\bibinfo{year}{2022}).

\bibitem{xu2021predicting}
\bibinfo{author}{Xu, F.} \emph{et~al.}
\newblock \bibinfo{title}{Predicting axillary lymph node metastasis in early breast cancer using deep learning on primary tumor biopsy slides}.
\newblock \emph{\bibinfo{journal}{Frontiers in oncology}} \textbf{\bibinfo{volume}{11}}, \bibinfo{pages}{759007} (\bibinfo{year}{2021}).

\bibitem{radford2021learning}
\bibinfo{author}{Radford, A.} \emph{et~al.}
\newblock \bibinfo{title}{Learning transferable visual models from natural language supervision}.
\newblock \emph{\bibinfo{journal}{arXiv preprint arXiv:2103.00020}}  (\bibinfo{year}{2021}).

\bibitem{zheng2014anchored}
\bibinfo{author}{Zheng, Z.} \emph{et~al.}
\newblock \bibinfo{title}{Anchored multiplex pcr for targeted next-generation sequencing}.
\newblock \emph{\bibinfo{journal}{Nature medicine}} \textbf{\bibinfo{volume}{20}}, \bibinfo{pages}{1479--1484} (\bibinfo{year}{2014}).

\bibitem{garcia2017validation}
\bibinfo{author}{Garcia, E.~P.} \emph{et~al.}
\newblock \bibinfo{title}{Validation of oncopanel: a targeted next-generation sequencing assay for the detection of somatic variants in cancer}.
\newblock \emph{\bibinfo{journal}{Archives of Pathology and Laboratory Medicine}} \textbf{\bibinfo{volume}{141}}, \bibinfo{pages}{751--758} (\bibinfo{year}{2017}).

\bibitem{jaume2024modeling}
\bibinfo{author}{Jaume, G.} \emph{et~al.}
\newblock \bibinfo{title}{Modeling dense multimodal interactions between biological pathways and histology for survival prediction}.
\newblock In \emph{\bibinfo{booktitle}{Proceedings of the IEEE/CVF Conference on Computer Vision and Pattern Recognition}} (\bibinfo{year}{2024}).

\bibitem{vaidya2024demographic}
\bibinfo{author}{Vaidya, A.} \emph{et~al.}
\newblock \bibinfo{title}{Demographic bias in misdiagnosis by computational pathology models}.
\newblock \emph{\bibinfo{journal}{Nature Medicine}} \textbf{\bibinfo{volume}{30}}, \bibinfo{pages}{1174--1190} (\bibinfo{year}{2024}).

\bibitem{acosta2022intratumoral}
\bibinfo{author}{Acosta, P.~H.} \emph{et~al.}
\newblock \bibinfo{title}{Intratumoral resolution of driver gene mutation heterogeneity in renal cancer using deep learning}.
\newblock \emph{\bibinfo{journal}{Cancer research}} \textbf{\bibinfo{volume}{82}}, \bibinfo{pages}{2792--2806} (\bibinfo{year}{2022}).

\bibitem{neto2024interpretable}
\bibinfo{author}{Neto, P.~C.} \emph{et~al.}
\newblock \bibinfo{title}{An interpretable machine learning system for colorectal cancer diagnosis from pathology slides}.
\newblock \emph{\bibinfo{journal}{NPJ precision oncology}} \textbf{\bibinfo{volume}{8}}, \bibinfo{pages}{56} (\bibinfo{year}{2024}).

\bibitem{pati2022hierarchical}
\bibinfo{author}{Pati, P.} \emph{et~al.}
\newblock \bibinfo{title}{Hierarchical graph representations in digital pathology}.
\newblock \emph{\bibinfo{journal}{Medical Image Analysis}} \textbf{\bibinfo{volume}{75}}, \bibinfo{pages}{102264} (\bibinfo{year}{2022}).

\bibitem{edwards2015cptac}
\bibinfo{author}{Edwards, N.~J.} \emph{et~al.}
\newblock \bibinfo{title}{{The CPTAC Data Portal: A Resource for Cancer Proteomics Research}}.
\newblock \emph{\bibinfo{journal}{Journal of Proteome Research}} \textbf{\bibinfo{volume}{14}}, \bibinfo{pages}{2707--2713} (\bibinfo{year}{2015}).

\bibitem{thangudu2020cptac}
\bibinfo{author}{Thangudu, R.~R.} \emph{et~al.}
\newblock \bibinfo{title}{{Abstract LB-242: Proteomic Data Commons: A resource for proteogenomic analysis}}.
\newblock \emph{\bibinfo{journal}{Cancer Research}} \textbf{\bibinfo{volume}{80}} (\bibinfo{year}{2020}).

\bibitem{liao2023proteogenomics}
\bibinfo{author}{Liao, Y.} \emph{et~al.}
\newblock \bibinfo{title}{A proteogenomics data-driven knowledge base of human cancer}.
\newblock \emph{\bibinfo{journal}{Cell Systems}} \textbf{\bibinfo{volume}{14}}, \bibinfo{pages}{777--787} (\bibinfo{year}{2023}).

\bibitem{wang2022weakly}
\bibinfo{author}{Wang, C.-W.} \emph{et~al.}
\newblock \bibinfo{title}{{Weakly supervised deep learning for prediction of treatment effectiveness on ovarian cancer from histopathology images}}.
\newblock \emph{\bibinfo{journal}{Computerized Medical Imaging and Graphics}} \textbf{\bibinfo{volume}{99}}, \bibinfo{pages}{102093} (\bibinfo{year}{2022}).

\bibitem{wilkinson2021nascent}
\bibinfo{author}{Wilkinson, S.} \emph{et~al.}
\newblock \bibinfo{title}{{Nascent Prostate Cancer Heterogeneity Drives Evolution and Resistance to Intense Hormonal Therapy}}.
\newblock \emph{\bibinfo{journal}{European Urology}} \textbf{\bibinfo{volume}{80}}, \bibinfo{pages}{746--757} (\bibinfo{year}{2021}).

\bibitem{touat2020mechanisms}
\bibinfo{author}{Touat, M.} \emph{et~al.}
\newblock \bibinfo{title}{Mechanisms and therapeutic implications of hypermutation in gliomas}.
\newblock \emph{\bibinfo{journal}{Nature}} \textbf{\bibinfo{volume}{580}}, \bibinfo{pages}{517--523} (\bibinfo{year}{2020}).

\bibitem{ayoub2024path}
\bibinfo{author}{Ayoub, G.} \emph{et~al.}
\newblock \bibinfo{title}{Path-50. ai-powered automated tissue segmentation improves outcome stratification in glioblastoma}.
\newblock \emph{\bibinfo{journal}{Neuro-Oncology}} \textbf{\bibinfo{volume}{26}}, \bibinfo{pages}{viii190--viii190} (\bibinfo{year}{2024}).

\bibitem{domingo2015next}
\bibinfo{author}{Domingo-Musibay, E.} \& \bibinfo{author}{Galanis, E.}
\newblock \bibinfo{title}{What next for newly diagnosed glioblastoma?}
\newblock \emph{\bibinfo{journal}{Future Oncology}} \textbf{\bibinfo{volume}{11}}, \bibinfo{pages}{3273--3283} (\bibinfo{year}{2015}).

\bibitem{martel2019assessment}
\bibinfo{author}{Martel, A.}, \bibinfo{author}{Nofech-Mozes, S.}, \bibinfo{author}{Salama, S.}, \bibinfo{author}{Akbar, S.} \& \bibinfo{author}{Peikari, M.}
\newblock \bibinfo{title}{Assessment of residual breast cancer cellularity after neoadjuvant chemotherapy using digital pathology [data set]}.
\newblock \emph{\bibinfo{journal}{The Cancer Imaging Archive}}  (\bibinfo{year}{2019}).

\bibitem{myles2024leveraging}
\bibinfo{author}{Myles, C.}, \bibinfo{author}{Um, I.~H.}, \bibinfo{author}{Harrison, D.~J.} \& \bibinfo{author}{Harris-Birtill, D.}
\newblock \bibinfo{title}{Leveraging foundation models for enhanced detection of colorectal cancer biomarkers in small datasets}.
\newblock In \emph{\bibinfo{booktitle}{Annual Conference on Medical Image Understanding and Analysis}}, \bibinfo{pages}{329--343} (\bibinfo{organization}{Springer}, \bibinfo{year}{2024}).

\bibitem{galland2022mbc}
\bibinfo{author}{Galland, L.} \emph{et~al.}
\newblock \bibinfo{title}{{Efficacy of platinum-based chemotherapy in metastatic breast cancer and HRD biomarkers: utility of exome sequencing}}.
\newblock \emph{\bibinfo{journal}{npj Breast Cancer}} \textbf{\bibinfo{volume}{8}}, \bibinfo{pages}{1--12} (\bibinfo{year}{2022}).

\bibitem{bergstrom2024deep}
\bibinfo{author}{Bergstrom, E.~N.} \emph{et~al.}
\newblock \bibinfo{title}{Deep learning artificial intelligence predicts homologous recombination deficiency and platinum response from histologic slides}.
\newblock \emph{\bibinfo{journal}{Journal of Clinical Oncology}} \textbf{\bibinfo{volume}{42}}, \bibinfo{pages}{3550--3560} (\bibinfo{year}{2024}).

\bibitem{boehm2022multimodal}
\bibinfo{author}{Boehm, K.~M.} \emph{et~al.}
\newblock \bibinfo{title}{{Multimodal data integration using machine learning improves risk stratification of high-grade serous ovarian cancer}}.
\newblock \emph{\bibinfo{journal}{Nat. Cancer}} \textbf{\bibinfo{volume}{3}}, \bibinfo{pages}{723--733} (\bibinfo{year}{2022}).

\bibitem{lin2017feature}
\bibinfo{author}{Lin, T.-Y.} \emph{et~al.}
\newblock \bibinfo{title}{Feature pyramid networks for object detection}.
\newblock In \emph{\bibinfo{booktitle}{Proceedings of the IEEE Conference on Computer Vision and Pattern Recognition (CVPR)}} (\bibinfo{year}{2017}).

\bibitem{yu2022coca}
\bibinfo{author}{Yu, J.} \emph{et~al.}
\newblock \bibinfo{title}{{CoCa: Contrastive Captioners are Image-Text Foundation Models}}.
\newblock \emph{\bibinfo{journal}{Transactions on Machine Learning Research}}  (\bibinfo{year}{2022}).

\bibitem{lu2024pathchat}
\bibinfo{author}{Lu, M.~Y.} \emph{et~al.}
\newblock \bibinfo{title}{{A multimodal generative AI copilot for human pathology}}.
\newblock \emph{\bibinfo{journal}{Nature}} \bibinfo{pages}{1--3} (\bibinfo{year}{2024}).

\bibitem{lu2021data}
\bibinfo{author}{Lu, M.~Y.} \emph{et~al.}
\newblock \bibinfo{title}{Data efficient and weakly supervised computational pathology on whole slide images}.
\newblock \emph{\bibinfo{journal}{Nature Biomedical Engineering}}  (\bibinfo{year}{2021}).

\bibitem{wang2024path}
\bibinfo{author}{Wang, H.} \emph{et~al.}
\newblock \bibinfo{title}{Path-gptomic: A balanced multi-modal learning framework for survival outcome prediction}.
\newblock \emph{\bibinfo{journal}{arXiv preprint arXiv:2403.11375}}  (\bibinfo{year}{2024}).

\bibitem{garrido2023rankme}
\bibinfo{author}{Garrido, Q.}, \bibinfo{author}{Balestriero, R.}, \bibinfo{author}{Najman, L.} \& \bibinfo{author}{Lecun, Y.}
\newblock \bibinfo{title}{Rankme: Assessing the downstream performance of pretrained self-supervised representations by their rank}.
\newblock In \emph{\bibinfo{booktitle}{International conference on machine learning}}, \bibinfo{pages}{10929--10974} (\bibinfo{organization}{PMLR}, \bibinfo{year}{2023}).

\bibitem{he2016deep}
\bibinfo{author}{He, K.}, \bibinfo{author}{Zhang, X.}, \bibinfo{author}{Ren, S.} \& \bibinfo{author}{Sun, J.}
\newblock \bibinfo{title}{Deep residual learning for image recognition}.
\newblock In \emph{\bibinfo{booktitle}{Proceedings of the IEEE conference on computer vision and pattern recognition}}, \bibinfo{pages}{770--778} (\bibinfo{year}{2016}).

\bibitem{deng2009imagenet}
\bibinfo{author}{Deng, J.} \emph{et~al.}
\newblock \bibinfo{title}{Imagenet: A large-scale hierarchical image database}.
\newblock In \emph{\bibinfo{booktitle}{2009 IEEE conference on computer vision and pattern recognition}}, \bibinfo{pages}{248--255} (\bibinfo{organization}{Ieee}, \bibinfo{year}{2009}).

\bibitem{sksurv}
\bibinfo{author}{P{\"o}lsterl, S.}
\newblock \bibinfo{title}{scikit-survival: A library for time-to-event analysis built on top of scikit-learn}.
\newblock \emph{\bibinfo{journal}{Journal of Machine Learning Research}} \textbf{\bibinfo{volume}{21}}, \bibinfo{pages}{1--6} (\bibinfo{year}{2020}).
\newblock \urlprefix\url{http://jmlr.org/papers/v21/20-729.html}.

\bibitem{dunn1961multiple}
\bibinfo{author}{Dunn, O.~J.}
\newblock \bibinfo{title}{Multiple comparisons among means}.
\newblock \emph{\bibinfo{journal}{Journal of the American statistical association}} \textbf{\bibinfo{volume}{56}}, \bibinfo{pages}{52--64} (\bibinfo{year}{1961}).

\bibitem{searle1980population}
\bibinfo{author}{Searle, S.~R.}, \bibinfo{author}{Speed, F.~M.} \& \bibinfo{author}{Milliken, G.~A.}
\newblock \bibinfo{title}{Population marginal means in the linear model: an alternative to least squares means}.
\newblock \emph{\bibinfo{journal}{The American Statistician}} \textbf{\bibinfo{volume}{34}}, \bibinfo{pages}{216--221} (\bibinfo{year}{1980}).

\bibitem{robertson2024decoding}
\bibinfo{author}{Robertson, H.} \emph{et~al.}
\newblock \bibinfo{title}{Decoding the hallmarks of allograft dysfunction with a comprehensive pan-organ transcriptomic atlas}.
\newblock \emph{\bibinfo{journal}{Nature Medicine}} \bibinfo{pages}{1--10} (\bibinfo{year}{2024}).

\bibitem{pedregosa2011scikit}
\bibinfo{author}{Pedregosa, F.} \emph{et~al.}
\newblock \bibinfo{title}{Scikit-learn: Machine learning in python}.
\newblock \emph{\bibinfo{journal}{the Journal of machine Learning research}} \textbf{\bibinfo{volume}{12}}, \bibinfo{pages}{2825--2830} (\bibinfo{year}{2011}).

\bibitem{kolesnikov2019revisiting}
\bibinfo{author}{Kolesnikov, A.}, \bibinfo{author}{Zhai, X.} \& \bibinfo{author}{Beyer, L.}
\newblock \bibinfo{title}{Revisiting self-supervised visual representation learning}.
\newblock In \emph{\bibinfo{booktitle}{Proceedings of the IEEE/CVF Conference on Computer Vision and Pattern Recognition (CVPR)}} (\bibinfo{year}{2019}).

\bibitem{russakovsky2015imagenet}
\bibinfo{author}{Russakovsky, O.} \emph{et~al.}
\newblock \bibinfo{title}{Imagenet large scale visual recognition challenge}.
\newblock \emph{\bibinfo{journal}{International journal of computer vision}} \textbf{\bibinfo{volume}{115}}, \bibinfo{pages}{211--252} (\bibinfo{year}{2015}).

\end{thebibliography}
\end{spacing}
\end{nolinenumbers}
\clearpage

\end{document}